\documentclass[lettersize,journal]{IEEEtran}

\DeclareUnicodeCharacter{2113}{\ensuremath{\ell}}

\usepackage{fancyhdr}
\usepackage{extramarks}

\usepackage{amsthm}
\usepackage{amsmath}
\usepackage{amsfonts}
\usepackage{comment,color}
\usepackage{csquotes}
\usepackage{xspace}
\usepackage{siunitx}
\usepackage{lmodern}  

\newtheorem{theorem}{Theorem}[section]

\newtheorem{lemma}[theorem]{Lemma}
\newtheorem{definition}{Definition}[section]

\newcommand{\yl}[1]{\textcolor{blue}{#1}}

\usepackage{amsmath} 
\usepackage{tikz}
\DeclareMathOperator*{\argmin}{arg\,min}

\newcommand{\prox}{\textbf{prox}}
\newcommand{\proj}{\textbf{proj}}
\usepackage{float}
\usepackage{algorithm}
\usepackage[ruled,algo2e]{algorithm2e}
\usepackage{amsfonts}
\usepackage{amssymb}
\usepackage{algpseudocode}
\usepackage[all]{xy}
\usepackage{tikz-cd}
\usepackage{multirow}
\usepackage{bm}
\usepackage{booktabs}
\usepackage[numbers,sort&compress]{natbib}
\usepackage{graphicx,subfigure}
\usepackage{palatino}
\usepackage{caption}
\usepackage{subcaption}
\usepackage{graphicx}

\usepackage[colorlinks,linkcolor=black,anchorcolor=black,citecolor=black,urlcolor=blue]{hyperref}

\usepackage{booktabs}
\usepackage{mathtools}
\usepackage{amssymb}
\usepackage{capt-of}
\usepackage{mciteplus}
\usepackage{mathrsfs}
\usepackage{parskip}
\usepackage{soul}
\usepackage{textcomp}
\usepackage[colaction]{multicol}
\usepackage[switch]{lineno}
\usepackage{lipsum}
\usepackage{etoolbox}
\usepackage{longtable}
\usepackage{array,bm}
\usepackage{tablefootnote}
\usepackage{ragged2e}
\usepackage{lscape}
\usepackage{amsmath,amsfonts}

\usepackage{array}
\usepackage{textcomp}
\usepackage{stfloats}
\usepackage{url}
\usepackage{verbatim}

\newcolumntype{C}[1]{>{\centering\arraybackslash}p{#1}}
\usetikzlibrary{automata,positioning}
\newcommand{\h}[1]{\mathbf{#1}}

\usetikzlibrary{automata,positioning}

\usepackage{tikz,xcolor,hyperref}
\definecolor{lime}{HTML}{A6CE39}
\DeclareRobustCommand{\orcidicon}{%
	\begin{tikzpicture}
	\draw[lime, fill=lime] (0,0)
	circle [radius=0.16]
	node[white] {{\fontfamily{qag}\selectfont \tiny ID}};
	\draw[white, fill=white] (-0.0625,0.095)
	circle [radius=0.007];
	\end{tikzpicture}
	\hspace{-2mm}
}
\foreach \x in {A, ..., Z}{%
	\expandafter\xdef\csname orcid\x\endcsname{\noexpand\href{https://orcid.org/\csname orcidauthor\x\endcsname}{\noexpand\orcidicon}}
}

\begin{document}
\setlength\parindent{0pt}
\title{A General Framework for Group Sparsity in Hyperspectral Unmixing Using Endmember Bundles}
%Auction Active learning
%MALADY: Multiclass Active Learning with Auction Dynamics  on Graphs  %% probably too NEURIPS-esque of an acronym, but it does fit.

\author{Gokul Bhusal, Yifei Lou, Cristina Garcia-Cardona, and Ekaterina Merkurjev 
\thanks{This work is supported in part by the NSF DMS-2052983 grant and NSF CAREER Award 2414705. CGC work was supported by the
U.S. Department of Energy, Office of Science, Office of Advanced Scientific Computing Research under Triad National
Security, LLC (‘Triad’) contract grant 89233218CNA000001 [FWP: LANLE2A2]. (Corresponding
author: Gokul Bhusal.)}

\thanks{G. Bhusal is with the Department of Mathematics, Michigan State University, MI 488824, USA. (email: bhusalgo@msu.edu)}
\thanks{Y. Lou is with the Department of Mathematics and the School of Data
Science and Society, The University of North Carolina at Chapel Hill, Chapel
Hill, NC 27599 USA. (email: yflou@unc.edu)}
\thanks{CG. Cardona is with Computer, Computational and Statistical Sciences Division
Los Alamos National Laboratory, Los Alamos, NM 87545 USA. (email: cgarciac@lanl.gov)}
%\thanks{J. Chanussot is with Univ. Grenoble Alpes, Inria, CNRS, Grenoble INP,
%LJK, Grenoble 38000, France. (e-mail: jocelyn.chanussot@grenoble-inp.fr)}
\thanks{E. Merkurjev is with the Department of Mathematics and the Department of Computational Mathematics, Science and Engineering, Michigan State University, MI 488824, USA. (email: merkurje@msu.edu)}
\thanks{The authors thank Dr. Jocelyn Chanussot for his valuable suggestions during the course of the project.}

}
\begin{comment}

\author{Gokul Bhusal$^1$, Yifei Lou$^3$, Cristina Garcia-Cardona$^4$,  Jocelyn Chanussot$^5$, and Ekaterina Merkurjev$^{1,2}$\footnote{Corresponding author,
	Email:  merkurje@msu.edu}  \\
    $^1$ Department of Mathematics, \\
Michigan State University, MI 48824, USA.\\
$^2$ Department of Computational Mathematics, Science and Engineering\\
Michigan State University, MI 48824, USA.\\
$^3$ Department of Mathematics, School of Data Science and Society\\
 The University of North Carolina at Chapel Hill, Chapel Hill, NC 27599, USA\\
$^4$ Computer, Computational and Statistical Sciences Division \\
Los Alamos National Laboratory, Los Alamos, NM 87545 USA\\
$^5$University of Grenoble Alpes, 38000 Grenoble, France}
\end{comment}

\maketitle

\begin{abstract}
Due to low spatial resolution,  hyperspectral data often consists of mixtures of contributions from multiple materials. This limitation motivates the task of hyperspectral unmixing (HU), a fundamental problem in hyperspectral imaging. HU aims to identify the spectral signatures (\textit{endmembers}) of the materials present in an observed scene, along with their relative proportions (\textit{fractional abundance}) in each pixel. A major challenge lies in the class variability in materials, which hinders accurate representation by a single spectral signature, as assumed in the conventional linear mixing model. 
%This situation can arise from factors such as variable illumination, atmospheric conditions, and the geometry and topography of the scene. Moreover, 
To address this issue, we propose using group sparsity after representing each material with a set of spectral signatures, known as endmember bundles, where each group corresponds to a specific material.
In particular, we develop a bundle-based framework that can enforce either inter-group sparsity or sparsity within and across groups (SWAG) on the abundance coefficients.  Furthermore, our framework offers the flexibility to incorporate a variety of sparsity-promoting penalties, among which the transformed $\ell_1$ (TL1) penalty is a novel regularization in the HU literature. Extensive experiments conducted on both synthetic and real hyperspectral data demonstrate the effectiveness and superiority of the proposed approaches.
\end{abstract}

\section{Introduction}
\IEEEPARstart{H}yperspectral imaging (HSI) is a technique that captures high-resolution spectral images that contain information in hundreds of narrow contiguous spectral bands, spanning from the visible to the near-infrared spectrum \citep{goetz2009three,meng2017,2014IEEE}. The high spectral resolution of these images enables better identification of different materials in the scene compared to traditional color images, as distinct materials typically exhibit unique spectral profiles, some of which are especially prominent in the infrared spectrum.  However, HSI suffers from relatively low spatial resolution, resulting in each observed measurement representing a combination of contributions from various materials in the scene. These so-called \textit{mixed-pixel} situations make hyperspectral data extremely challenging to analyze, which motivates the task of \textit{Hyperspectral Unmixing} (HU), or a source separation problem, where the objective is to recover the signature of pure materials (\textit{endmembers}) of the observed scene and to estimate the relative proportions (\textit{fractional abundance}) of various materials in each pixel of the observed image. 

A common approach to solve the HU problem is based on the Linear Mixing Model (LMM) \citep{bioucas2012hyperspectral, keshava2002spectral}, which assumes that the contributions of each endmember are linearly combined. Although LMM is the simplest and most widely used model---offering a computationally efficient framework---it has a major limitation: spectral variability of the endmembers. Spectral variability \citep{zare2013endmember,veganzones2014new, halimi2015unsupervised, halimi2016hyperspectral} refers to the phenomenon where the spectral signatures of pure constituent materials vary across different regions of a hyperspectral image, which is common in many real-world datasets. This variability can arise from factors such as changes in illumination, atmospheric conditions, and the geometry or topography of the scene. Furthermore, all materials exhibit some degree of intraclass variability, meaning they cannot be accurately represented by a single signature, as assumed in the conventional LMM.

%Several methodshave been proposed to address the problem of spectral variability in the literature. 
Broadly speaking, methods for addressing the spectral variability  
%\citep{zare2013endmember, hong2018augmented,borsoi2021spectral} \yl{[maybe better to mention [10,11] in the above paragraph, and no need for citation here]}
can be classified into two categories: bundle-based approaches and  model-based approaches. The main idea behind the bundle-based approach \citep{bateson2000endmember,somers2012automated,xu2015image,xu2018regional, uezato2019hyperspectral} is to use large libraries of spectra---also called $\textit{bundles}$---acquired as a prior. 
By regarding multiple spectral signatures for each pure material as a bundle or a group, one can further impose a group structure during the unmixing process. 
Some popular bundle construction methods include Vertex Component Analysis (VCA) \citep{nascimento2005vertex}, Orthogonal Subspace Projection (OSP) \citep{harsanyi1994hyperspectral}, N-FINDER \citep{winter1999n}, Simplex Volume Maximization (SiVM) \citep{heylen2011fully}, and the Minimum-Volume Enclosing Simplex method \citep{chan2009convex, craig1994minimum, drumetz2019hyperspectral}. 
In contrast, model-based approaches tackle spectral variability by incorporating additional variability terms within the model itself. Notable examples include the Perturbed Linear Mixing Model (PLMM) \citep{thouvenin2015hyperspectral}, Extended Linear Mixing Model (ELMM) \citep{drumetz2016blind}, Gaussian Mixture Models (GMM) \citep{zhou2018gaussian}, and Generalized Linear Mixing Model (GLMM) \citep{imbiriba2018generalized}, each of which introduces specific mechanisms to account for spectral variations.

This paper adopts a bundle-based approach and proposes a general framework specifically designed to address spectral variability. Leveraging prior work \citep{somers2012automated}, which enables automatic extraction of endmember bundles from hyperspectral data, we introduce a flexible framework that supports two types of group sparsity:  intra-group sparsity and sparsity within and across groups (SWAG). These sparsity-inducing strategies enforce structured sparsity on the expanded coefficient vectors associated with the endmember bundles. The main contributions of this paper are three-fold:
\begin{itemize}
    \item We propose a general framework that can incorporate a variety of sparsity-promoting penalties to enforce either inter-group sparsity or SWAG for bundle-based hyperspectral unmixing.
   % regularization to enforce inter-group sparsity for endmember bundles. In addition, we introduce a separate general regularization to enforce sparsity within and across groups (SWAG). These frameworks provide the flexibility to incorporate a variety of sparsity-promoting penalties into the unmixing problem.  
  
    \item We advocate the transformed $\ell_1$ (TL1) \citep{zhang2018minimization,nikolova2000local,zhang2014minimization} penalty as a novel regularization in the HU literature, when formulated under either inter-group sparsity or SWAG.
    \item Extensive experiments demonstrate empirically that SWAG is the most effective approach for addressing the spectral variability.
\end{itemize}

The remainder of the paper is organized as follows. Section \ref{literature_review} reviews bundle-based LMM and group sparsity. 
%Since sparsity is an important part of our work, in \ref{group_sparsity}, we review some group sparsity work in the literature. 
In  Section \ref{proposed_abundance}, we describe the proposed framework that can enforce group sparsity under the endmember bundles. Section \ref{experiments_and_results} details the experimental results. Finally, Section \ref{conclusion} provides concluding remarks.

%\gokul{Note: Scalars are denoted by lowercase letters in regular typeface: $ n,k$. Vectors are denoted by lowercase bold letters: $\bm{x}, \bm{y}$. Matrices are denoted by uppercase bold letters: $\bm{X}, \bm{B}$.}

%\yl{[Rough outline for intro]}

%{\tt what's hyperspectral imaging? Advantages and need for unmixing}

%{\tt Review some unmixing methods}

%{\tt Point out the spectral variability issue and say our contributions}

\section{Background}
\label{literature_review}

In this section, we begin by summarizing all notation in Table \ref{tab:notations}, followed by a literature review in Sections \ref{bundle-based-llm} and \ref{group_sparsity}.
%\yl{[each endmember may have different number of signatures?]}

 \begin{table}[!ht]
	\centering
	{
		\caption{Notation table}
		\label{tab:notations}
		\vspace{0.05cm}
		\begin{tabular}{lc}
         \hline
        Variable & Description\\
        \hline
		
			$n$ &  Number of pixels \\
                $w$ &  Number of spectral bands \\
                $k$ & Number of endmembers \\
                $m_l$ &  Number of signatures for the $l$-th endmember  \\
                $r$ &  Total number of signatures \\
                $\bm{X} \in \mathbb{R}^{w \times n}$& Hyperspectral image\\
                $\bm{M}  \in \mathbb{R}^{k \times n} $ & Abundance map \\
                 $\bm{S} \in \mathbb{R}^{w \times k}$ & Mixing matrix \\
                $\bm{A}  \in \mathbb{R}^{r \times n} $ & Extended (abundance) matrix\\
                $\bm{B}  \in \mathbb{R}^{w \times r}$ & (Extended) bundle matrix\\
                $\bm{0}_{m \times n}$ & Zero matrix of size $m \times n$. \\

			\hline
		\end{tabular}

    }
    \end{table}

\subsection{Bundle-Based Linear Mixing Model}
\label{bundle-based-llm}

Let $\bm{X} \in \mathbb{R}^{w \times n}$ be a hyperspectral image where $w$ is the number of spectral bands and $n$ is the number of pixels in the image. Suppose $k$ pure materials, called endmembers, are in the scene, each with its spectral signature denoted as $\{\h s_l\}_{l=1}^k.$ The LMM \citep{bioucas2012hyperspectral, keshava2002spectral} assumes that the spectral measurement at each pixel $i$ is a linear combination of endmember spectral signatures, i.e.,
\begin{equation}\label{eq:LMM}
    \h x_i = \sum_{l=1}^k m_{li} \; \h s_l  + \h e_i \: , \quad i = 1,\dots,n \, ,
\end{equation}
where $\h x_i$ is the $i$-th column of $\bm{X}$, $m_{li}$ is the proportion of the $l$-th endmember at the $i$-th pixel, and $\h e_i$ is assumed to be an additive noise following a Gaussian distribution. Due to low spatial resolution, the fractional abundance $m_{li}$ represents the fractional area occupied by the $l$-th endmember in the $i$-th pixel, and consequently, fractional abundances are subject to the following constraints: Abundance Nonnegativity Constraint (ANC), i.e.,  $m_{li}\geq0 \quad \forall (l,i)$, and the Abundance Sum-to-one Constraint (ASC), i.e., $\sum_{l=1}^k m_{li}=1, \; \forall i$. Concatenating all the spectral signatures $\h s_l$ into a {\it mixing matrix} $\bm{S} \in \mathbb{R}^{w \times k}$ and assembling all weights $m_{li}$ into a matrix $\bm{M} \in \mathbb{R}^{k \times n}$, referred to as an {\it abundance map}, the standard LMM with additive noise can be expressed as 
\begin{equation}
    \bm{X} = \bm{S M} + \bm{E} \, , 
\end{equation}
where $\bm{E} \in \mathbb{R}^{w \times n}$ is composed of the noise terms, with each column associated with a pixel $i$ in the noise term $\h e_i$.  When $\bm{S}$ is known \textit{a priori}, identifying the abundance map $\bm M$ from the hyperspectral data matrix
$\bm X$ is referred to as the non-blind hyperspectral unmixing, which is often formulated as the following regularized optimization problem,
\begin{equation}
\label{R-LLM}
    \min_{\bm{M} \in \Delta_{k \times n}} \frac{1}{2}\| \bm{X} - \bm{S M} \|_F^2 + \lambda R(\bm{M}) \, ,
\end{equation}
where $R(\bm{M})$ is a regularization term,  $\lambda>0$ is a balancing parameter, and $\Delta_{k \times n}$ denotes a constraint set of all $k \times n$  matrices with ANC and ASC,  i.e.
\begin{equation}
    \label{eq:ANC-ASC}
    \begin{aligned}
  %  \small
    \Delta_{k \times n} := &\{ \Lambda \in \mathbb{R}^{k \times n} \mid \Lambda_{li} \geq 0, \bm{1}_k^\top \Lambda = \bm{1}_n^\top, \\
    & l = 1,  \dots k, \ i = 1, \dots n\} \, . 
\end{aligned}
\end{equation}

One popular approach to addressing spectral variability is the use of spectral \textit{bundles}, which can be easily incorporated into the LMM framework by replacing the mixing matrix $\bm S$ by an extended bundle matrix. Specifically, we assume that there are $m_{l}$ elements available for the $l$-th pure material with its endmembers organized in a matrix form, i.e.,
\begin{equation}
\bm{B}^{(l)} = 
    \begin{bmatrix}
b_{1,1}^{(l)} & \cdots & b_{1,m_{l}}^{(l)}  \\
\vdots &  & \vdots  \\
b_{w,1}^{(l)} & \cdots & b_{w,m_{l}}^{(l)}  \\
%0 & \cdots & 0 & 1 & \cdots & 1 & 0 & \cdots & 0\\
%\vdots &  & \vdots & \vdots &  & \vdots & \vdots & &\vdots  \\
%0 & \cdots &  0 & \cdots &  1 & \cdots & 1\\
\end{bmatrix}\in\mathbb R^{w\times m_l},
\end{equation}
where each column of $\bm{B}^{(l)}$ corresponds to one spectral representation.  We can then concatenate these matrices into a bundle matrix, denoted by,
\begin{equation}
\label{group_structure_B}
\bm{B} = 
    \begin{bmatrix}
\begin{array}{c|c|c|c} 
  \bm{B}^{(1)} & \bm{B}^{(2)} & \cdots &\bm{B}^{(k)} \\
  \end{array} 
\end{bmatrix} \in \mathbb R^{w\times r},
\end{equation}
with $r:=m_{1}+\cdots +m_{k}$. Denote $\mathcal G_l$ as the index set corresponding to the $l$-th group, and we refer $\mathcal G=[\mathcal G_1, \mathcal G_2, \cdots, \mathcal G_k]$ as a group structure. Consequently, the extended (abundance) matrix, denoted by $\bm A,$ shall have the same group structure as $\mathcal G$, i.e., every column of $\bm A$ can be represented by 
\begin{equation}
\begin{split}
    \label{group_structure_a}
    \bm{a} = & \Bigl[\underbrace{a_{1}^{(1)}, a_{2}^{(1)}, \cdots, a_{m_{1}}^{(1)}}_{ \bm{a}^{\mathcal{G}_1}}, \underbrace{a_{1}^{(2)}, a_{2}^{(2)}, \cdots, a_{m_{2}}^{(2)}}_{ \bm{a}^{\mathcal{G}_2}}, \\
    & \cdots, \underbrace{a_{1}^{(k)}, a_{2}^{(k)}, \cdots, a_{m_{k}}^{(k)} }_{ \bm{a}^{\mathcal{G}_{k}}} \Bigr]^\top \in\mathbb R^r,
    \end{split}
\end{equation}
where $ {\bm{a}^{\mathcal{G}_l}}\in\mathbb R^{m_l}$ represents the $l$-th group of $\bm{a}.$ % and each group $\mathcal G_l$ has $m_{l}$ elements.

With such a bundle matrix $\bm{B}$ and the corresponding extended matrix $\bm{A}$, the regularized unmixing \eqref{R-LLM} becomes
\begin{equation}\label{eq:general-bundle}
 \widehat A:=   \argmin_{\bm{A} \in \Delta_{r \times n}} \frac{1}{2}\|\bm{X}-\bm{BA}\|_F^2 + \lambda R(\bm{A}).
\end{equation}
In this work, we assume the bundle matrix $\bm B$ is given; please refer to Section \ref{sec:bundle_construction} for a detailed discussion of bundle construction \citep{plaza2004quantitative,nascimento2005vertex,harsanyi1994hyperspectral}. Instead, we focus on the estimation of the matrix $\bm A$ as a sparse recovery problem \citep{iordache2013collaborative,drumetz2019hyperspectral,iordache2011hyperspectral,borsoi2018fast,iordache2012total,qian2011hyperspectral}. In particular, the goal is to identify a small set of spectral signatures from the library that best represents each observed pixel by enforcing sparsity and/or group sparsity of the abundance map together with the ANC and ASC constraints.

Once the optimal solution $\widehat A$ is obtained from \eqref{eq:general-bundle}, we can find the global abundance \citep{drumetz2019hyperspectral} of the \( l \)-th material for pixel $i$ as follows. First, we calculate the total abundance coefficient for \( l \) as:  
\begin{equation}
    \label{eqn:global_abundance}
\widehat{m}_{l,i} = \sum_{j=1}^{m_{l}} \widehat{a}_{j,i}^{(l)},
\end{equation}
where \( \widehat{a}_{j,i}^{(l)} \) represents the coefficient of the \( j \)-th component within group \( \mathcal{G}_l \). Then the corresponding spectral signature of material $l$ is computed as: 
%\yl{[should we put hat on s?]}
\begin{equation} 
\label{global_endmember}
\widehat{s}_{l,i} = \frac{\sum_{j=1}^{m_{\mathcal{G}_l}} \widehat{a}_{j,i}^{(l)} b_{:,j}^{(l)}}{\sum_{j=1}^{m_{\mathcal{G}_l}} \widehat{a}_{j,i}^{(l)}} \in\mathbb R^w, 
\end{equation}
where \( b_{:,j}^{(l)} \) is the \( j \)-th column of the block matrix \( \bm{B}^{\mathcal{G}_l} \).  
Note that \( \widehat{\h s}_l \) represents the weighted mean of all the available instances of material \( l \), ensuring that the global abundance satisfies both the non-negativity and sum-to-one constraints. 
The expression for a single pixel in terms of the ``global'' abundance of the $l$-th material is given by:
\begin{equation}
    \label{gloobal_abundance}
   \h x_i = \sum_{l=1}^k \bigg(  \sum_{j=1}^{m_{\mathcal{G}_l}} \widehat{a}_{j,i}^{(l)}\bigg) \bigg(\frac{\sum_{j=1}^{m_{\mathcal{G}_l}} \widehat{a}_{j,i}^{(l)} b_{:,j}^{\mathcal{G}_l}}{\sum_{j=1}^{m_{\mathcal{G}_l}} \widehat{a}_{j,i}^{(l)}}\bigg). 
\end{equation}

\subsection{Group Sparsity}
\label{group_sparsity}

%\yl{[check group sparse or group sparsity or group-sparse]}

Group sparsity is an active research area with applications spanning multiple domains, including dynamic MRI \citep{usman2011k}, DNA microarrays \citep{parvaresh2008recovering}, color imaging \citep{majumdar2010compressed}, source localization \citep{malioutov2005sparse}, etc.  We shall specify three types of group sparsity. 
\begin{itemize}
    \item \textit{Intra-group sparsity} refers to sparsity \textit{within} individual groups, which implies that only a few elements within each group are nonzero. 
    \item \textit{Inter-group sparsity} or \textit{group-level sparsity} refers to sparsity \textit{across} groups, meaning that only a few groups contain nonzero elements, while most groups are entirely zero. 
    \item \textit{Sparsity within and across groups} (SWAG) \citep{bayram2017penalty} indicates only a few groups are nonzero (inter-group sparsity), and within each selected group, only a few elements are nonzero (intra-group sparsity). 
\end{itemize}

%Lastly, 
%\yl{[Go through the paper, add ``-'' between inter/intra and group]} \gokul{done.}

%\yl{[merge with this section]}
 %The $\ell_1$ norm is a widely used regularization to promote sparsity, which is also known as the least absolute shrinkage and selection operator (LASSO) \citep{tibshirani1996regression}. However, it has been observed in \citep{fan2001variable} that LASSO introduces bias towards coefficients with larger magnitude. To reduce the bias, various nonconvex regularization penalties have been proposed, including smoothly clipped absolute deviation (SCAD) \citep{fan2001variable}, minimax concave penalty (MCP) \citep{zhang2010nearly},  $\ell_{\frac{1}{2}}$ penalty \citep{lai2013improved, xu2012l_half},  logarithm penalty \citep{candes2008enhancing},   transformed $\ell_1$ (TL1) \citep{nikolova2000local}, $\ell_1 -\ell_2$ \citep{esser2013method, yin2014ratio, lou2015computing, lou2018fast}, $\ell_1/\ell_2$ \citep{peng2016polynomial,rahimi2019scale, wang2020accelerated}, and error function (ERF) \citep{guo2021novel}.

We review some regularization-based methods to enforce group sparsity. 
For a vector $\bm{a} \in \mathbb{R}^r$ with a group structure in (\ref{group_structure_a}), the mixed norm or the $\ell_{(p,q)}$ regularization \citep{kowalski2009sparsity, drumetz2019hyperspectral} is commonly used, which is defined by,
$$
\|\bm{a}\|_{\mathcal{G},p,q} := \bigg(\sum_{l=1}^k \|a^{\mathcal{G}_l}\|_{p}^q \bigg)^{1/q},
$$
where $\|\bm a\|_p=(\sum_{i=1}^n |a_i|^p)^{1/p}$ for $\bm a\in\mathbb R^n.$
%as a measure of the sparsity of $\bm{a}$. 
The $\ell_{(p,q)}$ regularization considers the $\ell_{p}$ regularization within each group and the $\ell_{q}$ regularization among groups. Depending on the values of $p,q,$ the $\ell_{(p,q)}$ regularization can enforce one of the three aforementioned types of group sparsity. For example, $p=2$ is often used for inter-group sparsity, which uses each group's $\ell_2$ norm to encourage the entire group to be zero while allowing nonzero values within a selected group. Specifically,   the $\ell_{(2,1)}$ norm is referred to as group LASSO \citep{yuan2006model}, which promotes inter-group sparsity by encouraging only a small number of groups to be active ($q=1$), while favoring dense mixtures within each group ($p=2$). 
Other regularization techniques include collaborative sparse regularization \citep{iordache2013collaborative}, which applies the mixed \( \ell_{2,1} \) norm to the entire abundance matrix. Moreover, various nonconvex regularizers have been proposed to promote inter-group sparsity,  including group capped-$\ell_1$ \citep{pan2021group,phan2019group}, group MCP \citep{huang2012selective}, group SCAD \citep{wang2007group}, $\ell_{(2,0)}$ \citep{shi2018collaborative}, and group LOG \citep{ke2021iteratively}, with implicitly setting $p=2$.
%These methods aim to recover the desired group sparsity patterns effectively. \yl{[I assume sparsity in this paragraph is not in hyperspectral]} \gokul{yes}
On the other hand, the elitist LASSO penalty \citep{kowalski2009sparsity}, denoted by $\ell_{(1,2)}$, has been proposed to promote intra-group sparsity by encouraging a larger number of groups to be active while favoring a small number of variables to be active within each group. Alternatively, the $\ell_{0,2}$ penalty \citep{jiao2016group} directly enforces intra-group sparsity by using the $\ell_0$ regularization on each group. 

% In some applications, it is advantageous to promote sparsity within and across groups (SWAG) \citep{bayram2017penalty}. The SWAG penalty favors solutions where the number of active variables in each group is smaller, but the number of active groups is larger. 

Group sparsity is widely used in the hyperspectral unmixing problem. Iordache, et al.~\citep{iordache2011hyperspectral} considered the group LASSO for the unmixing problem.  Both inter-group and intra-group sparsity have been discussed in \citep{esser2013method} with the use of $\ell_1-\ell_2$ and $\ell_1/\ell_2$ regularizations. 
Furthermore, Drumetz et al.~\citep{drumetz2019hyperspectral} introduced mixed norms, including group LASSO, elitist LASSO, and fractional penalties, to promote inter-group sparsity,  intra-group sparsity, and SWAG, respectively. Their experimental results show that the fractional penalty, which enforces both inter-group and intra-group sparsity significantly improves performance compared to other penalties. Cui et al.~\citep{cui2023unrolling} suggested unrolling the non-negative matrix factorization (NMF) method via %group sparsity. In particular, a spatial group sparsity regularizer is taken into account for the abundance estimation to promote
intra-group sparsity.  A regularized NMF with 
spatial group sparsity was proposed by
 Wang et al.~\citep{wang2017spatial}. Ren et al.~\citep{ren2021nonconvex} presented a nonconvex framework for SU that incorporates the group structure of the spectral library. Ayres et al. \citep{ayres2024generalized} extended 
a multiscale spatial regularization approach 
to a bundle-based formulation by incorporating the group sparsity-inducing mixed norms.
%\yl{[so only three papers on group-sparsity for HU? you may want to double check the literature]}

%Moreover, \citep{iordache2011hyperspectral} propose spatial group sparsity regularized nonnegative matrix factorization for hyperspectral unmixing. Furthermore, the authors in \citep{esser2013method} propose hyperspectral unmixing using a Gaussian mixture model combined with spatial group sparsity. 

\section{Proposed Abundance Estimation Methods}
\label{proposed_abundance}

We introduce a general regularization to enforce group sparsity, given by 
%\yl{[I changed the subscript right after R; check the rest of the paper. In SWAG, you don't have the square root factor, make both consistent.]}
%\begin{equation}
%\label{normp}
   % R_{\mathcal{G},p,f}(\bm{a}) = \sum_{l=1}^k \sqrt{m_{l}}f(\|\bm{a}_{\mathcal{G}_l}\|_p),
%\end{equation}
\begin{equation}
\label{normp}
    R_{\mathcal{G},p,f}(\bm{a}) = \sum_{l=1}^k f(\|\bm{a}^{\mathcal{G}_l}\|_p),
\end{equation}
where $\mathcal G$ denotes the group structure defined in \eqref{group_structure_a} with the $l$-th group having $m_l$ elements, $p$ indicates the $\ell_p$ norm used within each group, referred to as the group norm, and $f(\cdot)$ is a univariate function. 
%The factor of $\sqrt{m_l}$ in front of $f(\cdot)$ evaluated at the $l$-th group norm serves as a normalizing weight in such a way that the weighted sum in \eqref{normp}  approximates the number of non-zero groups.
Depending on the choice of $p$ and $f(\cdot),$ the proposed regularization $R_{\mathcal{G} , p, f}(\cdot)$ is applicable to any of the three aforementioned types of group sparsity. For example, if $p=2$ and $f$ is the absolute value function, then  $R_{\mathcal{G},p,f}$ becomes the group LASSO. We include the $\ell_q$ regularization for $q<1$ \citep{drumetz2019hyperspectral} and transformed $\ell_1$ (TL1) \citep{zhang2018minimization, nikolova2000local} as alternatives of $f(\cdot)$ in experiments. 
%\yl{[we may have to remove L1-L2 in experiments, as it's not separable.]}\gokul{done}

We incorporate the regularization \eqref{normp} into a bundle-based hyperspectral unmixing framework in \eqref{eq:general-bundle}, thus leading to 
\begin{equation}
\label{eqn:general_framework}
    \argmin_{\bm{A}\in \Delta_{r \times n}} \frac{1}{2} \|\bm{X}- \bm{BA}\|_F^2 + \lambda \sum_{i=1}^n R_{\mathcal{G} , p, f}(\bm{a}_i) ,%+ \mathcal{I}_{} (\bm{A}),
    \end{equation}
where the group sparsity penalty $R_{\mathcal{G},p,f}(\cdot)$ is applied to every column of the matrix $\bm A$, i.e., $\bm A = [\bm a_1, \bm a_2, \cdots, \bm a_n],$ the constraint set $\Delta_{r \times n}$ is defined in \eqref{eq:ANC-ASC} with $r$ vertices in a unit simplex in the bundle setting, and $\lambda>0$ is a balancing parameter (to be tuned).
%\yl{[isn't $\Delta_r$ defined in \eqref{eq:ANC-ASC}? if so make same notation]} \gokul{done}
%\yl{[you only have ANC for (7) and unit simplex is a combined constraint of ANC and ASC, which you use $\Omega$; make it consistent]} \gokul{done}.

We elaborate on the inter-group sparsity by setting $p=2$ along with its numerical algorithm in Section \ref{sect:inter}, while SWAG with $p=1$ is discussed in Section \ref{sect:SWAG}.

\subsection{Inter-Group Sparsity for Endmember Bundles} \label{sect:inter}

Given an abundance vector $\bm{a} \in \mathbb{R}^r$ with a predefined group structure $\mathcal{\mathcal{G}}$ that  consists of $k$ groups, as described in Section \ref{bundle-based-llm}, we set $p=2$ in \eqref{normp} to measure the group norm, ensuring that all elements within a group are either all active (non-zero) or discarded (zero), thereby preserving inter-group sparsity.

To simplify the notation in the algorithmic description, we denote 
\begin{equation}
\label{TL1_penalty}
    R_{\text{inter}}(\bm{A}) = \sum_{i=1}^n R_{\mathcal{G} , 2, f}(\bm{a}_i),
\end{equation}
as a regularization applied onto the matrix $\bm A\in \mathbb R^{r\times n}$ with $i$-th column $\bm a_i$ for $i=1,\cdots, n.$ To deal with the constraint in \eqref{eqn:general_framework}, we introduce the indicator function of the unit simplex, ensuring that each column of $\bm A$ satisfies both the ANC and ASC constraints. Specifically, for a vector $\bm{a} \in \mathbb{R}^r$, we define the indicator function $\mathcal{I}_{\Delta_r}$ as: 
\begin{equation}
\label{indicator_function}
\mathcal{I}_{\Delta_r}(\bm{a}) =
    \begin{cases}
        0, & \text{if } \bm{a} \in \Delta_{r}\\
        + \infty,  & \text{otherwise},
        \end{cases}
 \end{equation}
which can be extended to a matrix $\bm{A} \in \mathbb{R}^{r \times n}$ with $n$ pixels as:
 $\mathcal{I}_{\Delta_{r \times n}} (\bm{A}) = \sum_{i=1}^n \mathcal{I}_{\Delta_r}(\bm{a}_i). $
 %where $\bm{a}_i$ is the \( i \)-th column of $\bm{A}$. 
Therefore, we can express the optimization problem \eqref{eqn:general_framework} with the specific regularization \eqref{TL1_penalty} as follows,
\begin{equation}
\label{problem1_beforeADMM}
   % \begin{aligned}
    \argmin_{\bm{A}}  \frac{1}{2} \|\bm{X} -\bm{BA}\|_{F}^2 + \lambda R_{\text{inter}}(\bm{A})  + \mathcal{I}_{\Delta_{r\times n}} (\bm{A}).
    %\end{aligned}
\end{equation}

To find the optimal solution of \eqref{problem1_beforeADMM}, we adopt the Alternating Direction Method of Multipliers (ADMM) \citep{boyd2011distributed} due to its flexibility and efficiency. To this end, we introduce two auxiliary split variables $\bm{U}$ and $\bm{V}$, rewriting the problem (\ref{problem1_beforeADMM}) equivalently as: 
%\yl{[I'd suggest swap A and U, so that the R function is applied on A]}\gokul{done}
\begin{equation}
\label{problem1}
    \begin{aligned}
    \argmin_{\bm{A},\bm{U},\bm{V} } & \frac{1}{2} \|\bm{X} -\bm{BU}\|_{F}^2 + \lambda R_{\text{inter}}(\bm{A})  + \mathcal{I}_{\Delta_{r\times n}} (\bm{V}) \\
    \text{s.t.} & \ \bm{U} = \bm{A}, \ \bm{U} = \bm{V}.
    \end{aligned}
\end{equation}
The corresponding augmented Lagrangian function is expressed by
%\ekaterina{Why do you have U,V;U,V repeated in L(A,U,V;U,V)} \gokul{done}
\begin{equation}
 \label{AL}
   \begin{aligned}
     &\mathcal{L}_\rho(\bm{A},\bm{U},\bm{V}; \bm{C},\bm{D}) \\
     = & \frac{1}{2} \|\bm{X} -\bm{BU}\|_{F}^2 + \lambda R_{\text{inter}}(\bm{A})  + \mathcal{I}_{\Delta_{r \times n}} (\bm{V}) \\
      & \quad + \frac{\rho}{2} \|\bm{U}- \bm A- \bm C\|_F^2 + \frac{\rho}{2} \|\bm{U}- \bm V- \bm D\|_F^2,
     \end{aligned}
 \end{equation}
 where $\bm{C}$ and $\bm{D}$ are two dual variables and $\rho$ is a positive constant. ADMM iterates as follows:
 \begin{equation}
    \label{ADMM1}
   \left\{ \begin{aligned}
    \bm{U}^{\tau+1} = &  \argmin_{\bm{U}} \mathcal{L}_\rho(\bm{A}^{\tau},\bm{U},\bm{V}^{\tau}; \bm{C}^{\tau},\bm{D}^{\tau}),  \\
    \bm{A}^{\tau+1} =& \argmin_{\bm{A}}  \mathcal{L}_\rho(\bm{A},\bm{U}^{\tau+1},\bm{V}^{\tau}; \bm{C}^{\tau},\bm{D}^{\tau}), \\
    \bm{V}^{\tau+1} = & \argmin_{\bm{V} }\mathcal{L}_\rho(\bm{A}^{\tau+1},\bm{U}^{\tau+1},\bm{V}; \bm{C}^{\tau},\bm{D}^{\tau}),  \\
    \bm{C}^{\tau+1} =& \bm{C}^{\tau} +\bm A^{\tau+1}- \bm U^{\tau+1},\\
    \bm{D}^{\tau+1} = & \bm{D}^{\tau} + \bm V^{\tau+1}-\bm U^{\tau+1},
\end{aligned}\right.
\end{equation}
%\ekaterina{Why do you have $U^T,V^T;U^T,V^T$ repeated} \gokul{done.}
where $\tau$ counts the iteration number.  We detail the update rules for each subproblem. 
%\yl{[double check dual updates; I made the change in (18); if it's correct, update Algorithm 1. Also update your python codes to align with these variables when sharing codes on github]}\gokul{done}

%Next, we will introduce the proposed penalties and the update rules for the subproblems.  %\yl{[move the closed-form solutions of A and V here; then how to solve the U-step depends on which type of sparsity]}

% formulate a generalized group regularizer as follows:
% \begin{equation}
% \label{norm1}
%     F(\bm{a})_{\mathcal{G},2,f} = \sum_{l=1}^k \sqrt{m_{\mathcal{G}_l}}f(\|\bm{a}_{\mathcal{G}_l}\|_2),
% \end{equation}
% where $f: \mathbb{R} \rightarrow \mathbb{R}$ denotes a sparsity promoting function   such as transformed $\ell_1$ \citep{zhang2018minimization, nikolova2000local}. This regularizer enforces sparsity on a vector whose entries are the \(\|\bm{a}_{\mathcal{G}_l}\|_2\). When one of the entries in this vector is zero, the entire corresponding group is discarded, effectively promoting inter-group sparsity. The use of 
%  \yl{[Add a sentence to justify the use of L2 norm in (9) for inter-group spasrity]} \gokul{done} 

% The vector formulation \eqref{norm1} can be easily extended to a matrix 
%$\bm{A} \in \mathbb{R}^{r \times n}$ by applying it column-wise and summing the results across all pixels, i.e.,

%With the regularization function $R(\bm{A})_{\mathcal{G},2,f}$ in (\ref{TL1_penalty}) in the general framework (\ref{eqn:general_framework}), we solve each subproblem (\ref{ADMM1}) as follows:
 The $\bm{U}$-subproblem in \eqref{ADMM1} is equivalent to 
 \[
\arg \min_{\bm{U}}\frac{1}{2} \|\bm{X}-\bm{BU}\|_F^2 +  \frac{\rho}{2} \|\bm{U}-\bm A^{\tau}-\bm C^{\tau}\|_F^2 + \frac{\rho}{2} \|\bm{U}-\bm V^{\tau}-\bm D^{\tau}\|_F^2 ,
  \]
 which has a closed-form solution: %since it only involves quadratic terms. 
\begin{equation}
\begin{split}
    \label{eq:Inter_Uupdate}
    \bm{U}^{\tau+1} = &(\bm{B}^T \bm{B} + 2 \rho \bm{I}_r)^{-1} \\ &\qquad\bigg(\bm{B}^T \bm{X} + \rho(\bm{A}^{\tau}+\bm{V}^{\tau}+\bm{C}^{\tau}+\bm{D}^{\tau})\bigg),
    \end{split}
\end{equation}
 where $\bm{I}_r$ denotes the $r \times r$ identity matrix.

With the definition of $R_{\text{inter}}$ in \eqref{TL1_penalty}, the $\bm{A}$-subproblem in \eqref{ADMM1} can be written in terms of a separable sum, i.e., % \yl{[express R(U) by sum of column and change U to A]}
%\[
%\lambda R(\bm{A})_{\mathcal{G}, 2 ,f} +  \frac{\rho}{2} \|\bm{A}-\bm{U}-%\bm{C}\|_F^2.
%\]
\begin{equation}\label{eq:A-update}
\arg  \min_{\bm A} \sum_{i=1}^n \left(\lambda R_{\mathcal{G},2,f}(\bm{a}_i)+ \frac{\rho}{2}\|\bm{u}_i-\bm{a}_i-\bm{c}_i\|_2^2\right),
\end{equation}
where $\bm u_i$ and $\bm c_i$ denote the $i$-th columns of the matrices $\bm U, \bm C,$ respectively.
The minimization formulation \eqref{eq:A-update} suggests that each column $\bm{a}_i \in \mathbb{R}^r$ of a matrix $\bm{A} \in \mathbb{R}^{r \times n}$ can be minimized with a closed-form solution given by a proximal operator \citep{parikh2014proximal}. Specifically for the $i$-th column $\bm a_i$, the minimization problem \eqref{eq:A-update} can be expressed as
\begin{equation}\label{eq:A-update2}
\argmin_{\bm a}\sum_{l=1}^k \left(\lambda f(\|\bm{a}^{\mathcal{G}_l}\|_p) + \frac{\rho}{2}\|\bm{u}_i^{\mathcal{G}_l}-\bm{a}^{\mathcal{G}_l}-\bm{c}_i^{\mathcal{G}_l}\|_2^2\right),
\end{equation}
enabling the following update for each group  $l \in \{1,2, \cdots, k\}$:
\begin{equation}\label{eq:columnAupdate}
    \bm{a}_i^{\mathcal{G}_l} = \frac{\bm{u}_i^{\mathcal{G}_l}-\bm{c}_i^{\mathcal{G}_l}}{\|\bm{u}_i^{\mathcal{G}_l}-\bm{c}_i^{\mathcal{G}_l}\|_2} \prox_{f}\bigg(\|\bm{u}_i^{\mathcal{G}_l}-\bm{c}_i^{\mathcal{G}_l}\|_2,\frac{\lambda }{\rho}\bigg),
    \end{equation} 
where $\prox$ denotes the proximal operator of the function $f$; please refer to Appendix \ref{sect:appendB} for formulas of inter-group sparsity.

Lastly, the $\bm{V}$-subproblem is given by
 \[
\arg \min_{\bm{V}}\mathcal{I}_{\Delta_{r \times n}} (\bm{V}) + \frac{\rho}{2} \|\bm{U}-\bm{V}-\bm{D}\|_F^2, \]
which can be solved by a simple projection:
 \begin{equation}
     \bm{V} = \proj_{{\Delta}_{r \times n}} (\bm{U} -
     \bm{D}),
 \end{equation}
 where $\proj_{\Delta}$ is the projection operator onto the set $\Delta$ which can be implemented efficiently by a fast algorithm in \citep{wang2013projection}.

For initialization, we follow the work \citep{drumetz2019hyperspectral}, which applied the Fully Constrained Linear Spectral Unmixing (FCLS) \cite{heinz2001fully}, originally designed for the abundance map $\bm M$ with a least-squares problem under the ANC and ASC constraints, to the bundle matrix $\bm A$. 
%via a modified  \citep{drumetz2019hyperspectral}.  The FCLSU method solves a linear least squares problem with nonnegativity constraints to estimate the abundances.  \yl{[FCLS is without bundle; how to adapt to our bundle setting, more details?]} \gokul{FCLSU in 16 is adapted for bundle setting.} 
 We present the overall algorithm for minimizing (\ref{problem1}) in Algorithm \ref{alg:ADMM1}.

%\yl{[make X and B as \textbf{input}; then \textbf{Choose} lambda and T; \textbf{initialize} A =V, C=D=0. usually initial for C/D is zero, how did you initialize A ? ]} \gokul{done}
 
\vspace{0.2cm}
\SetKwComment{Comment}{/* }{ */}

\begin{algorithm}
\caption{Enforcing inter-group sparisty via ADMM }
\label{alg:ADMM1}
{\textbf{Input}: hyperspectral data $\bm{X}$ and bundle matrix $\bm{B}$}\\
{\textbf{Choose}: $\lambda>0, \text{maximum iteration } T $}\\
\textbf{Initialize}: $\bm{A}^0$, and $\bm{V}^0$ are obtained by FCLSU \citep{drumetz2019hyperspectral}, $\bm{C}^0 =\bm{D}^0 =\bm{0}_{r \times n}$, $\tau =0$ ;\\
\While{$\tau \leq T$ }{

\vspace{0.1cm}
 Update $\bm{U}^{\tau+1}$ via \eqref{eq:Inter_Uupdate}\;

\vspace{0.1cm}
 Update $\bm{A}^{\tau+1}$ columnwise via \eqref{eq:columnAupdate} 
 %\frac{\bm{U}-\bm{C}}{\|\bm{U}-\bm{C}\|_2} \bm{\prox}_{(\lambda/ \rho),\mathcal{G},f}(\|\bm{U} -\bm{C}\|_F)$\; \yl{[may need to change notation]}
 
 \vspace{0.1cm}
 $\bm{V}^{\tau+1} = \bm{\proj}_{{\Delta}_{r \times n}} (\bm{U}^{\tau+1} - \bm{D}^{\tau})$\;
 
\vspace{0.1cm}
 $\bm{C}^{\tau+1} = \bm{C}^{\tau}+\bm{A}^{\tau+1}-\bm{U}^{\tau+1}$\;

 \vspace{0.1cm}

   $\bm{D}^{\tau+1} = \bm{D}^{\tau}+\bm{V}^{\tau+1}-\bm{U}^{\tau+1}$\;

   \vspace{0.1cm}
 
  $ \tau = \tau + 1 $\;

     }
     {\textbf{Output}: $\widehat{\bm{A}} = \bm A^{\tau}$, $\widehat{\bm M}, \widehat{\bm S}$ via \eqref{eqn:global_abundance} and \eqref{global_endmember}, respectively.}
     %\KwResult{$\bm{A}$ \yl{[can we say output?]}} 
     
\end{algorithm}

\subsection{SWAG  for Endmember Bundles}
\label{sect:SWAG}
%\subsection{Intra Group Sparsity for Endmember Bundles}

To promote sparsity within and across groups (SWAG), we propose using the $\ell_1$ norm for within-group sparsity and the non-convex univariate function $f(\cdot)$ to promote sparsity across groups. We define the SWAG reguarlization by
% For an abundance vector $\bm{a} \in \mathbb{R}^r$ with a predefined group structure $\mathcal{G}$ consisting of $k$ groups and a sparsity promoting function  $f: \mathbb{R} \rightarrow \mathbb{R}$,  we define a regularizer as follows:
% \begin{equation}
% \label{norm2}
%     F(\bm{a})_{\mathcal{G},1,f} = \sum_{l=1}^k f(\|\bm{a}_{\mathcal{G}_l}\|_1).
% \end{equation}
% This can easily be extended to a matrix 
% $\bm{A} \in \mathbb{R}^{r \times n}$ by applying it columnwise and summing the results across all pixels:
\begin{equation}
\label{SWAG_penalty}
    R_{\text{SWAG}}(\bm{A}) = \sum_{i=1}^n R_{\mathcal{G} , 1, f}(\bm{a}_i),
\end{equation}
which is analogous to $R_{\text{inter}}$ in \eqref{TL1_penalty} but with $p=1.$
 %where $F(\bm{a}_i)_{\mathcal{G},1,f}$ represents the value for pixel $i$. The regularization function $R(\bm{A})_{\mathcal{G},1,f}$ as defined in (\ref{SWAG_penalty}) within the general framework (\ref{eqn:general_framework}) is designed to promote sparsity within and across groups. 

 Due to the ANC for the matrix $\bm A,$ the $\ell_1$ norm of each column reduces to the summation, and hence we introduce the following matrix 
\begin{equation}
\bm{Z} = 
    \begin{bmatrix}
1 & \cdots & 1 & 0 & \cdots & 0 & 0 & \cdots & 0\\
0 & \cdots & 0 & 1 & \cdots & 1 & 0 & \cdots & 0\\
\vdots &  & \vdots & \vdots &  & \vdots & \vdots & &\vdots  \\
0 & \cdots & 0 & 0 & \cdots & 0 & 1 & \cdots & 1\\
\end{bmatrix} \in \mathbb{R}^{k \times r},
\end{equation}
whose $l$-th row contains $m_{\mathcal{G}_l}$ consecutive ones corresponding to the $l$-th group. Then we can rewrite \eqref{SWAG_penalty}
\begin{equation}
\label{SWAG_penalty2}
    R_{\text{SWAG}}(\bm{A}) = \sum_{i=1}^n \sum_{l=1}^k f(u_{li}),
\end{equation}
where $u_{li}$ is the $(l,i)$ element of the matrix $\bm U = \bm Z \bm A\in\mathbb R^{k\times n}.$ In other words, for each pixel $i,$ $u_{li}$ is the sum of the $l$-th group coefficients coded in $\bm a_i$.

% Given that each column of $\bm{A}$ follows the group structure (\ref{group_structure_a}), each column of the resulting matrix $\bm{A}$ forms a vector whose entries are the $\ell_1$ norms of the abundance coefficients of each group. 

Using \eqref{SWAG_penalty2} to enforce SWAG, we can rewrite (\ref{eqn:general_framework}) as the following regularized problem: 
\begin{equation}
\label{problem3}
    \begin{aligned}
    \argmin_{{\bm{A},\bm{U},\bm{V} }} & \frac{1}{2} \|\bm{X} -\bm{BA}\|_{F}^2 + \lambda \sum_{i=1}^n \sum_{l=1}^k f(u_{li}) + \mathcal{I}_{\Delta_{r \times n}} (\bm{V}) \\
    \text{s.t.} & \ \bm{ZA} = \bm{U}, \ \bm{V} = \bm{A},
    \end{aligned}
\end{equation}
%Similarly to the previous formulation, we change the constrained problem into an unconstrained optimization problem via the augmented Lagrangian. 
 with its augmented Lagrangian defined as
\begin{equation}
    \label{AL_least_square}
     \begin{aligned}
     &\mathcal{L}_\rho(\bm{A},\bm{U},\bm{V}; \bm{C},\bm{D}) \\
     = & \frac{1}{2} \|\bm{X} -\bm{BA}\|_{F}^2 + \lambda \sum_{i=1}^n \sum_{l=1}^k f(u_{li}) + \mathcal{I}_{\Delta_{r \times n}} (\bm{V}) \\
     & + \frac{\rho}{2} \|\bm{ZA}- \bm U-\bm C\|_F^2     + \frac{\rho}{2} \|\bm{A}- \bm V- \bm D\|_F^2,
     \end{aligned}
\end{equation}
for dual variables $\bm{C}, \bm{D},$ and a positive constant $\rho$. Similarly to \eqref{ADMM1}, we now discuss each subproblem required by the ADMM iterations.
%consider the following subproblems:
\begin{comment}
    \begin{equation}
    \label{ADMM2}
   \left\{ \begin{aligned}
    \bm{A}^{\tau+1} = &  \argmin_{\bm{A}} \mathcal{L}_\rho(\bm{A}^{\tau},\bm{U}^{\tau},\bm{V}^{\tau}; \bm{C}^{\tau},\bm{D}^{\tau})  \\
    \bm{U}^{\tau+1} =& \argmin_{\bm{U}}  \mathcal{L}_\rho(\bm{A}^{\tau+1},\bm{U}^{\tau}; \bm{C}^{\tau}) \\
    \bm{V}^{\tau+1} = & \argmin_{\bm{V} }\mathcal{L}_\rho(\bm{A}^{\tau+1},\bm{V}^{\tau+1}; \bm{D}^{\tau})  \\
    \bm{C}^{\tau+1} =& \bm{C}^{\tau} + \bm U^{\tau+1} -\bm A^{\tau+1}\\
    \bm{D}^{\tau+1} = & \bm{D}^{\tau} +\bm V^{\tau+1}-\bm A^{\tau+1},
\end{aligned}\right.
\end{equation}

where $\tau$ counts the iteration number.
\end{comment}
\begin{comment}

\begin{equation}
    \label{ADMM2}
      \begin{aligned}
    \bm{A} \leftarrow & \arg \min_{\bm{A}}  \frac{1}{2} \|\bm{X}- \bm{BA}\|_F^2 +  \frac{\rho}{2} \|\bm{MA}-\bm U-\bm C\|_F^2 \\
    &+ \frac{\rho}{2} \|\bm{A}-\bm V-\bm D\|_F^2  \\
    \bm{U} \leftarrow& \arg \min_{\bm{U}}  \lambda R(\bm{U})_{f} +  \frac{\rho}{2} \|\bm{MA}- \bm U-\bm C\|_F^2\\
    \bm{V} \leftarrow & \arg \min_{\bm{V} }\mathcal{I}_{\Delta_r} (\bm{V}) + \frac{\rho}{2} \|\bm{A} -\bm V-\bm D\|_F^2 \\
    \bm{C} \leftarrow & \bm{ C } + \bm{U} -\bm{MA}\\
    \bm{D} \leftarrow & \bm{ D}+\bm V-\bm A
\end{aligned}
\end{equation}
\end{comment}

 The $\bm{A}$-subproblem  for (\ref{AL_least_square}) is equivalent to 
\[ \arg \min_{\bm{A}}\frac{1}{2} \|\bm{X}- \bm{BA}\|_F^2 +  \frac{\rho}{2} \|\bm{ZA}-\bm U-\bm C\|_F^2 
    + \frac{\rho}{2} \|\bm{A}-\bm V-\bm D\|_F^2,\]
which has a closed-form solution:
\begin{eqnarray}\label{eq:SWAG-Aupdate}
   && \bm{A}^{\tau +1}= (\bm{B}^T \bm{B} +  \rho \bm{Z}^T \bm{Z}+\rho \bm{I}_r)^{-1}\\
   && \quad \bigg(\bm{B}^T \bm{X} + \rho \bm{Z}^T (\bm{U}^{\tau} +\bm{C}^{\tau})+  \rho(\bm{V}^{\tau}+\bm{D}^{\tau})\bigg).
  \notag
\end{eqnarray}
% where $\bm{I}_r$ is $r \times r$ identity matrix.

 The $\bm{U}$-subproblem  for (\ref{AL_least_square}) can be expressed by
\[\arg \min_{\bm{U}}\lambda \sum_{i=1}^n \sum_{l=1}^k f(u_{li}) +  \frac{\rho}{2} \|\bm{U} - (\bm {ZA}^{\tau+1}-\bm C^{\tau})\|_F^2,\]
with the closed-form solution given elementwise by
\begin{equation}\label{eq:SWAG-U}
    \bm U^{\tau+1} = \prox_{f}(\bm {ZA}^{\tau+1}-\bm C^{\tau}, \lambda/\rho),
\end{equation}
where $\prox$ is evaluated component-wise.
Please refer to Appendix \ref{sect:prox-operators} for the proximal operator of the respective function $f$.
%defined in (\ref{SWAG_penalty}). Specifically, the proximal operators for different sparsity functions are given as follows: transformed $\ell_1$ (TL1) in (\ref{prox_tl1}), $\ell_1 -\ell_2$ in (\ref{prox_l_{1/2}}),  and $\ell_{\frac{1}{2}}$ in (\ref{prox_L12}).

The $\bm{V}$-subproblem defined by
\[\arg \min_{\bm{V}}\mathcal{I}_{\Delta_r} (\bm{V}) + \frac{\rho}{2} \|\bm{A}^{\tau+1} -\bm V-\bm D^{\tau}\|_F^2,\]
which can be solved by a simple projection:
\begin{equation}
     \bm{V} = \bm{\proj}_{{\Delta}_{r \times n}} (\bm{A}^{\tau+1} - \bm{D}^{\tau}).
 \end{equation}

%\ekaterina{Fix the (??) above}
We initialize the matrix $\bm A^0$ in the same way as in Algorithm \ref{alg:ADMM1}, set $\bm U^0=\bm Z\bm A^0$, and summarize the overall procedure in Algorithm \ref{alg:ADMM2}.
% After solving the problem, one has to simply calculate the global abundances using \eqref{eqn:global_abundance} for the final result. 
\vspace{0.5cm}

%\SetKwComment{Comment}{/* }{ */}

\begin{algorithm}
\caption{Enforcing SWAG via ADMM}\label{alg:ADMM2}
{\textbf{Input}: hyperspectral data $\bm{X}$ and bundle matrix $\bm{B}$}\\
{\textbf{Choose}: $\lambda, \text{maximum iteration } T $}\\
\textbf{Initialize}: $\bm{A}^{0}$ and $\bm{V}^{0}$ are obtained by FCLSU \citep{drumetz2019hyperspectral}, then  $\bm U^{0}= \bm{ZA}^0$; set $\bm{C}^0 = \bm{0}_{k \times n}, \bm{D}^0 =\bm{0}_{r \times n}$, $\tau =0$ ;\\
%\KwData{$\bm{X},\bm{B}, \text{maximum iteration } T $}
%\KwResult{$\bm{A}$}
%Initialize $\bm{A} $, $\tau =0$, and choose $\lambda$;\\
\While{$\tau \leq T $}{

\vspace{0.1cm}
 Update $\bm{A}^{\tau+1}$ via \eqref{eq:SWAG-Aupdate}\;

 \vspace{0.1cm}
 Update  $ \bm{U}^{\tau+1} $ via \eqref{eq:SWAG-U}\;

   \vspace{0.1cm}
   $\bm{V}^{\tau+1} = \bm{\proj}_{{\Delta}_r} (\bm{A}^{\tau+1} - \bm{D}^{\tau})$\;

\vspace{0.1cm}
   $\bm{C}^{\tau+1} = \bm{C}^{\tau}+\bm{U}^{\tau+1}-\bm{ZA}^{\tau+1}$\;

\vspace{0.1cm}
   $\bm{D}^{\tau+1} = \bm{D}^{\tau}+\bm{V}^{\tau+1}-\bm{A}^{\tau+1}$\;

   \vspace{0.1cm}
   $\tau = \tau + 1 $\;
}
{\textbf{Output}: $\widehat{\bm{A}} = \bm A^{\tau}$, $\widehat{\bm M}, \widehat{\bm S}$ via \eqref{eqn:global_abundance} and \eqref{global_endmember}, respectively.  }
\end{algorithm}

\section{EXPERIMENTS AND RESULTS}
\label{experiments_and_results}
In this section, we perform a comprehensive evaluation of various sparsity-promoting regularizers under certain group structures. We list all competing methods including two proposed approaches (Inter-TL1 and SWAG-TL1) in Section \ref{sect:compete-method}. We outline the bundle construction in Section \ref{sec:bundle_construction}.  Section \ref{sect:syn-data} is devoted to synthetic data experiments, while three real datasets of Samson, Jasper Ridge, and Houston are examined in Section \ref{sect:real-data}.
All experiments were performed on a 1.4 GHz QuadCore Intel Core i5 computer using MATLAB R2022a.

 % After solving the problem, one has to simply calculate the global abundances using \eqref{eqn:global_abundance} for the final result. The regularization parameters for each method, along with the total signatures $r$ for each dataset, are provided in the supplementary material. 

To quantitatively assess the abundance estimation performance of different unmixing methods, we employ the Root Mean Square Error (RMSE) between the estimated abundances $\widehat{\bm{M}}$ and the true abundances $\bm M$: %\yl{[we need to change A to M using (9)]}
\begin{equation}
\label{RMSE}
    \text{RMSE}(\widehat{\bm{M}}) = \frac{1}{n} \sum_{i=1}^n \sqrt{\frac{1}{k} \sum_{l=1}^k (m_{li} - \widehat{m}_{li})^2}.
\end{equation}
Using each material $l$, the overall spectral angle mapper (SAM) between the estimated one $\widehat{\bm S}$ and the true signature $\bm S$ is given by: %\yl{[do we need absolute value $|s_{li}^\top \widehat{s}_{li}|$? wait, $\widehat{s_{li}}$ is a scalar? ]} \gokul{$\widehat{s_{li}}$  is a vector. In term of matrix $\widehat{S} \in \mathbb R^{w \times k \times n}$ where $n$ is total pixels, $w$ is total spectral bands and $k$ is total number of endmembers.  $\widehat{s_{l,i}} $ is the $l$-th column of matrix $\widehat{S}$.}
\begin{equation}
    \label{SAM}
     \text{SAM}(\widehat{\bm{S}} ) = \frac{1}{k}\sum_{l=1}^k \arccos\bigg(\frac{\langle\h s_{l} , \widehat{\h s}_{l}\rangle}{\|\h s_{l}\|_2 \|\widehat{\h s}_{l}\|_2}\bigg),
\end{equation}
where $\h s_l$ is defined in \eqref{global_endmember}.
Lastly, we calculate the RMSE between the reconstructed data matrix $\widehat{\bm X}$ by using equation~(\ref{gloobal_abundance}) and the data matrix $\bm X$, i.e., %\yl{[how to recon using A or M? relate these calculations to (9) to (11)]}
\begin{equation}
\label{RE}
\text{RMSE}(\widehat{\bm{X}})  =     \frac{1}{n} \sum_{i=1}^n \sqrt{\frac{1}{w} \sum_{p=1}^w (x_{pi} - \widehat{x}_{pi})^2}.
\end{equation}
This metric is particularly helpful where there is no true abundance available. 
\begin{comment}
    To evaluate the performance in terms of material variability retrieval, we compute the RMSE between the true signatures and the estimated signatures: 
\begin{equation}
    \text{RMSE}(\widehat{\bm{S}} ) = \frac{1}{nk} \sum_{i=1}^n \sum_{l=1}^k \frac{1}{\sqrt{w}}\|s_{li} - \widehat{s}_{li}\|_2.
\end{equation}
\end{comment}

\begin{comment}
    \begin{equation}
    \label{SAM}
     \text{SAM} = \frac{1}{nk} \sum_{i=1}^n \sum_{l=1}^k \arccos\bigg(\frac{\langle s_{l,i} , \widehat{s_{l,i}}\rangle}{\|s_{l,i}\|_2 \|\widehat{s_{l,i}}\|_2}\bigg)
\end{equation}
\end{comment}

\begin{comment}

\begin{equation}
\label{SAM}
    \text{SAM} = \frac{1}{nk} \sum_{i=1}^n \sum_{l=1}^k \arccos\bigg(\frac {b_{li}^\top \widehat{s}_{li}}{||b_{li}||_2 ||\widehat{s}_{li}||_2}.\bigg)
\end{equation}
\end{comment}
%\subsection{Method Comparison}
\subsection{Comparison and Proposed Methods}
\label{sect:compete-method}
Here is an overview of the comparison and proposed methods that we consider in our experiments. 
%We previously discussed \textbf{Group LASSO}, \textbf{Elitist LASSO}, and \textbf{Fractional LASSO} in Section~(\ref{group_sparsity}). In addition to these, we also consider the following penalties:

\begin{itemize}

%\vspace{0.2cm}
\item \textbf{Fully Constrained Least Squares (FCLS)} \citep{heinz2001fully} solves (\ref{R-LLM}) without any additional regularization term, i.e., setting $\lambda$ to zero, nor does it take into account the sparsity and the bundle matrix.

\item \textbf{Inter-L1} uses the \(\ell_{\mathcal{G},2,1}\) mixed norm, which promotes sparsity in the vector whose entries are \(\|a^{\mathcal{G}_i}\|_2\). It encourages a small number of groups to be active, but prefers a dense mixture within each group.  In the literature, this penalty is also referred to as group LASSO \citep{yuan2006model}. To maintain consistency with our notation, we refer to it as Inter-L1. 
\vspace{0.15cm}

\item \textbf{Intra-L1}  adopts the \(\ell_{\mathcal{G},1,2}\) mixed norm, which promotes sparsity within each group. Unlike Group LASSO, it favors all groups to be active while selecting only a few active components within each group.  In the literature, this penalty is also referred to as Elitist LASSO  \citep{kowalski2009sparsity}. To stay consistent with our notation, we refer to it as Intra-L1. 
\vspace{0.15cm}

\item \textbf{SWAG-Lq} \citep{drumetz2019hyperspectral} uses the \(\ell_{\mathcal{G},1,q}\) regularization with $q<1$ to promote sparsity within and across groups. It encourages only a few groups to be active while also promoting sparsity within each group. In the literature, this penalty is also referred to as fractional LASSO. To maintain consistency with our notation, we refer to it as SWAG-Lq. 
\vspace{0.15cm}
\begin{comment}
    \item \textbf{Fractional LASSO \citep{drumetz2019hyperspectral} :} This method uses the \(\ell_{\mathcal{G},1,q<1}\) norm to promote sparsity within and across groups. It encourages only a few groups to be active while also promoting sparsity within each group. 
\vspace{0.15cm}
\end{comment}

\item \textbf{Inter-TL1:} This penalty effectively promotes inter-group sparsity where we use $\ell_2$-norm within each group and the TL1 function  \citep{zhang2018minimization,nikolova2000local,zhang2014minimization} to promote inter-group sparsity. The corresponding algorithm is detailed in Algorithm \ref{alg:ADMM1}  with its proximal operator defined in Appendix \ref{sect:appendB}.

%\item \textbf{SWAG-L12:} This penalty effectively promotes sparsity within and across groups. In particular, we use the $\ell_1$-norm within each group and the $\ell_1 - \ell_2$ (L12) \citep{esser2013method, yin2014ratio, lou2015computing, lou2018fast} penalty function across groups. The corresponding algorithm is detailed in  Algorithm \ref{alg:ADMM2} and the definition of L12 with its proximal operator is provided in \eqref{prox_L12}.  

 \item \textbf{SWAG-TL1:} This penalty works similarly to SWAG-L1 where we use the $\ell_1$ norm to promote sparsity within group and the TL1 function to promote sparsity across groups. The corresponding algorithm is detailed in Algorithm \ref{alg:ADMM2}.
 %and the definition of TL1 with its proximal operator is provided in (\ref{TL1}). 
%\vspace{0.2cm}
  \end{itemize}

 %\yl{[check the literature again to make sure this statement is true:]} 
 To the best of our knowledge, Inter-TL1 and SWAG-TL1 have never been used in hyperspectral unmixing and 
  are introduced in this work as two novel penalty terms.  For a fair comparison, we optimize the parameters of all competing methods to achieve the best performance in terms of RMSE$(\widehat{\bm M})$. For the Houston dataset, parameter tuning is performed with respect to RMSE$(\widehat{\bm X})$.
  %\yl{[Houston doesn't have GT M, how did you tune parameters]}

\subsection{Bundle Construction}
\label{sec:bundle_construction}
%\yl{[We assume $S$ is given, this paragraph is to describe the bundle idea, by focusing on the idea that each group/bundle is composed by different appearance of the same endmembber]}
Constructing spectral bundles is one of the straightforward approaches when addressing endmember variability by generating a library of candidate signatures for each pure material. In this work, we use the  Automated Endmember Bundles (AEB) method to construct an endmember bundle. Its idea is to randomly sample a certain subset of the data  and run the Endmember Extraction
Algorithms (EEA) \citep{somers2012automated} on each subset to form an endmember candidate. With a total of \( k \) subsets selected without replacement, we obtain a total of \( r = k \cdot m_l \) signatures, where \( m_l \) represents the number of elements in the $l$-th endmember group. Sometimes, it is possible that the endmembers are not aligned from one subset to the other, so a clustering step, e.g., $k$-means,  is required to group the candidate endmembers into classes. 
%Similarly to \citep{somers2012automated}, the clustering step is performed using the $k$-means algorithm with the spectral angle as a similarity measure. 
The output of this technique is the bundle matrix $\bm{B}$ of endmember candidates.

\subsection{Synthetic Data}\label{sect:syn-data}

We start with experiments using a synthetic hyperspectral image of $50 \times 50$ pixels with spatially correlated values from a Gaussian random field as the abundance map. We consider three pure materials, i.e., vegetation, dirt, and water, each of which has a library of signatures with spectral variability.  
In particular, the vegetation spectra are created using the PROSPECT-D model \citep{feret2017prospect}, while dirt and water spectra are generated using simplified Hapke \citep{hapke1981bidirectional} and atmospheric models, respectively. We select 198 bands and 30 representative endmember signatures for each material to construct the bundle matrix. For each pixel, endmembers are randomly selected from the set of synthesized signatures, and the data matrix $\bm X$ is generated from \eqref{eq:LMM} with an added Gaussian noise 
%with variability,  which generalizes \eqref{eq:LMM} by allowing a different endmember matrix for each pixel \yl{[Here instead of allowing, describe how to generate different endmembers]}. Finally, white Gaussian noise is added to the image 
to achieve a signal-to-noise ratio of $30$ dB. Additional details on the synthetic data generation and supporting software can be found in \citep{borsoi2021spectral}. 
%After generating the synthetic data, we construct a bundle matrix using the AEB method, resulting in a set of $30$ representative endmember signatures.

%We use the Automated Endmember Bundle (AEB) method to extract a bundle matrix. Specifically, we sample, without replacement, 10 subsets that contain 10\% of the pixels in the data set.  For each subset, we extract three endmembers using the VCA algorithm. The resulting bundle matrix has dimensions of $198 \times 30$. For initialization, we use the FCLSU algorithm \citep{drumetz2019hyperspectral}, which is a modified classical Fully Constrained Least Squares (FCLS) method. The
%regularization parameters of each method are recorded in the supplementary material file. 

We record the quantitative results of RMSE$(\widehat{\bm{M}})$,  SAM$(\widehat{\bm{S}}),$ and RMSE$(\widehat{\bm{X}})$  recovered by the competing methods in Table \ref{tab:synthetic}.  It is evident that the Inter-TL1 and SWAG-TL1 penalties outperform others in terms of estimating the abundance map.
%and the endmember signatures. %In contrast, the Intra-L1 and SWAG-L12 penalties are the poorest performers in abundance estimation.
Intra-L1 achieves the best result in terms of SAM$(\widehat{\bm{S}})$, but performs poorly under other metrics. Moreover, we provide the computational time in  Table \ref{tab:synthetic}, showing that FCLS and Inter-L1 are the fastest methods, while other methods have comparable speeds.
%For endmember recovery, Inter-TL1, Inter-L1, and SWAG-TL1 achieve the best RMSE and  SAM values, indicating their superior ability in recovering endmembers close to the ground truth. 
%In terms of the RMSE between the reconstruction of pixels and true pixels values, both SWAG-TL1 and Inter-TL1 outperform other penalties.

Figure \ref{fig:synthetic_abundance} presents the abundance maps estimated by various methods
with the ground truth displayed in the first row. 
Visually speaking, SWAG-TL1 achieves results that are the closest to the reference maps, although all the penalties exhibit some level of noise in the estimation. 
%SWAG-L12, and SWAG-Lq have similar visual results for all three materials. This similarity is expected, as both penalties promote sparsity both within and across groups, leading to comparable abundance maps.
Specifically, the Intra-L1 penalty has a tendency to favor dense mixtures across groups, which might affect its overall performance. 

%The abundance estimation using Intra-L1 performs worse compared to other methods. Although  promotes sparsity within groups, it also 

\begin{table*}[!ht]

\centering
\resizebox{\textwidth}{!}{
\begin{tabular}{ |p{1.4cm}|p{1.0cm}|p{1.0 cm}|p{1.0 cm}|p{1.0 cm}|p{1.4 cm}|p{1.0 cm}|}

\hline
\multicolumn{7}{|c|}{Synthetic Data } \\
\hline
Methods & FCLS & Inter-L1&  Intra-L1  & SWAG-Lq & Inter-TL1& SWAG-TL1 \\  %& SWAG-Lq\\
\hline
RMSE$(\widehat{\bm{M}})$ &  0.065 &  0.055  &0.101 & \textcolor{black}{0.065}
 &   \textcolor{blue}{ 0.043} & \textcolor{red}{0.032}  \\ %&& 0.0893\\
%RMSE$(\widehat{\bm{S}})$ &  -- & \textcolor{black}{0.043} &  0.056 & \textcolor{black}{0.051} &  \textcolor{blue}{ 0.039} & \textcolor{red}{0.038} \\ %& &0.0867\\

SAM$(\widehat{\bm{S}})$ & --  &  \textcolor{black}{3.380} & \textcolor{blue}{2.394} & \textcolor{black}{3.452}&   3.248& \textcolor{red}{2.194}\\%&  &8.1411\\
RMSE$(\widehat{\bm{X}})$ &  0.014 & 0.012 & 0.014  & \textcolor{blue}{0.011} &   \textcolor{blue}{0.011}& \textcolor{red}{0.009}\\ %&& 0.0128\\
Time (s) & 0.34  & 0.74 & 6.32 & 8.44 & 5.65& 6.82 \\%& &14.01 \\
%$\lambda$ & $\times$  & 1 & 2  & 0.4 & 0.04\\
\hline

\end{tabular} 
}
\caption{Comparison of the synthetic data with the best value in red and the second best in blue. Since FCLS does not output the mixing matrix, we indicate its error on $\widehat{\bm S}$ with a dash (--). }%\yl{[use -- rather than -; keep 3-digit accuracy except for time (2 digit) or can you do 4-digit precision for SAM? List these methods in the order of FCLS, inter-L1, intra-L1, SWAG-Lq, Inter-TL1, and SWAG-TL1; also swap the rows of RMSE X and SAM S ]} 
\label{tab:synthetic}
\end{table*}\vspace{0.2cm}

\begin{comment}
\begin{tabular}{ |p{1.4cm}|p{1.4cm}|p{1.2cm}|p{1.2cm}|p{1.8cm}|}
\hline
\multicolumn{5}{|c|}{Synthetic Data }  \\
\hline
Methods
 & PLMM & ELMM & GLMM & NMF-QMV\\
\hline
RMSE(A) & 0.0636&0.0628 &   0.0634 & 0.0564
\\

RMSE(X) &0.0084 &0.0057  &   0.0011  & 0.0113 
\\

\hline
\end{tabular} 
\end{comment}

\def\picw{0.3}
\begin{figure}[!ht]
\centering
\setlength{\tabcolsep}{1pt}
	\begin{tabular}{cccc}
&Water & Soil & Vegetation\\
		\raisebox{0.5\normalbaselineskip}[0pt][0pt]{\rotatebox{90}{\footnotesize Reference}}&
        \includegraphics[width=\picw\linewidth, height=0.065\textheight]{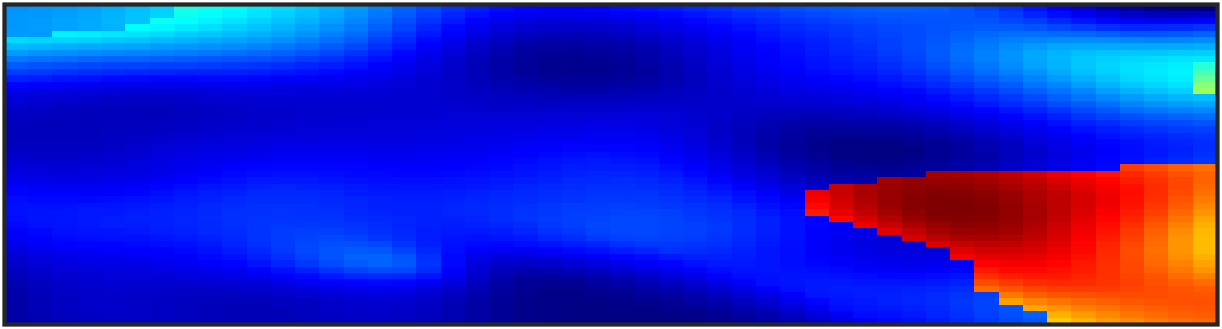}&
		\includegraphics[width=\picw\linewidth, height=0.065\textheight]{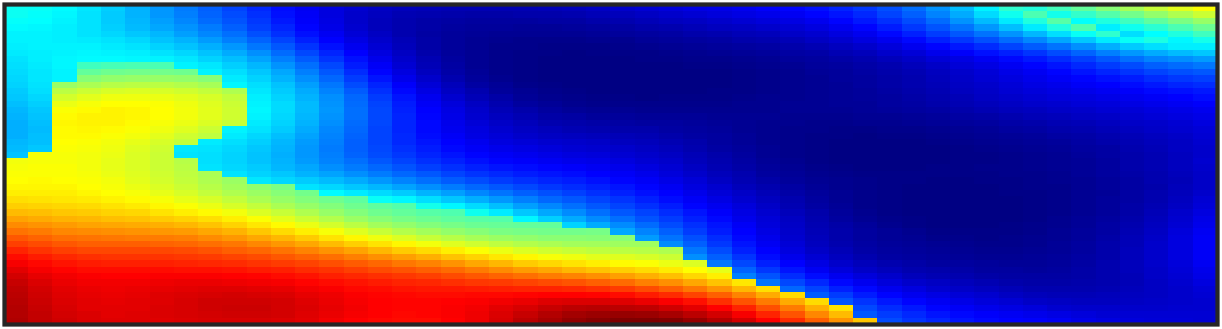}&
		\includegraphics[width=\picw\linewidth, height=0.065\textheight]{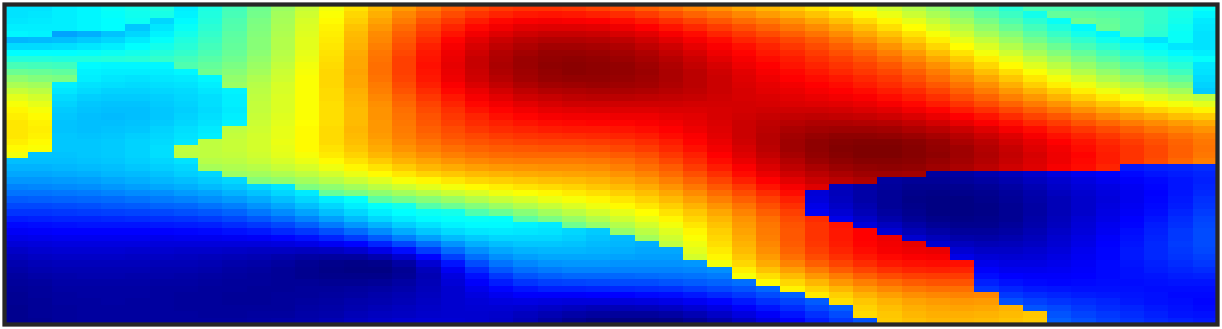}\\
		\raisebox{1.0\normalbaselineskip}[0pt][0pt]{\rotatebox{90}{\footnotesize FCLS}}&
        \includegraphics[width=\picw\linewidth,height=0.065\textheight,]{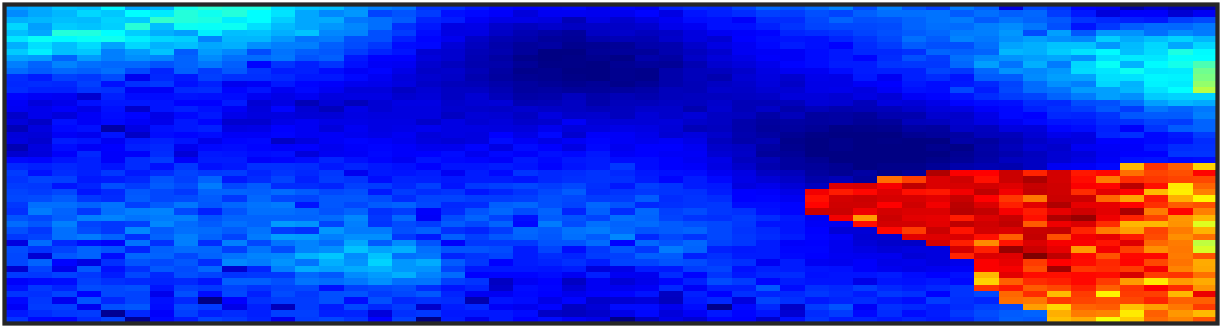}&
		\includegraphics[width=\picw\linewidth,height=0.065\textheight]{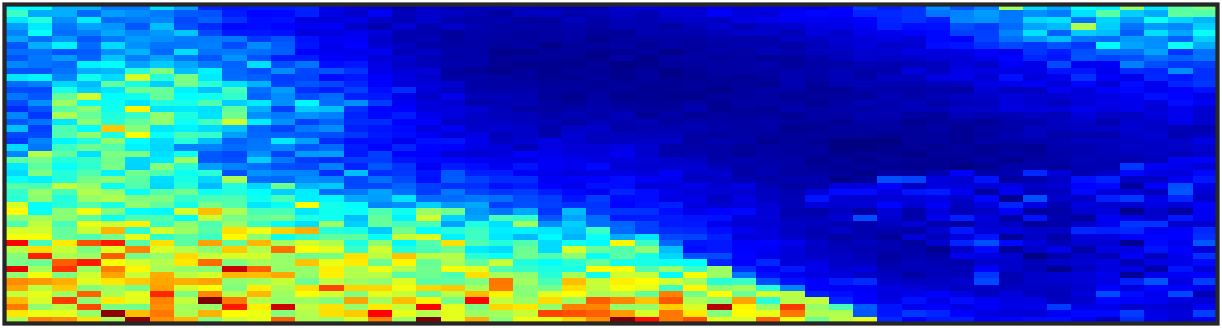}&
        \includegraphics[width=\picw\linewidth,height=0.065\textheight]{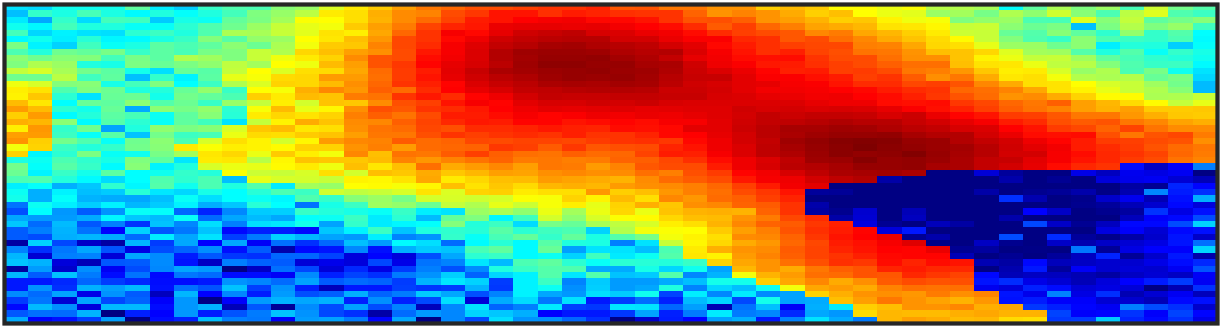}\\

	\raisebox{1.0\normalbaselineskip}[0pt][0pt]{\rotatebox{90}{\footnotesize \shortstack{Inter -\\L1}}}&
        \includegraphics[width=\picw\linewidth,height=0.065\textheight]{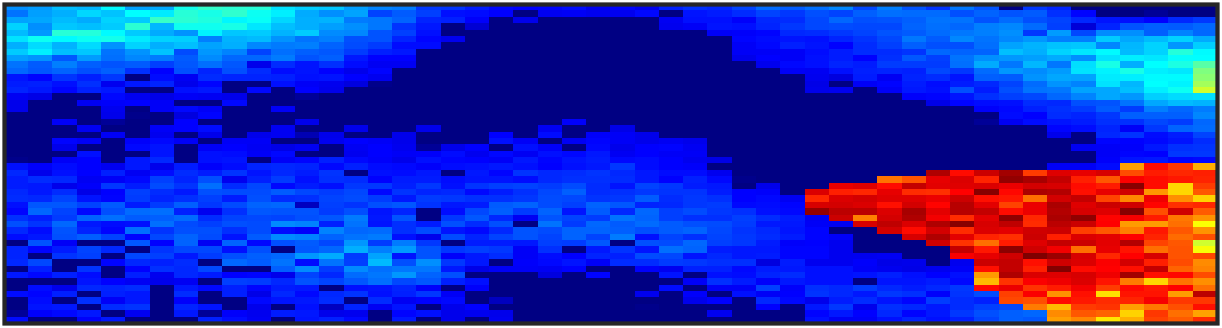}&
		\includegraphics[width=\picw\linewidth,height=0.065\textheight]{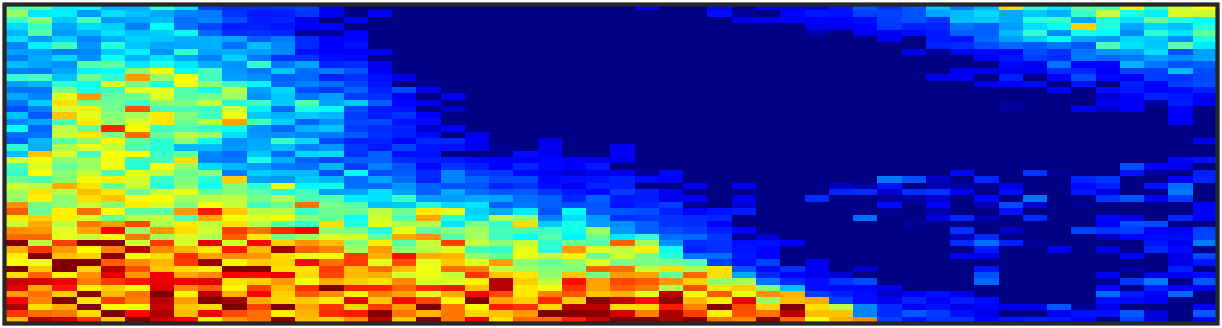}&
		\includegraphics[width=\picw\linewidth,height=0.065\textheight]{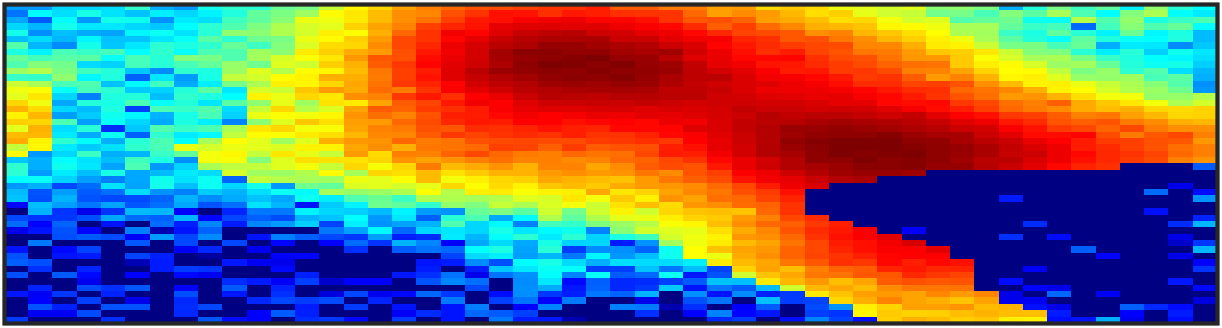}\\

        \raisebox{1.0\normalbaselineskip}[0pt][0pt]{\rotatebox{90}{\footnotesize \shortstack{Intra -\\ L1}}}&
        \includegraphics[width=\picw\linewidth,height=0.065\textheight]{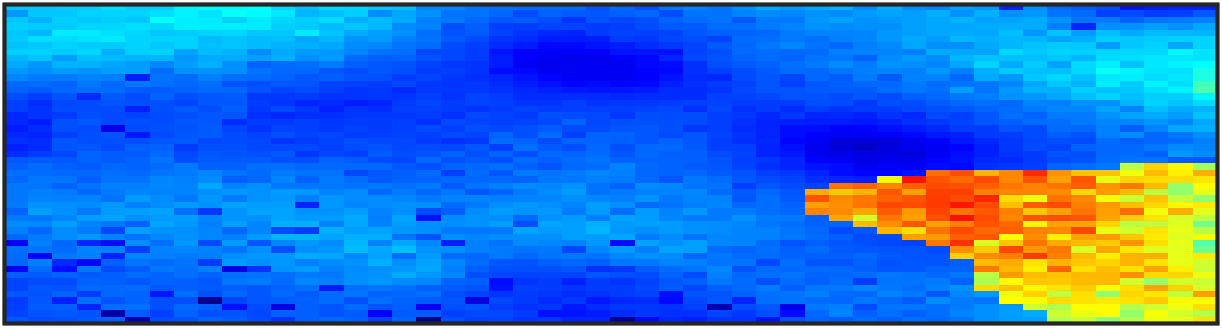}&
		\includegraphics[width=\picw\linewidth,height=0.065\textheight]{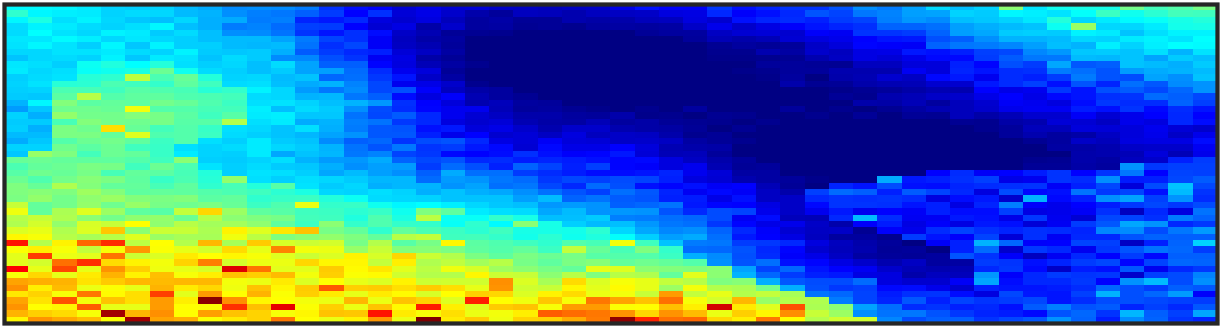}&
		\includegraphics[width=\picw\linewidth,height=0.065\textheight]{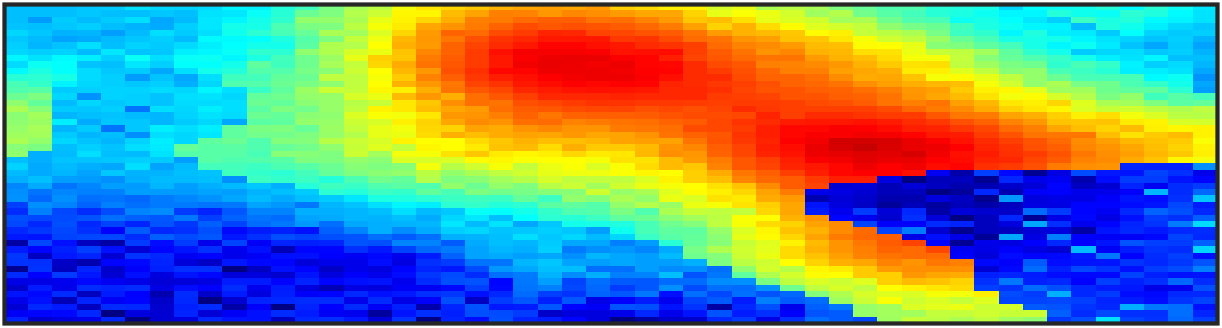}\\
	
 \raisebox{0.5\normalbaselineskip}[0pt][0pt]{\rotatebox{90}{\footnotesize \shortstack{SWAG - \\Lq}}}&
        \includegraphics[width=\picw\linewidth,,height=0.065\textheight]{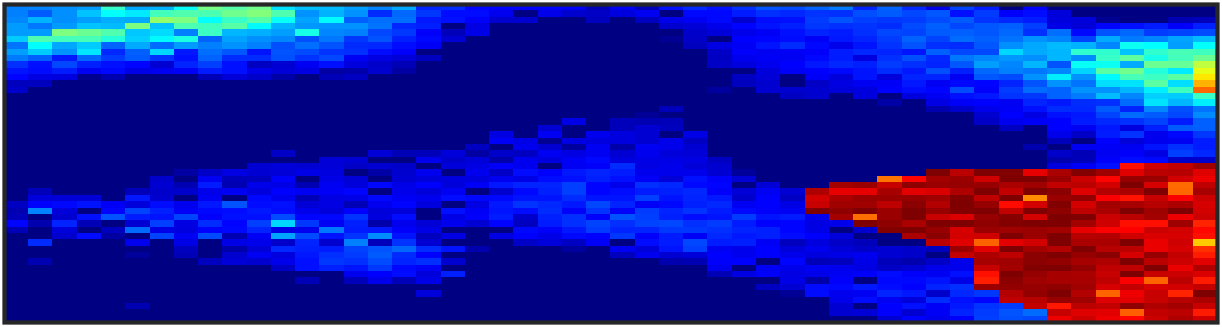}&
		\includegraphics[width=\picw\linewidth,height=0.065\textheight]{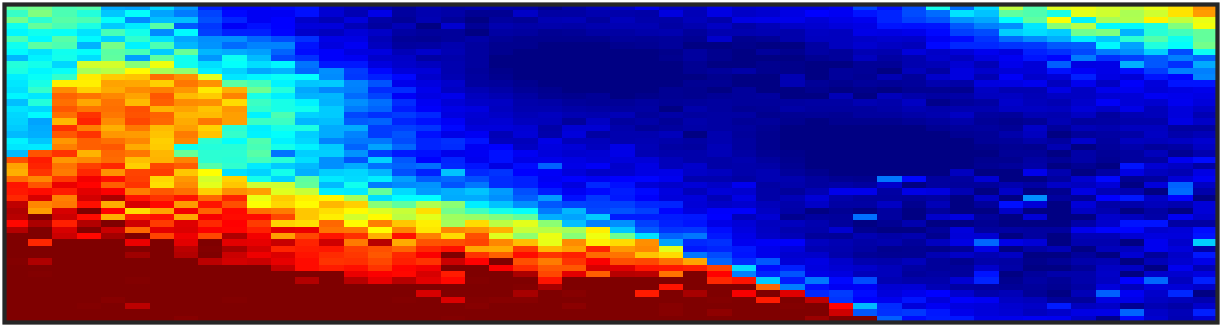}&
		\includegraphics[width=\picw\linewidth,height=0.065\textheight]{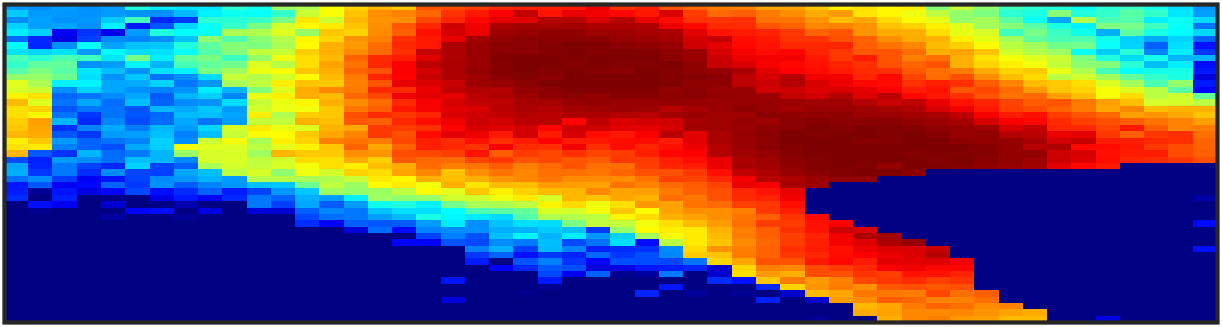}\\

			\raisebox{1.0\normalbaselineskip}[0pt][0pt]{\rotatebox{90}{\footnotesize \shortstack{Inter - \\ TL1}}}&
        \includegraphics[width=\picw\linewidth,height=0.065\textheight]{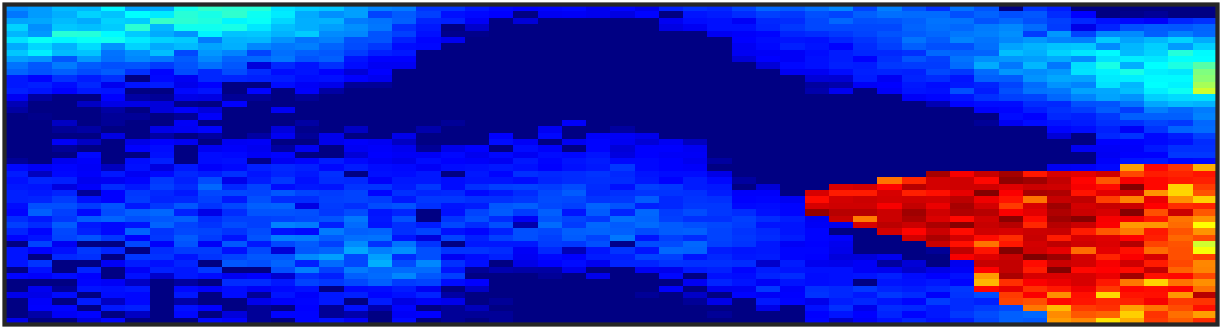}&
		\includegraphics[width=\picw\linewidth,height=0.065\textheight]{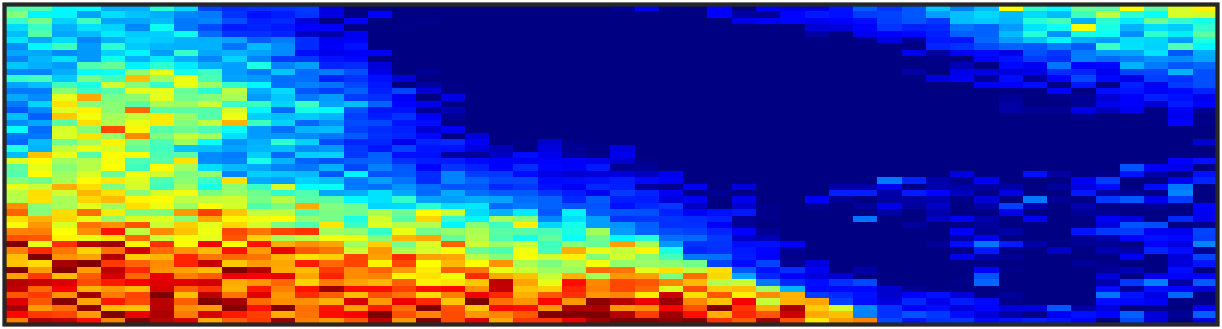}&
		\includegraphics[width=\picw\linewidth,height=0.065\textheight]{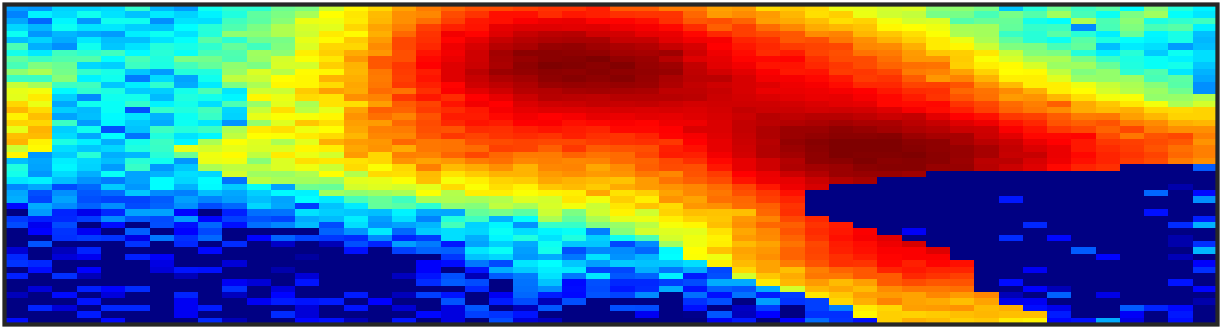}\\

        \raisebox{1.0\normalbaselineskip}[0pt][0pt]{\rotatebox{90}{\footnotesize \shortstack{SWAG - \\TL1}}}&
        \includegraphics[width=\picw\linewidth,height=0.065\textheight]{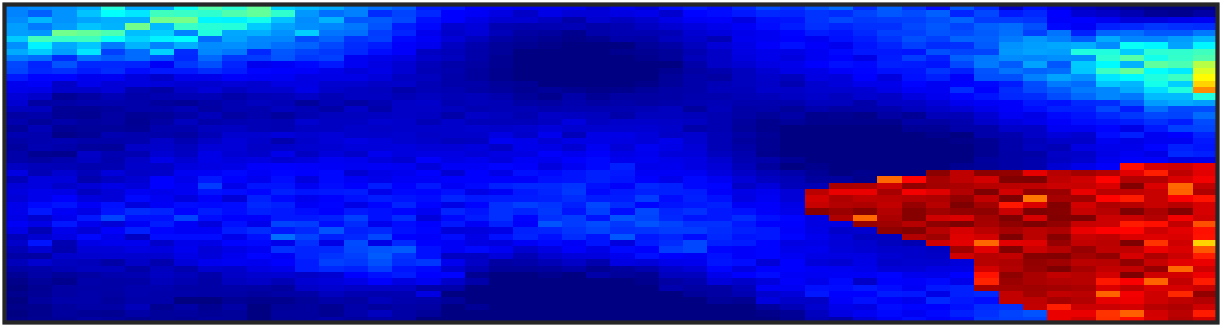}&
		\includegraphics[width=\picw\linewidth,height=0.065\textheight]{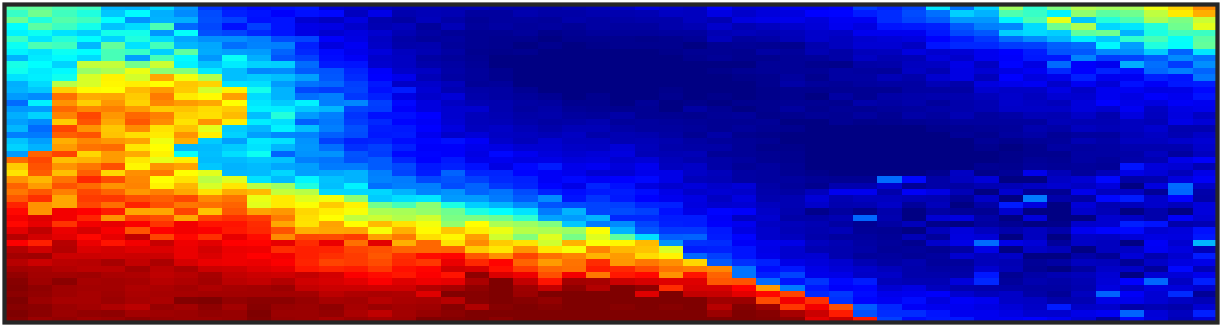}&
		\includegraphics[width=\picw\linewidth,height=0.065\textheight]{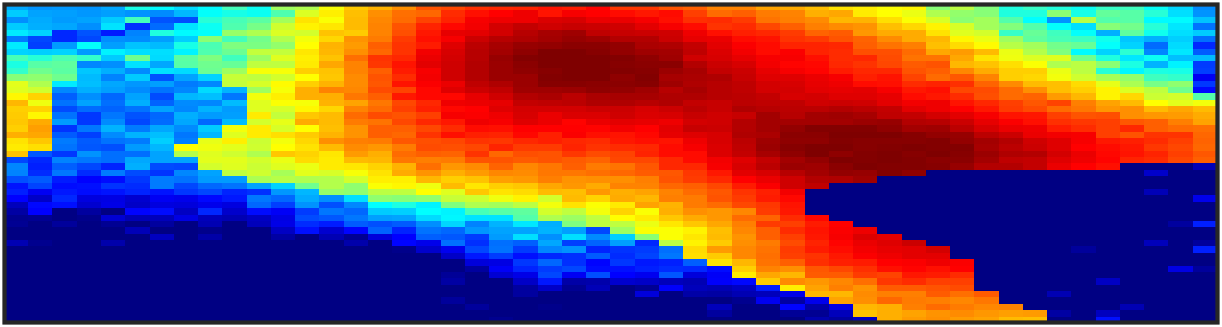}\\

	\begin{comment}
		\raisebox{1.0\normalbaselineskip}[0pt][0pt]{\rotatebox{90}{\footnotesize \shortstack{SWAG -\\ L12}}}&
        \includegraphics[width=\picw\linewidth,height=0.065\textheight]{synthetic/A_L12_im_Material1.png}&
		\includegraphics[width=\picw\linewidth,height=0.065\textheight]{synthetic/A_L12_im_Material2.png}&
		\includegraphics[width=\picw\linewidth,height=0.065\textheight]{synthetic/A_L12_im_Material3.png}\\

		\raisebox{1.0\normalbaselineskip}[0pt][0pt]{\rotatebox{90}{\footnotesize \shortstack{SWAG - \\Lq}}}&
        \includegraphics[width=\picw\linewidth,height=0.065\textheight]{synthetic/A_LP_im_Material1.png}&
		\includegraphics[width=\picw\linewidth,height=0.065\textheight]{synthetic/A_LP_im_Material2.png}&
		\includegraphics[width=\picw\linewidth,height=0.065\textheight]{synthetic/A_LP_im_Material3.png}\\
        \end{comment}
	\end{tabular}
	 \caption{Abundance maps estimated by the competing methods on the synthetic data with the ground truth (top row).}% \yl{[List these methods in the order of FCLS, inter-L1, intra-L1, SWAG-Lq, Inter-TL1, and SWAG-TL1]}}
    \label{fig:synthetic_abundance}
\end{figure}

\subsection{Real Data}\label{sect:real-data}

We investigate three real datasets, namely Samson, Jasper Ridge, and Houston, the last of which does not have ground-truth abundance maps.

\medskip
\subsubsection{Samson Data}
The Samson dataset \citep{Samson} has dimensions of $952 \times 952 \times 156$ and consists of three materials: soil, tree, and water. Due to the large size of the original image, which makes computations highly expensive, we use a smaller subimage of size $95 \times 95 \times 156$ for our experiments.  As is common with many real-world datasets, a reference library of endmembers is unavailable.  We construct a bundle matrix using the AEB method with a total of 30 endmember signatures. 

We use the ground truth abundance maps provided on the website \citep{Samson} to compute the RMSE between the estimated and true abundance values. As shown in Table \ref{tab:samson}, the group sparsity enforced by SWAG outperforms other types of group sparsity, as both SWAG-Lq and SWAG-TL1 achieve the best abundance estimation.
%all methods outperform the FCLS baseline in terms of abundance estimation. Among them, the SWAG-TL1 method achieves the best results for abundance estimation, while 
%Furthermore, the Inter-TL1 method excels in minimizing the RMSE and SAM \yl{[Inter-TL1 is not the best for SAM]} of the reconstructed image.  
In Figure \ref{fig:samson_abundance}, we show the abundance plot of the three materials with the ground truth in the first row. One can see that both inter-group sparsity and SWAG better preserve the hot spot regions in vegetation and water, compared to intra-group sparsity and FCLS, while SWAG-TL1 produces the best approximation in the soil abundance. 

% SWAG-Lq provides sparser abundance maps compared to other penalties; however, they fail to capture the boundary region of the materials accurately. Although all penalties exhibit some level of noise in the estimation, SWAG-TL1 achieves results that are the closest to the reference maps. \\

\begin{table*}[!ht]
\centering

\resizebox{\textwidth}{!}{
\begin{tabular}{ |p{1.4cm}|p{1.0cm}|p{1.0 cm}|p{1.0 cm}|p{1.0 cm}|p{1.4 cm}|p{1.0 cm}|}

\hline
\multicolumn{7}{|c|}{Samson Data } \\
\hline
Methods & FCLS &  Inter-L1 & Intra-L1& SWAG-Lq& Inter-TL1& SWAG-TL1 \\%& SWAG-L12 \\%&& SWAG-Lq\\
\hline
RMSE$(\widehat{\bm{M}})$ & 0.521 & \textcolor{blue}{ 0.199} &  0.227
 & \textcolor{red}{0.164} &  \textcolor{blue}{ 0.199} &\textcolor{red}{0.164}\\% &0.1650 \\%& &0.1329\\
%RMSE$(\widehat{\bm{S}})$ &  -- &  & -  & - &    & \\
SAM$(\widehat{\bm{S}})$ & -- &  \textcolor{black}{3.362} & \textcolor{blue}{3.331} & \textcolor{red}{3.298}& \textcolor{black}{3.360}  &\textcolor{black}{3.390}\\% & \textcolor{red}{3.3002}\\%&&5.2393\\
RMSE$(\widehat{\bm{X}})$ &  0.010 & 0.010 & 0.012 & \textcolor{black}{ 0.010} &   \textcolor{blue}{0.009} &\textcolor{red}{0.008} \\%&  0.0133\\%&& 0.0122\\

 Time(s) & 0.86 & 2.30 & 14.60 & 20.05 &15.65 &24.10 \\%&21.60 \\%&&27.14 \\
%$\lambda$ & $\times$  & 1 & 2  & 0.4 & 0.04\\
\hline

\end{tabular} 
}
\caption{Comparison of the Samson data with the best value in red and the second best in blue. Since FCLS does not output the mixing matrix, we indicate its error on $\widehat{\bm S}$ with a dash (--). }%\yl{[check IEEE to see whether caption is above or below the tab/fig; and adjust accordingly; why not including RMSE of S?]}}
\label{tab:samson}
\end{table*}

\begin{comment}
    \begin{figure}[!ht]
    \centering
    \includegraphics[width=1.0\linewidth, height=0.5\textheight]{figures/samson_abundance.eps}
    \caption{Abundance maps estimated by the tested methods with the ground truth (top row).}
    \label{fig:samson_abundance}
\end{figure}
\end{comment}

\def\picw{0.3}
\begin{figure}
\centering
\setlength{\tabcolsep}{1pt}
	\begin{tabular}{cccc}
&Vegetation & Water & Soil\\
		\raisebox{0.5\normalbaselineskip}[0pt][0pt]{\rotatebox{90}{\footnotesize Reference}}&
        \includegraphics[width=\picw\linewidth, height=0.065\textheight]{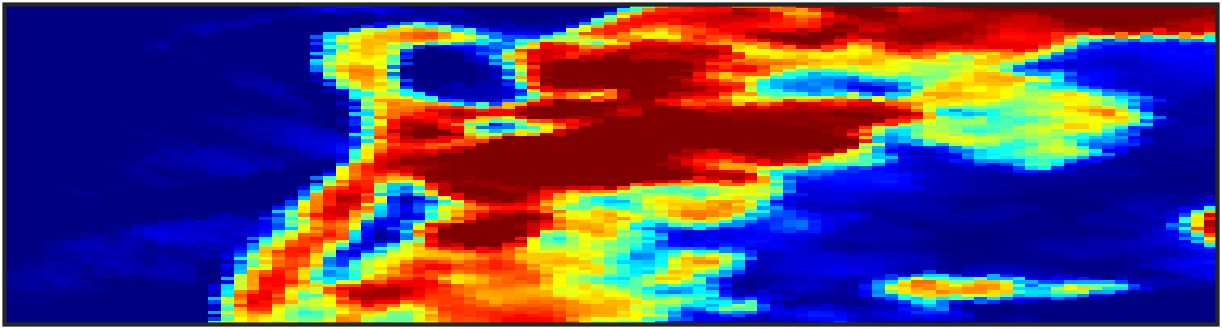}&
		\includegraphics[width=\picw\linewidth, height=0.065\textheight]{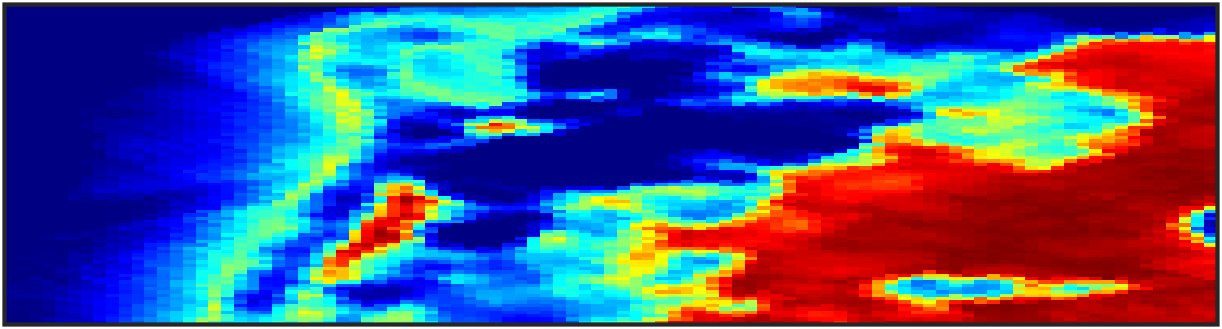}&
		\includegraphics[width=\picw\linewidth, height=0.065\textheight]{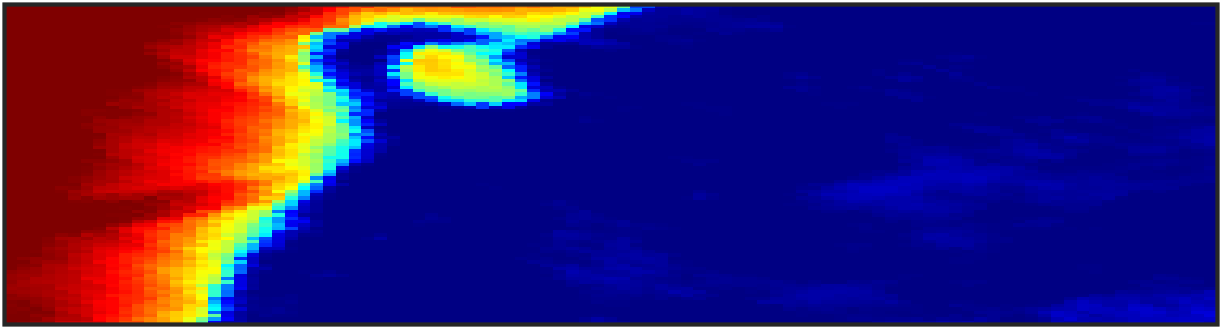}\\
		\raisebox{1.0\normalbaselineskip}[0pt][0pt]{\rotatebox{90}{\footnotesize FCLS}}&
        \includegraphics[width=\picw\linewidth,height=0.065\textheight,]{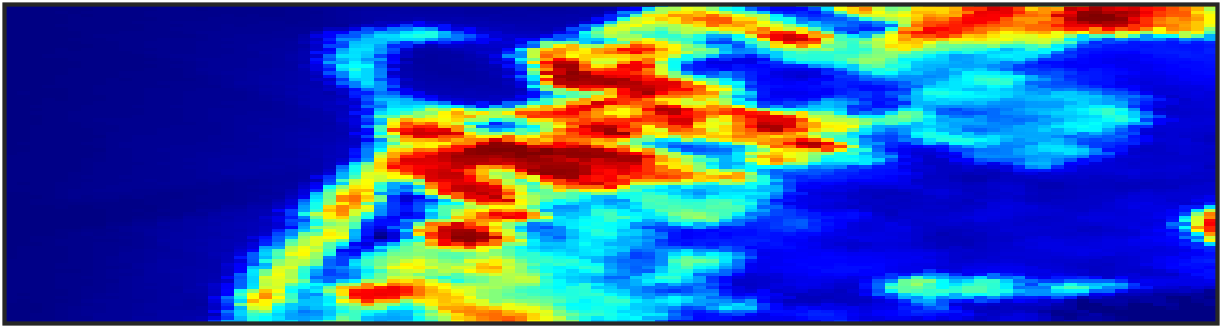}&
		\includegraphics[width=\picw\linewidth,height=0.065\textheight]{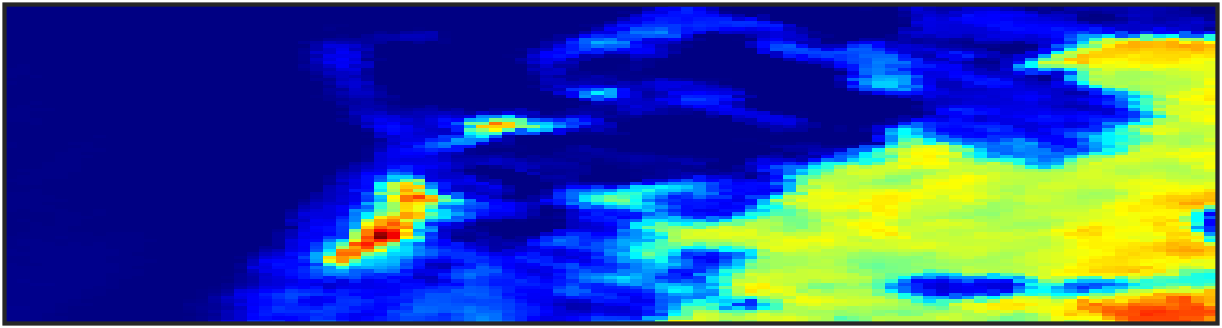}&
        \includegraphics[width=\picw\linewidth,height=0.065\textheight]{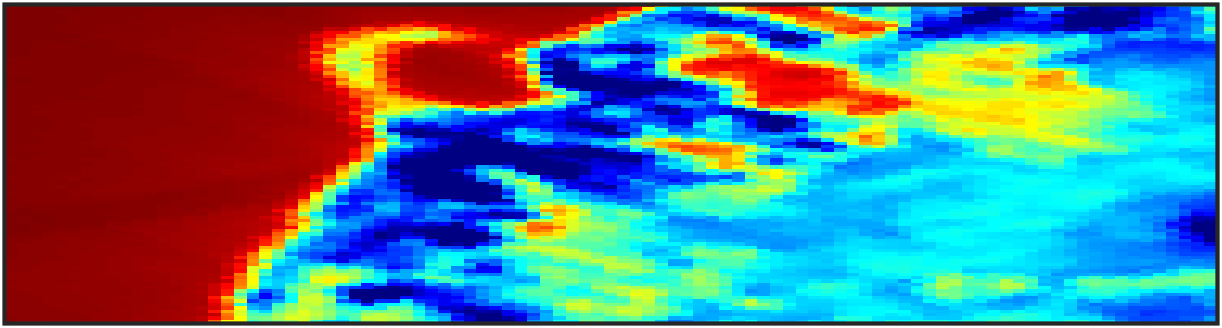}\\

        \raisebox{1.0\normalbaselineskip}[0pt][0pt]{\rotatebox{90}{\footnotesize \shortstack{Inter - \\L1}}}&
        \includegraphics[width=\picw\linewidth,height=0.065\textheight]{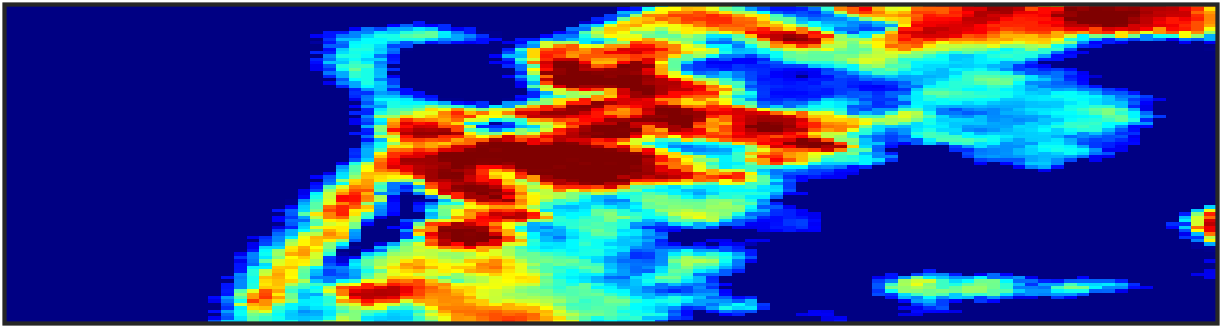}&
		\includegraphics[width=\picw\linewidth,height=0.065\textheight]{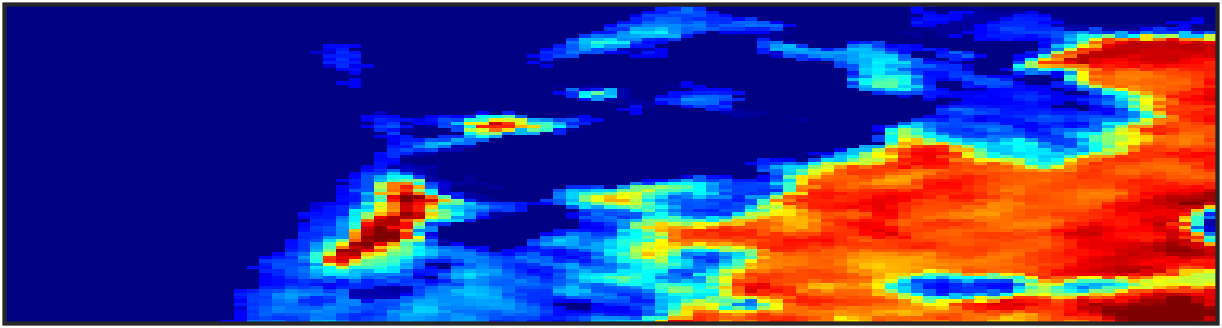}&
		\includegraphics[width=\picw\linewidth,height=0.065\textheight]{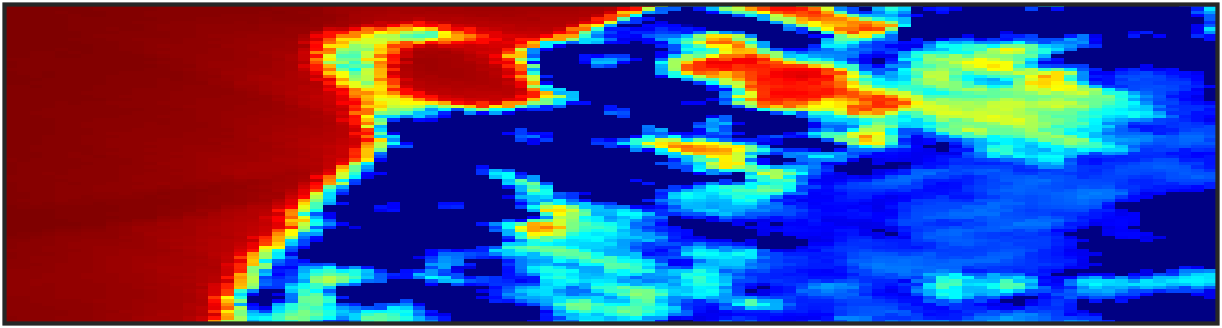}\\
        
        \raisebox{1.0\normalbaselineskip}[0pt][0pt]{\rotatebox{90}{\footnotesize \shortstack{Intra -\\ L1}}}&
        \includegraphics[width=\picw\linewidth,height=0.065\textheight]{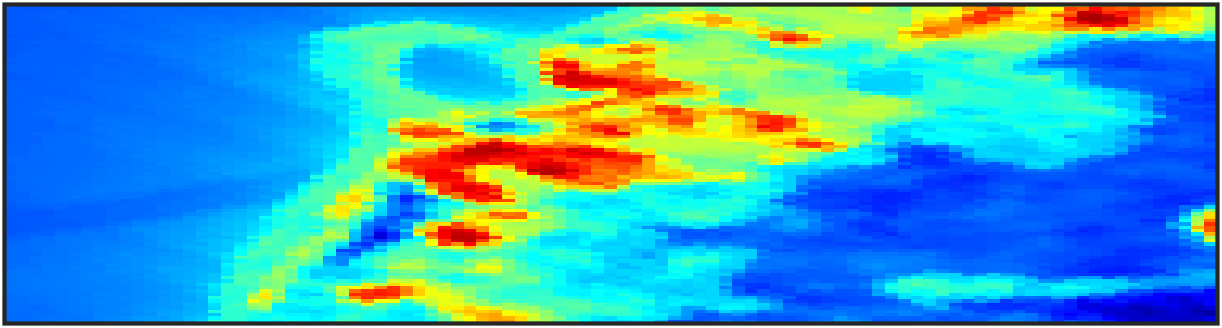}&
		\includegraphics[width=\picw\linewidth,height=0.065\textheight]{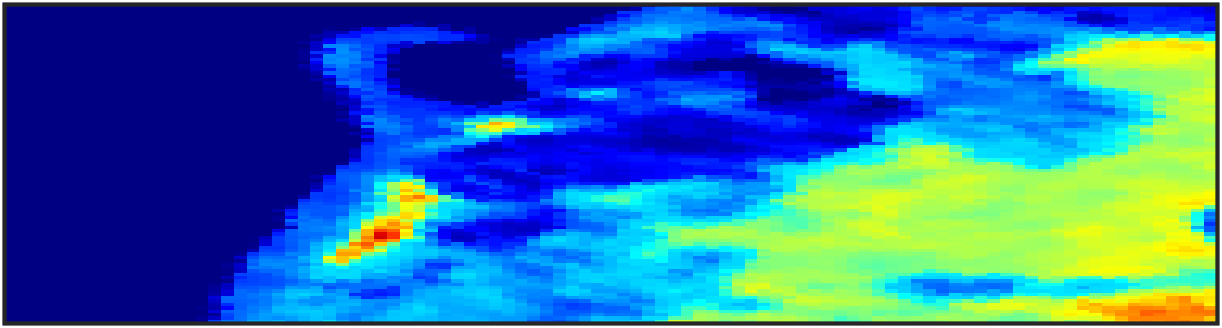}&
		\includegraphics[width=\picw\linewidth,height=0.065\textheight]{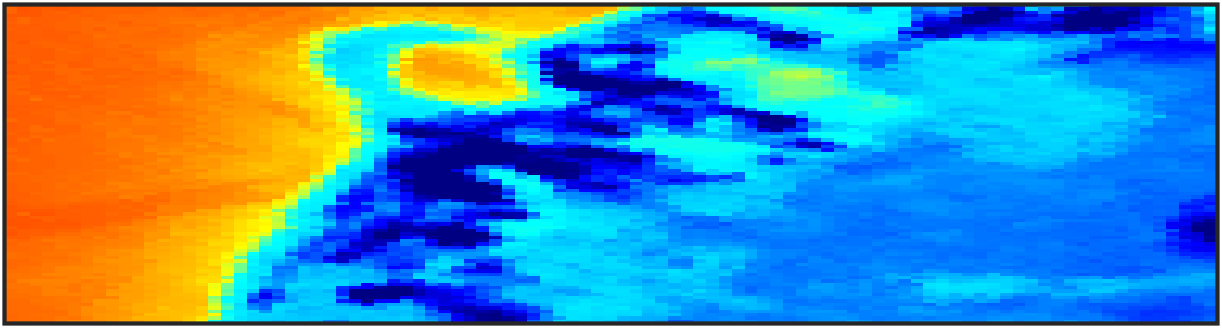}\\

	\raisebox{0.5\normalbaselineskip}[0pt][0pt]{\rotatebox{90}{\footnotesize \shortstack{SWAG - \\Lq}}}&
        \includegraphics[width=\picw\linewidth,,height=0.065\textheight]{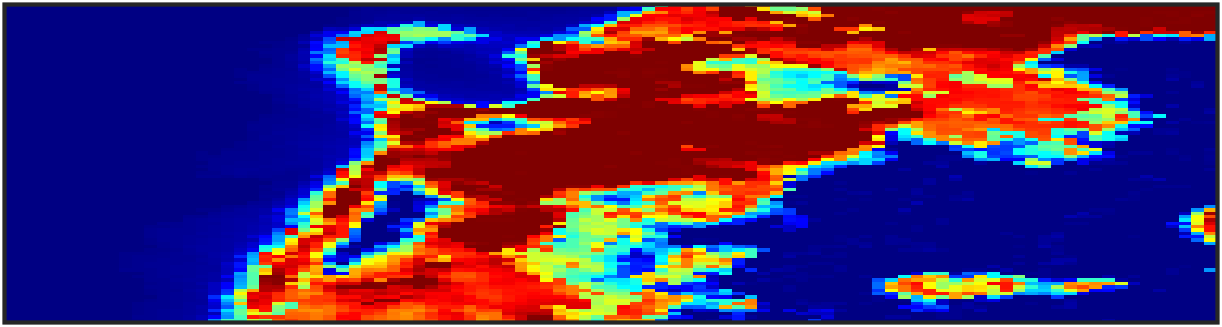}&
		\includegraphics[width=\picw\linewidth,height=0.065\textheight]{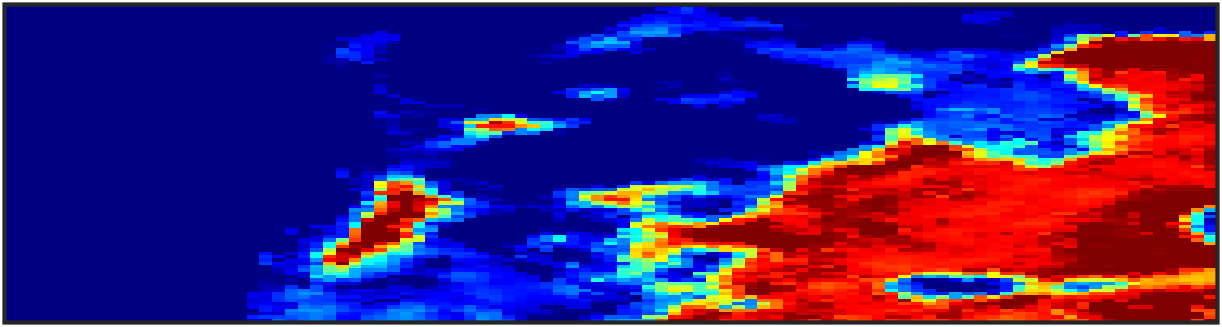}&
		\includegraphics[width=\picw\linewidth,height=0.065\textheight]{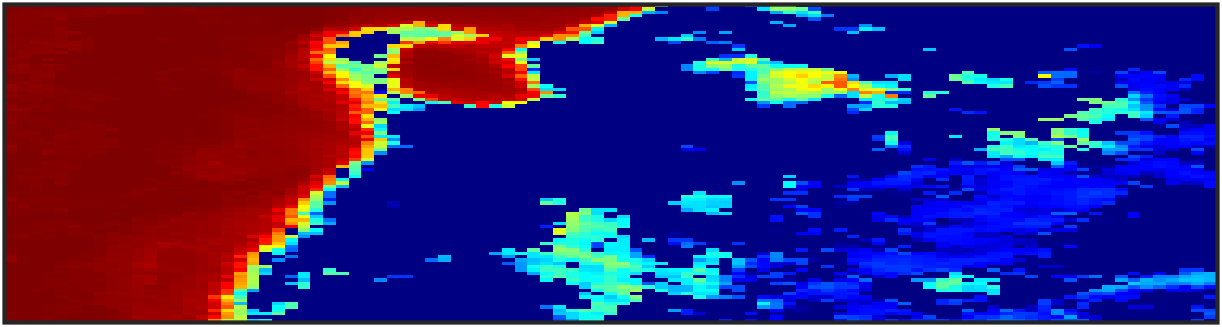}\\

			\raisebox{1.0\normalbaselineskip}[0pt][0pt]{\rotatebox{90}{\footnotesize \shortstack{Inter - \\ TL1}}}&
        \includegraphics[width=\picw\linewidth,height=0.065\textheight]{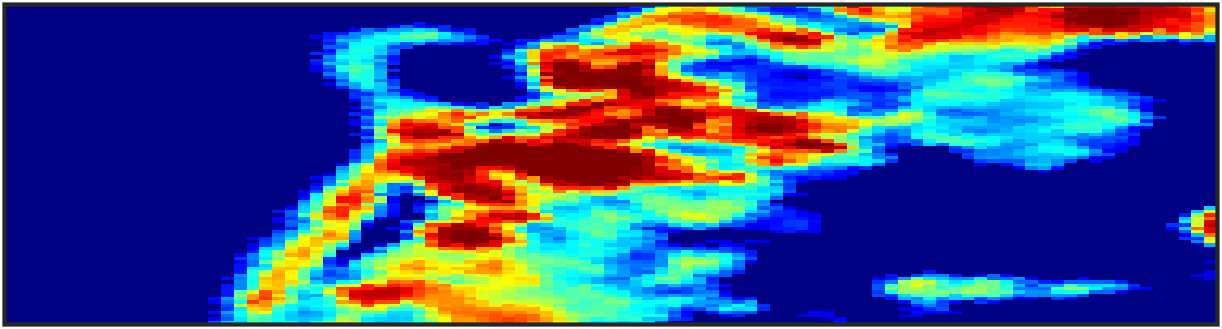}&
		\includegraphics[width=\picw\linewidth,height=0.065\textheight]{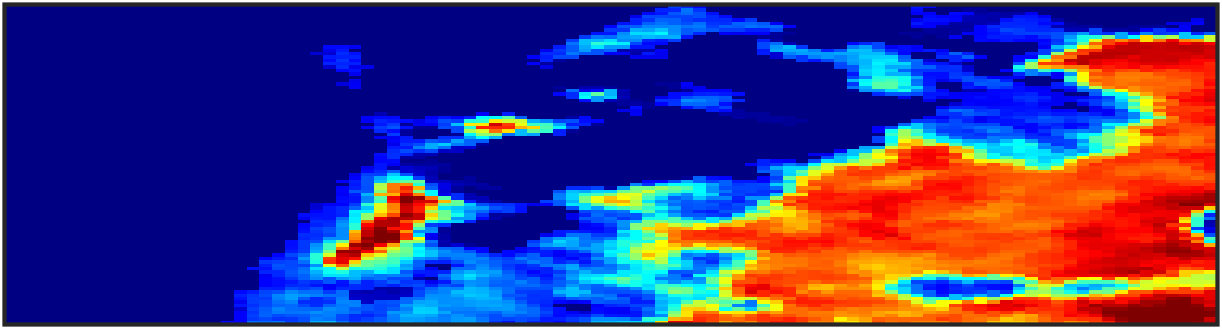}&
		\includegraphics[width=\picw\linewidth,height=0.065\textheight]{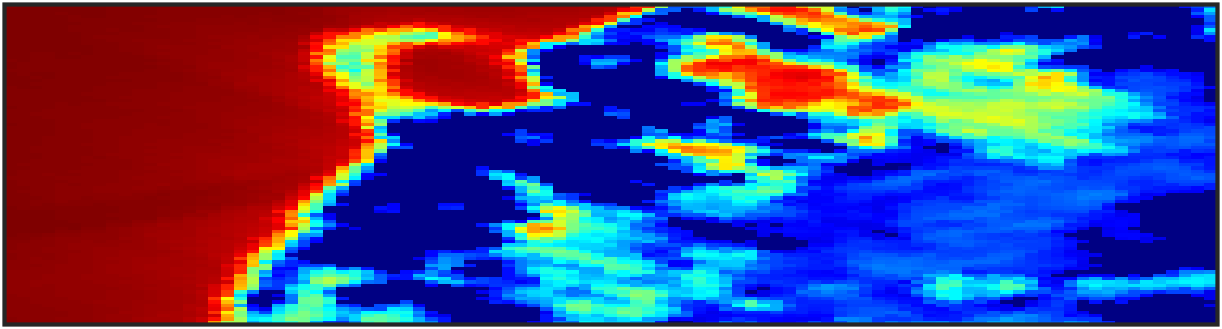}\\

	\raisebox{1.0\normalbaselineskip}[0pt][0pt]{\rotatebox{90}{\footnotesize \shortstack{SWAG - \\TL1}}}&
        \includegraphics[width=\picw\linewidth,height=0.065\textheight]{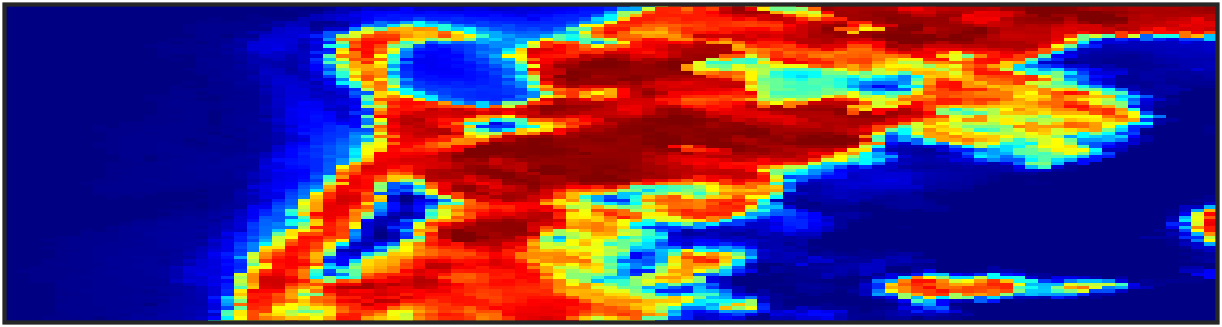}&
		\includegraphics[width=\picw\linewidth,height=0.065\textheight]{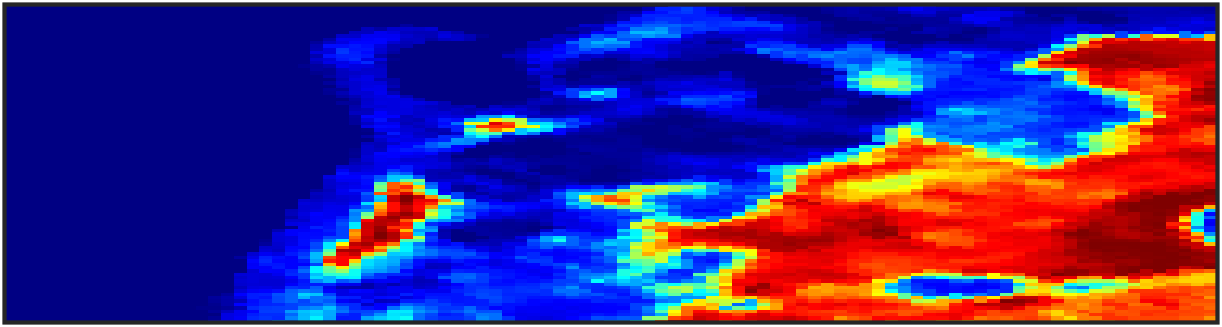}&
		\includegraphics[width=\picw\linewidth,height=0.065\textheight]{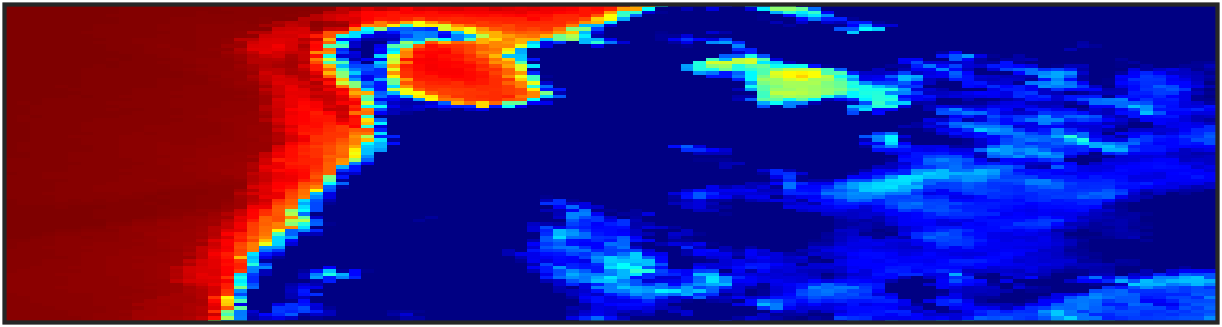}\\
        \begin{comment}
	
		\raisebox{1.0\normalbaselineskip}[0pt][0pt]{\rotatebox{90}{\footnotesize \shortstack{SWAG -\\ L12}}}&
        \includegraphics[width=\picw\linewidth,height=0.065\textheight]{samson/A_L12_im_Material1.png}&
		\includegraphics[width=\picw\linewidth,height=0.065\textheight]{samson/A_L12_im_Material2.png}&
		\includegraphics[width=\picw\linewidth,height=0.065\textheight]{samson/A_L12_im_Material3.png}\\

		\raisebox{1.0\normalbaselineskip}[0pt][0pt]{\rotatebox{90}{\footnotesize \shortstack{SWAG - \\Lq}}}&
        \includegraphics[width=\picw\linewidth,height=0.065\textheight]{samson/A_LP_im_Material1.png}&
		\includegraphics[width=\picw\linewidth,height=0.065\textheight]{samson/A_LP_im_Material2.png}&
		\includegraphics[width=\picw\linewidth,height=0.065\textheight]{samson/A_LP_im_Material3.png}\\

   \end{comment}
	
	\end{tabular}
	 \caption{Abundance maps estimated by the competing methods on 
the Samson dataset with the ground truth (top row).}
    \label{fig:samson_abundance}
\end{figure}

\medskip

\subsubsection{Jasper Ridge}
The Jasper Ridge dataset \citep{jasper} is one of the most widely used benchmarks for hyperspectral image classification. The dataset includes four latent endmembers: vegetation, water, soil, and road. 
%\yl{[make sure to capitalize the first letter of the endmember classes]}
 Its spatial dimension is $512 \times 614$ with 198 channels\footnote{Note that the original Jasper Ridge data contains 224 hyperspectral channels ranging from 380 to 2500 nm. However, it is common practice to remove 26 channels during preprocessing to account for dense water vapor and atmospheric effects.}. For our experiments, we consider a subimage of $100 \times 100$ pixels with 198 channels. The bundle matrix is obtained by the AEB method with a total of $40$ signatures.
We use the ground truth information provided on the website \citep{jasper} to compute the RMSE between the estimated and the true abundance values and record them in Table \ref{tab:jasper}, showing that SWAG-TL1 achieves the best result among the competing methods. In Figure \ref{fig:jasper_abundance}, we show the abundance plot of the four materials with the ground truth in the first row. From the plot, it is evident that none of the methods effectively detect the road in the image. However, SWAG methods produce the best approximation of the water abundance. %\yl{[Fig?]} \gokul{done}

\begin{table*}[!ht]
\centering
\resizebox{\textwidth}{!}{
\begin{tabular}{ |p{1.4cm}|p{1.0cm}|p{1.0cm}|p{1.0 cm}|p{1.0 cm}|p{1.4 cm}|p{1.0 cm}|}

\hline
\multicolumn{7}{|c|}{Jasper Ridge  } \\
\hline
Methods & FCLS & Inter-L1 & Intra-L1 & SWAG-Lq & Inter-TL1 & SWAG-TL1 \\%& SWAG-L12 \\%& & SWAG-Lq\\
\hline
RMSE$(\widehat{\bm{M}})$ &  \textcolor{black}{0.152} &0.156  & 0.203  &\textcolor{blue}{0.114}  & 0.156& \textcolor{red}{0.113}\\%& \textcolor{blue}{0.1984}\\%& &0.2011\\
%RMSE(E) &   &     &    & & & &\\
SAM$(\widehat{\bm{S}})$ & --  &  \textcolor{black}{15.122} & \textcolor{red}{14.158}  & 14.722  & 15.104 &\textcolor{blue}{14.396}\\%&14.5866\\
RMSE$(\widehat{\bm{X}})$ & 0.040  & 0.025   &\textcolor{black}{0.018}  & \textcolor{blue}{0.014} &\textcolor{black}{0.025} &\textcolor{red}{0.013}\\%& 0.01577\\%&&0.0153\\

 Time(s) & 0.74  & 14.09 & 24.34 & 31.42 &76.28 &35.27 \\%& 29.006\\%&&44.25\\

\hline
\end{tabular} 
}
\caption{Comparison of the Jasper Ridge data with the best value in red and the second best in blue. Since FCLS does not output the mixing matrix, we indicate its error on $\widehat{\bm S}$ with a dash (--).}
\label{tab:jasper}
\end{table*}

\begin{comment}
    \begin{figure}[!ht]
    \centering
    \includegraphics[width=1.0\linewidth, height=0.5\textheight]{figures/jasper_abundance.eps}
    \caption{Abundance maps estimated by the tested methods with the ground truth (top row).}
    \label{fig:samson_abundance}
\end{figure}
\end{comment}

\def\picw{0.22}
\begin{figure}
\centering
\setlength{\tabcolsep}{1pt}
	\begin{tabular}{ccccc}
&Vegetation & Water & Road& Soil\\
		\raisebox{0.5\normalbaselineskip}[0pt][0pt]{\rotatebox{90}{\footnotesize Reference}}&
        \includegraphics[width=\picw\linewidth, height=0.065\textheight]{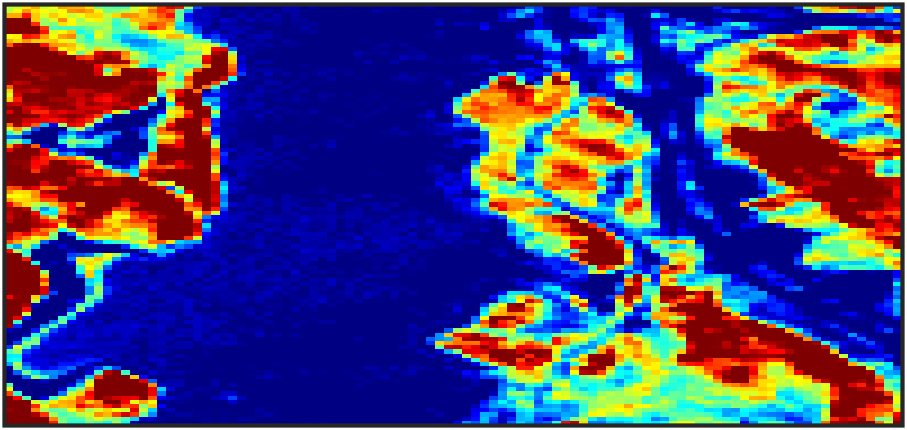}&
		\includegraphics[width=\picw\linewidth, height=0.065\textheight]{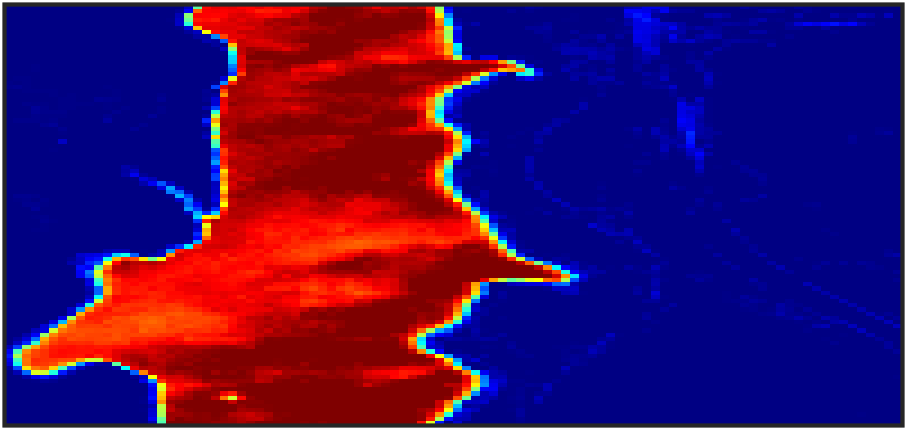}&
        \includegraphics[width=\picw\linewidth, height=0.065\textheight]{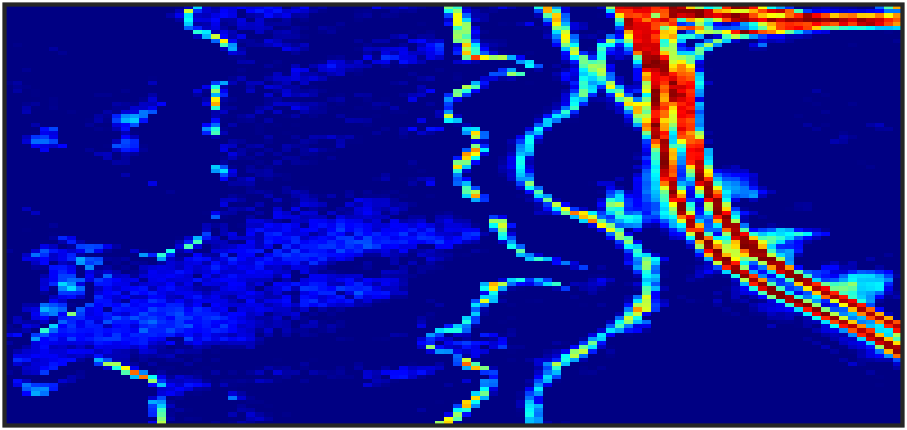}&
        \includegraphics[width=\picw\linewidth, height=0.065\textheight]{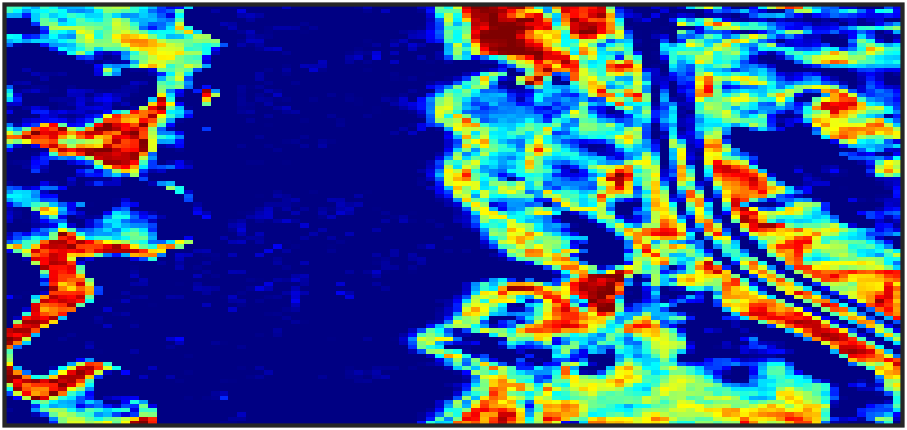}\\

		\raisebox{1.0\normalbaselineskip}[0pt][0pt]{\rotatebox{90}{\footnotesize FCLS}}&
        \includegraphics[width=\picw\linewidth,height=0.065\textheight,]{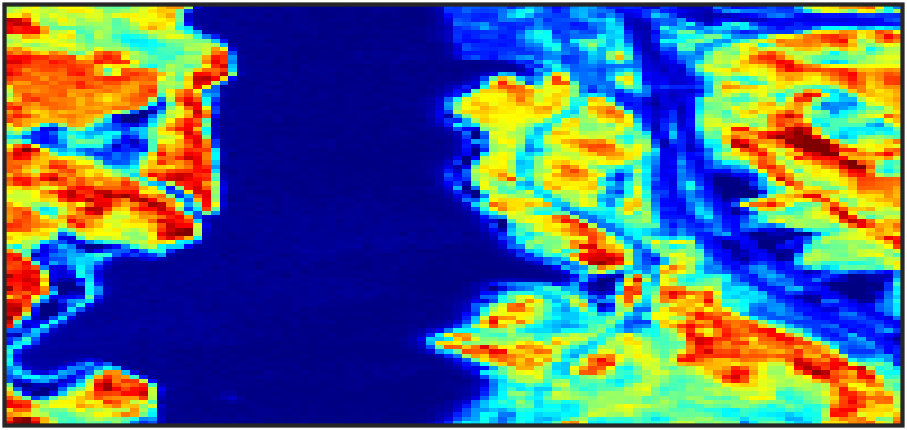}&
		\includegraphics[width=\picw\linewidth,height=0.065\textheight]{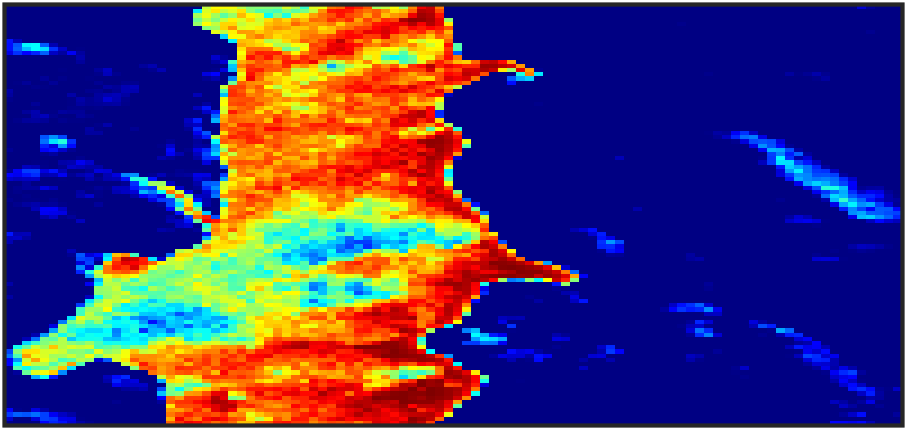}&
          \includegraphics[width=\picw\linewidth,height=0.065\textheight,]{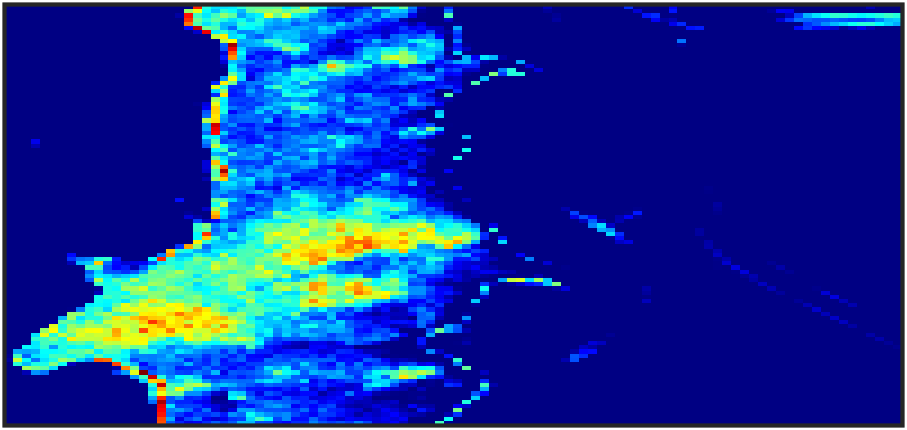}&
            \includegraphics[width=\picw\linewidth,height=0.065\textheight,]{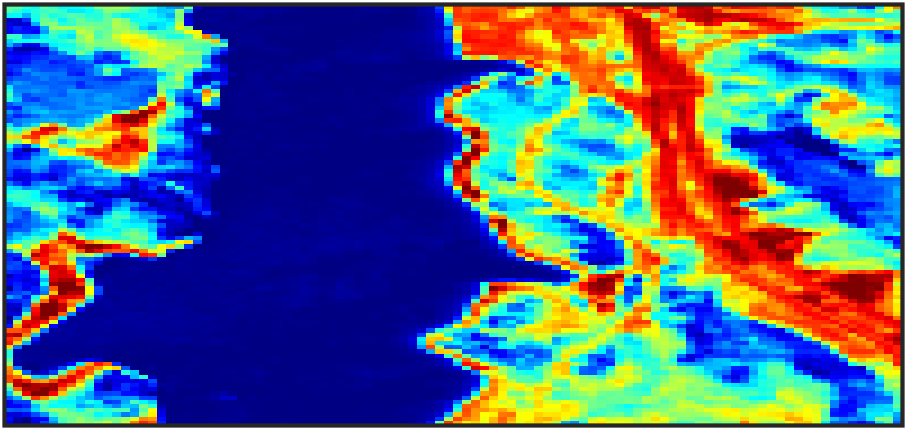}\\
            
            \raisebox{1.0\normalbaselineskip}[0pt][0pt]{\rotatebox{90}{\footnotesize \shortstack{Inter - \\L1}}}&
        \includegraphics[width=\picw\linewidth,height=0.065\textheight]{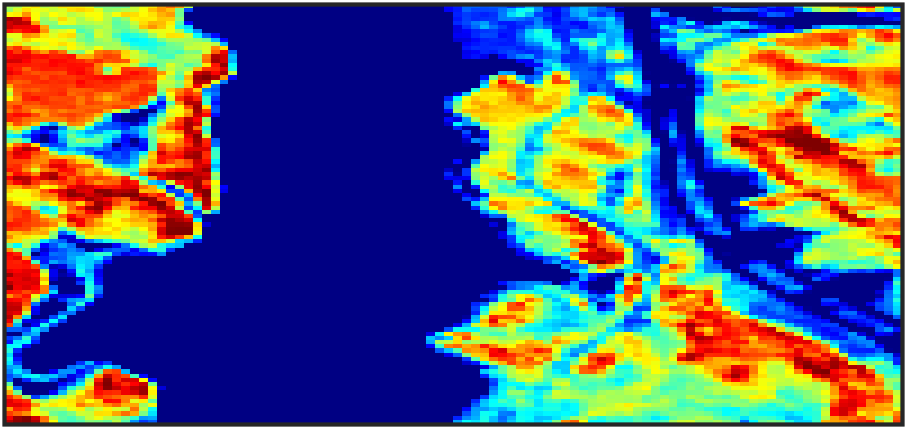}&
		\includegraphics[width=\picw\linewidth,height=0.065\textheight]{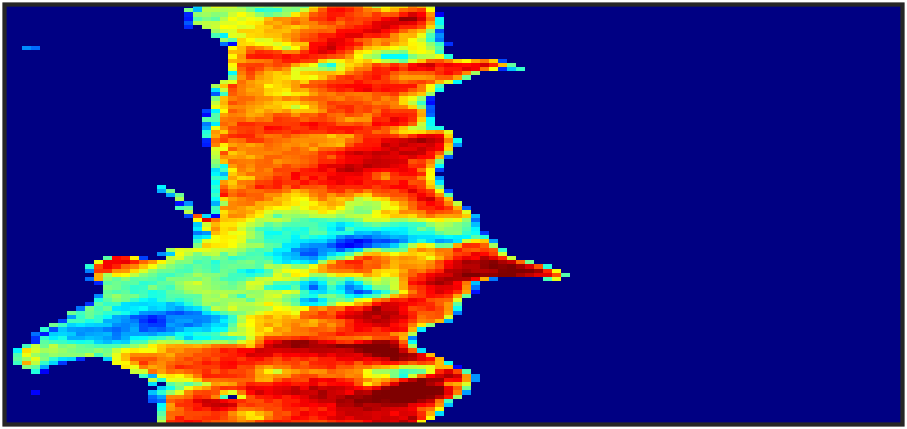}&
        \includegraphics[width=\picw\linewidth,height=0.065\textheight]{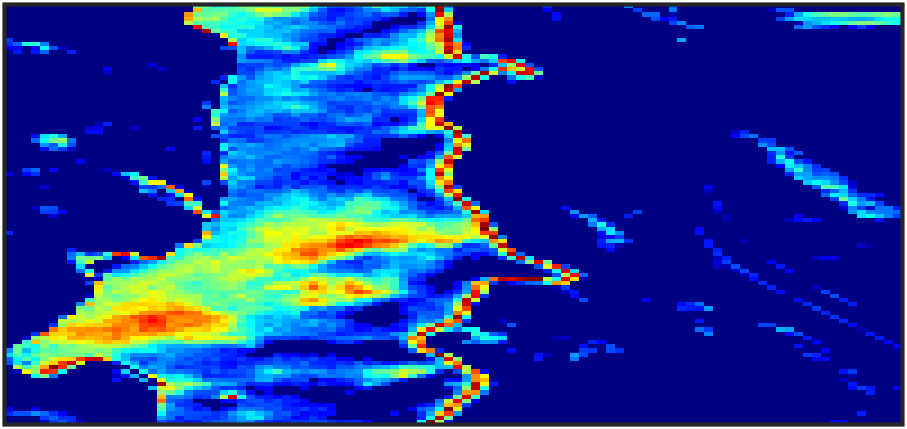}&
        \includegraphics[width=\picw\linewidth,height=0.065\textheight]{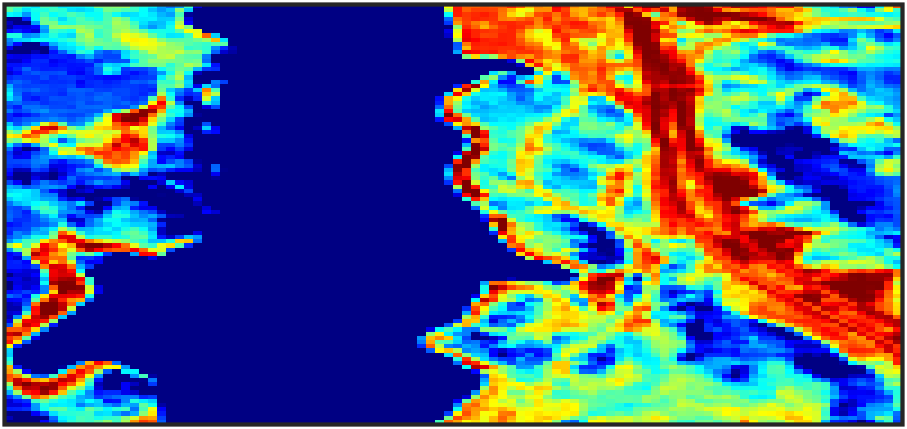}\\

\raisebox{1.0\normalbaselineskip}[0pt][0pt]{\rotatebox{90}{\footnotesize \shortstack{Intra - \\ L1}}}&
        \includegraphics[width=\picw\linewidth,height=0.065\textheight]{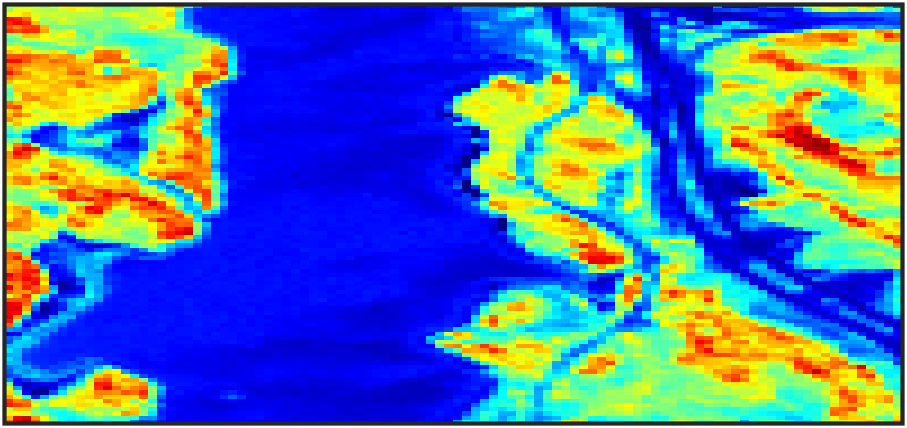}&
		\includegraphics[width=\picw\linewidth,height=0.065\textheight]{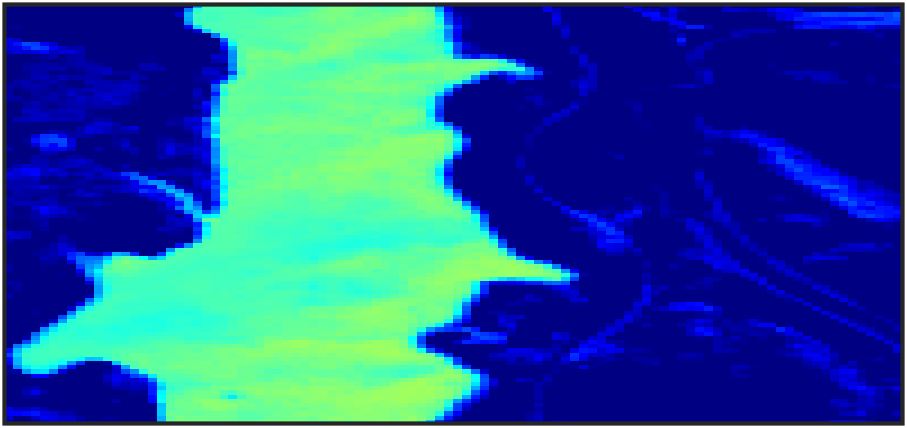}&
        \includegraphics[width=\picw\linewidth,height=0.065\textheight]{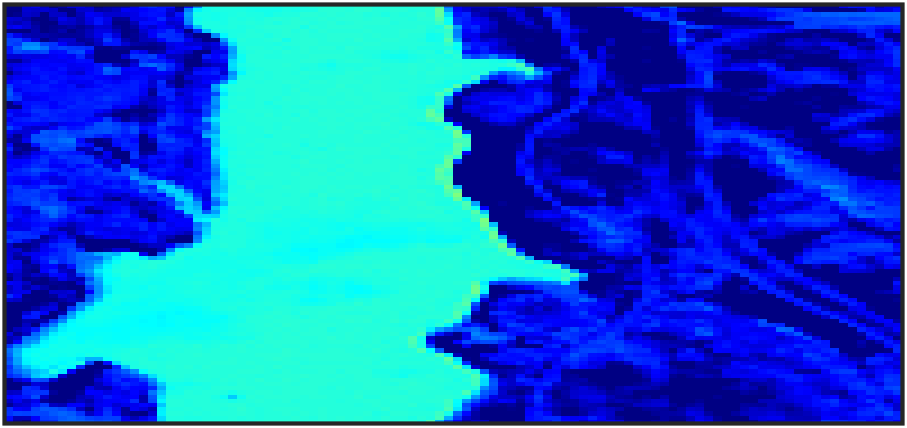}&
        \includegraphics[width=\picw\linewidth,height=0.065\textheight]{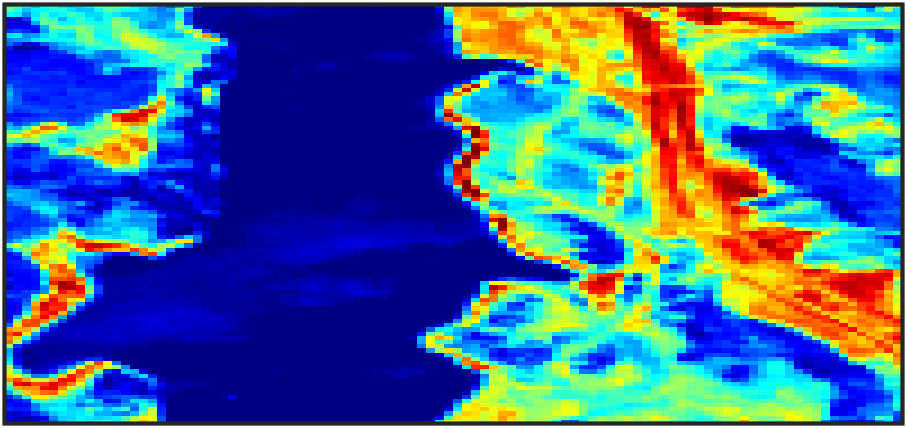}\\

	\raisebox{0.5\normalbaselineskip}[0pt][0pt]{\rotatebox{90}{\footnotesize \shortstack{SWAG - \\Lq}}}&
        \includegraphics[width=\picw\linewidth,,height=0.065\textheight]{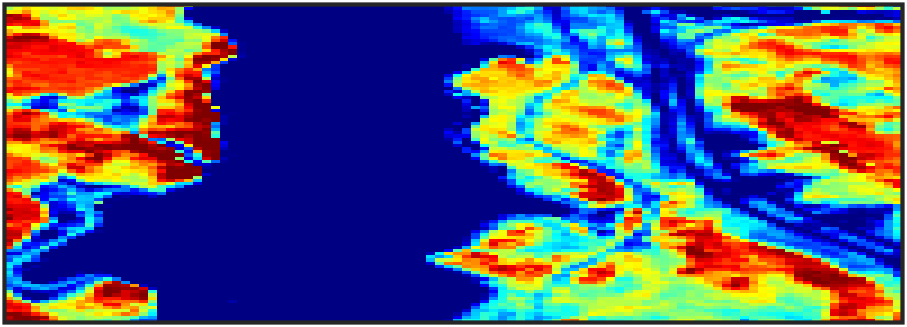}&
		\includegraphics[width=\picw\linewidth,height=0.065\textheight]{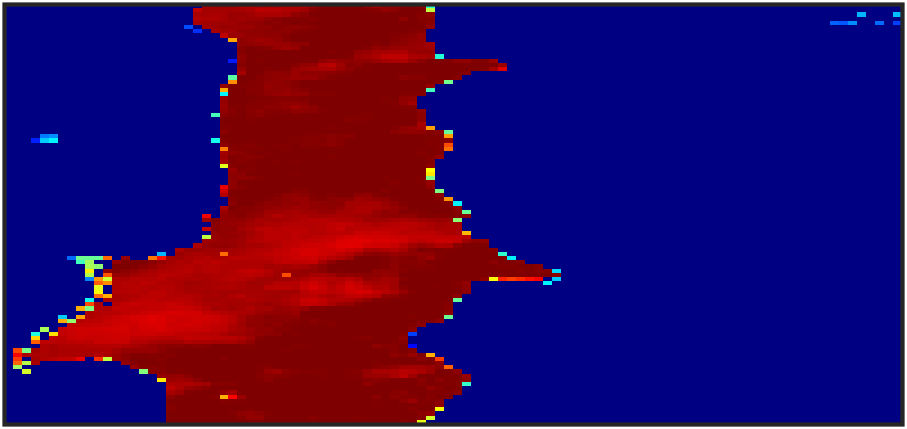}&
        \includegraphics[width=\picw\linewidth,height=0.065\textheight]{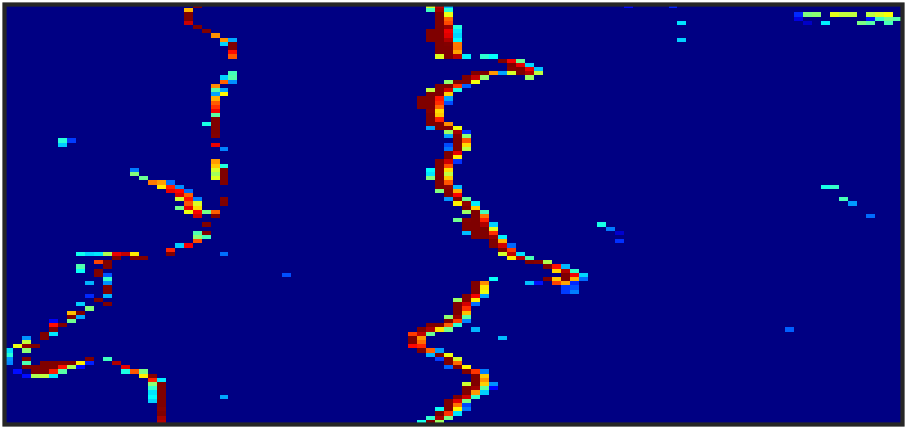}&
        \includegraphics[width=\picw\linewidth,height=0.065\textheight]{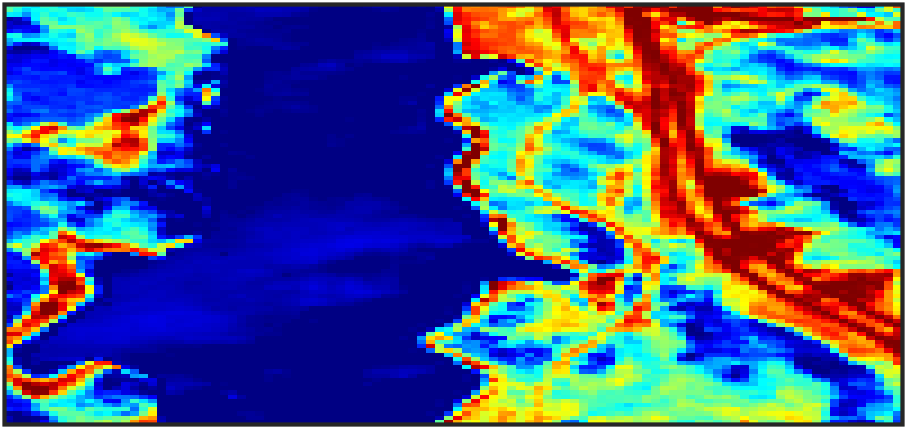}\\

		\raisebox{1.0\normalbaselineskip}[0pt][0pt]{\rotatebox{90}{\footnotesize \shortstack{Inter - \\ TL1}}}&
        \includegraphics[width=\picw\linewidth,height=0.065\textheight]{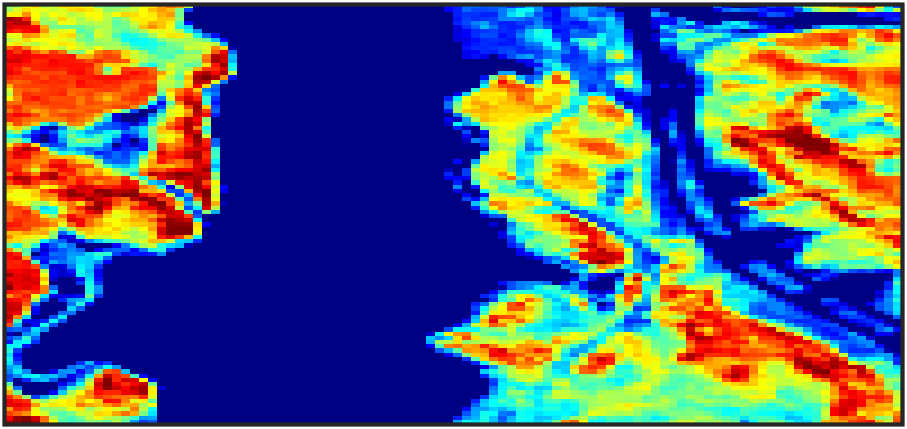}&
		\includegraphics[width=\picw\linewidth,height=0.065\textheight]{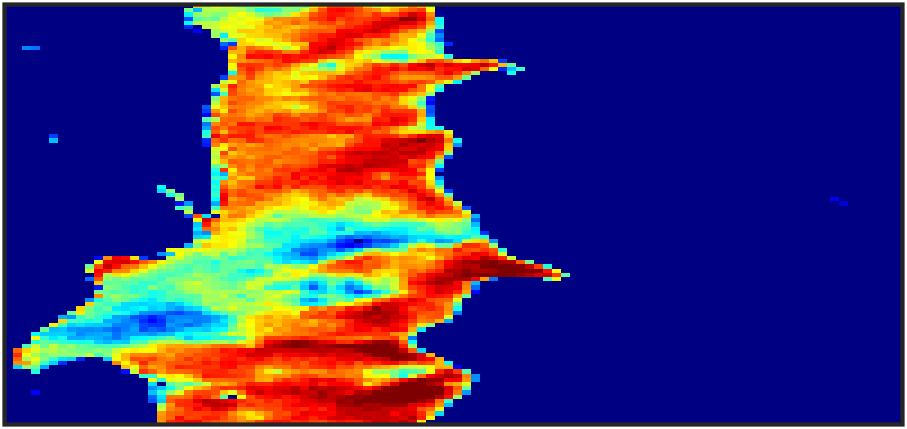}&
        \includegraphics[width=\picw\linewidth,height=0.065\textheight]{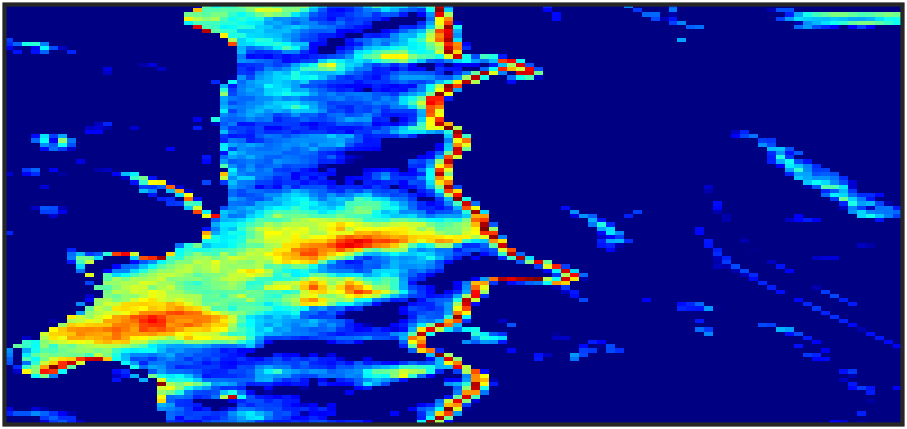}&
        \includegraphics[width=\picw\linewidth,height=0.065\textheight]{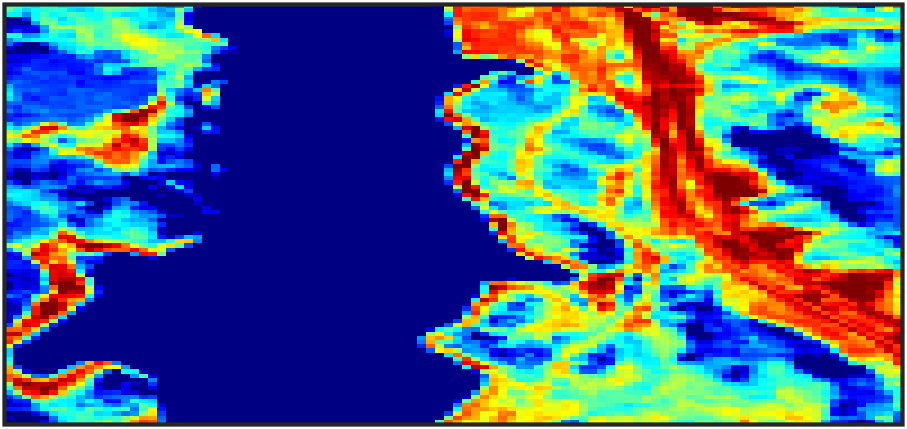}\\

		  \raisebox{1.0\normalbaselineskip}[0pt][0pt]{\rotatebox{90}{\footnotesize \shortstack{SWAG - \\TL1}}}&
        \includegraphics[width=\picw\linewidth,height=0.065\textheight]{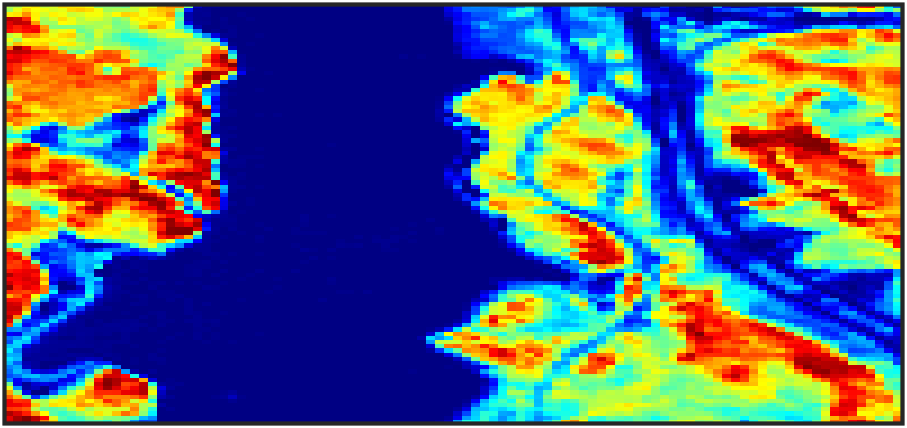}&
		\includegraphics[width=\picw\linewidth,height=0.065\textheight]{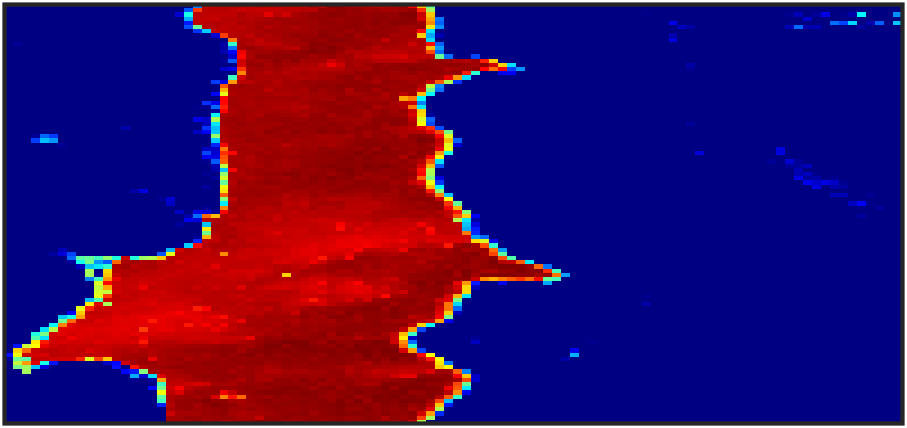}&
        	\includegraphics[width=\picw\linewidth,height=0.065\textheight]{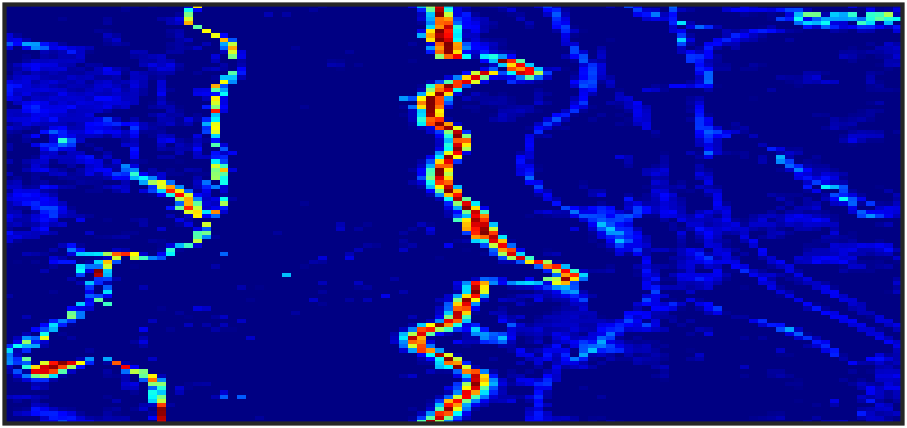}&
            	\includegraphics[width=\picw\linewidth,height=0.065\textheight]{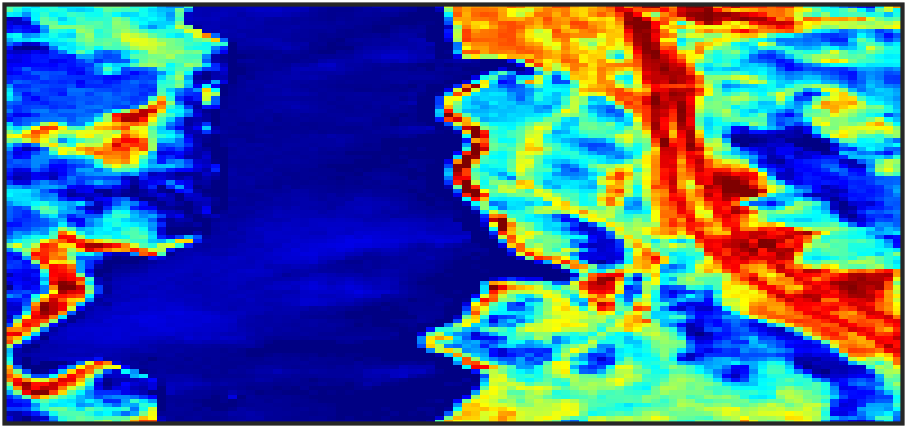}\\

	\begin{comment}
		\raisebox{1.0\normalbaselineskip}[0pt][0pt]{\rotatebox{90}{\footnotesize \shortstack{SWAG -\\ L12}}}&
        \includegraphics[width=\picw\linewidth,height=0.065\textheight]{jasper/A_L12_im_Material1.png}&
		\includegraphics[width=\picw\linewidth,height=0.065\textheight]{jasper/A_L12_im_Material2.png}&
        	\includegraphics[width=\picw\linewidth,height=0.065\textheight]{jasper/A_L12_im_Material3.png}&
            	\includegraphics[width=\picw\linewidth,height=0.065\textheight]{jasper/A_L12_im_Material4.png}\\

		\raisebox{1.0\normalbaselineskip}[0pt][0pt]{\rotatebox{90}{\footnotesize \shortstack{SWAG - \\Lq}}}&
        \includegraphics[width=\picw\linewidth,height=0.065\textheight]{jasper/A_LP_im_Material1.png}&
		\includegraphics[width=\picw\linewidth,height=0.065\textheight]{jasper/A_LP_im_Material2.png}&
        	\includegraphics[width=\picw\linewidth,height=0.065\textheight]{jasper/A_LP_im_Material3.png}&
            	\includegraphics[width=\picw\linewidth,height=0.065\textheight]{jasper/A_LP_im_Material4.png}\\
	
	\end{comment}

	\end{tabular}
	 \caption{Abundance maps estimated by the competing methods on the Jasper Ridge data with the ground truth (top row).}
    \label{fig:jasper_abundance}
\end{figure}

\medskip
\subsubsection{Houston Dataset}
Lastly, we use the subset of an image acquired over the Robertson stadium of the campus of the University of Houston and its surroundings, referred to as the Houston data. It was used for the 2013 IEEE GRSS Data Fusion Contest (DCF) \citep{debes2014hyperspectral}, but it does not contain the ground truth abundance map for hyperspectral unmixing.
The subset of the image we use in experiments is of size $152 \times 108 \times 144$. 
%The Houston data doesn't have a reference library of endmembers available. 
The bundle matrix containes  a total of $40$ signatures. 

Since the ground-truth abundance is unavailable, we only report the RMSE and SAM between the reconstructed image and the true image. The results are listed in Table \ref{tab:houston_data}, showing that SWAG-TL1 achieves significant improvements over the others. SWAG with either Lq or TL1 outperforms other types of group sparsity.
Furthermore, we visualize the abundance maps obtained from each method in Figure \ref{fig:houston_abundance}. From the plot, one can see that all methods seem to detect the materials present in the image. The SWAG-Lq penalty seems to produce sparser abundance maps compared to the other methods. %We also compute the RMSE and SAM between the reconstruction of the pixels and the true pixel values, and record in Table \ref{tab:houston_data}. Inter-TL1 and SWAG-TL1 have the best values compared to other methods. We also record the computational time of each method in the Table \ref{tab:houston_data}. 

%observation: swag favours big bundle matrix.  

\begin{table*}[!ht]
\centering
\label{tab:houston}
\resizebox{\textwidth}{!}{
\begin{tabular}{ |p{1.4cm}|p{1.0cm}|p{1.0 cm}|p{1.0 cm}|p{1.0 cm}|p{1.4 cm}|p{1.0 cm}|}

\hline
\multicolumn{7}{|c|}{Houston Data } \\
\hline
Methods & FCLS &  Inter-L1 & Intra-L1 & SWAG-Lq & Inter-TL1 & SWAG-TL1 \\%& SWAG-L12 \\%&& SWAG-Lq\\
\hline
%RMSE(A) &  \textcolor{red}{-} & \textcolor{blue}{ -} & -  & - &   \textcolor{blue}{ -} & & & &\\
%RMSE(E) &  \textcolor{red}{-} & \textcolor{blue}{ -} & -  & - &   \textcolor{blue}{ -} & & & &\\
RMSE$(\widehat{\bm{X}})$ & 0.021 & 0.010 & 0.017 & \textcolor{blue}{0.086} &   \textcolor{black}{0.009}& \textcolor{red}{0.005}\\%& 0.0123 \\ %&& 0.0121 \\
SAM$(\widehat{\bm{X}})$ & 9.486  &  2.697
 & 5.502 & \textcolor{blue}{2.503} &  \textcolor{black}{ 2.517}& \textcolor{red}{1.452}\\% & 3.3556\\%& &  2.9595
 
Time(s) & 1.15 & 3.49 & 25.49 & 38.17 & 29.71& 44.20\\%&40.44 \\%& &44.12 \\
%$\lambda$ & $\times$  & 1 & 2  & 0.4 & 0.04\\
\hline
\end{tabular} 
}
\caption{Comparison of the Houston data with the best value in red and the second best in blue.}
\label{tab:houston_data}
\end{table*}

\begin{comment}
    \begin{figure}[!ht]
    \centering
    \includegraphics[width=1.0\linewidth, height=0.5\textheight]{figures/houston_abundance.eps}
    \caption{Abundance maps estimated by the tested methods. }
    \label{fig:houston_abundance}
\end{figure}
\end{comment}
\def\picw{0.22}
\begin{figure}
\centering
\setlength{\tabcolsep}{1pt}
	\begin{tabular}{ccccc}
&Vegetation  & Asphalt& Metal Roofs & Concrete\\

		\raisebox{1.0\normalbaselineskip}[0pt][0pt]{\rotatebox{90}{\footnotesize FCLS}}&
        \includegraphics[width=\picw\linewidth,height=0.065\textheight,]{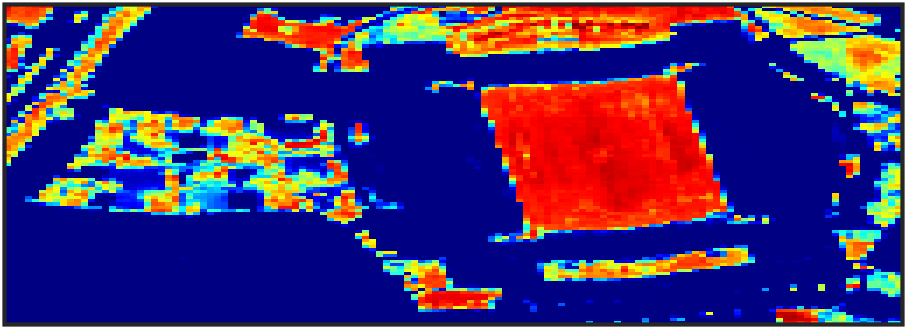}&
		\includegraphics[width=\picw\linewidth,height=0.065\textheight]{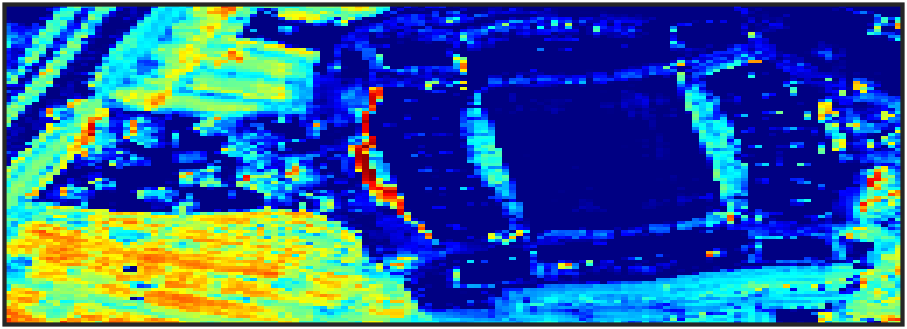}&
        \includegraphics[width=\picw\linewidth,height=0.065\textheight]{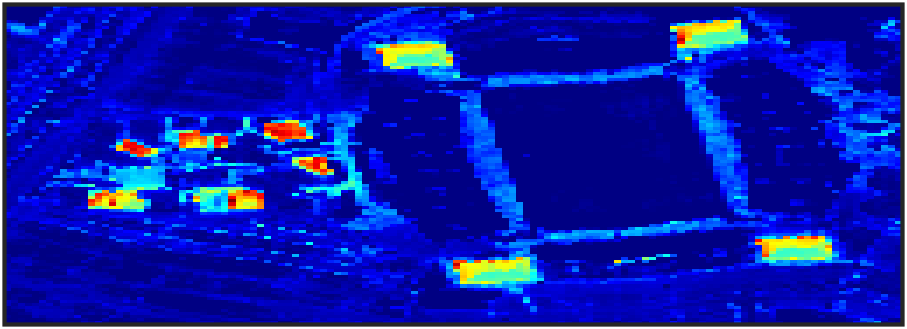}&
        \includegraphics[width=\picw\linewidth, height=0.065\textheight]{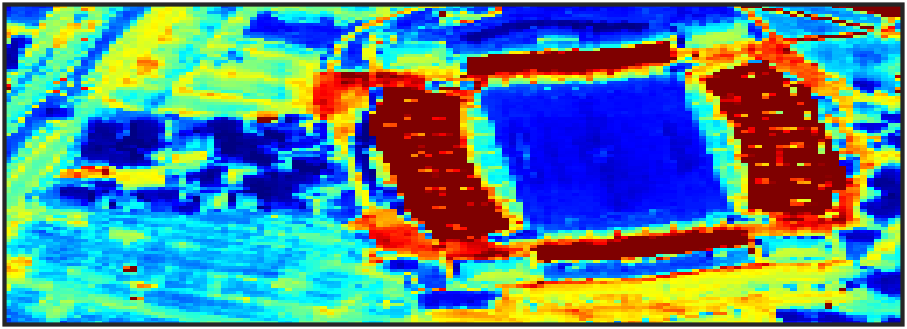}\\

        	\raisebox{1.0\normalbaselineskip}[0pt][0pt]{\rotatebox{90}{\footnotesize \shortstack{Inter - \\L1}}}&
        \includegraphics[width=\picw\linewidth,height=0.065\textheight]{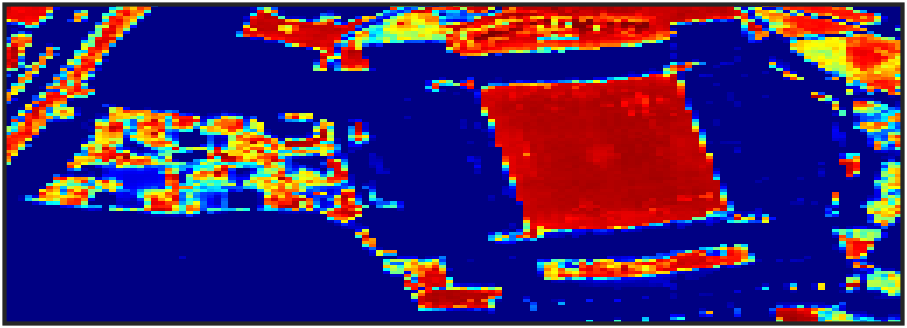}&
		\includegraphics[width=\picw\linewidth,height=0.065\textheight]{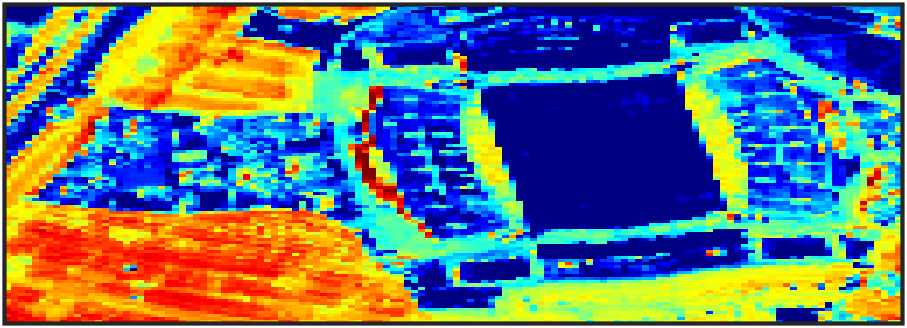}&
		\includegraphics[width=\picw\linewidth,height=0.065\textheight]{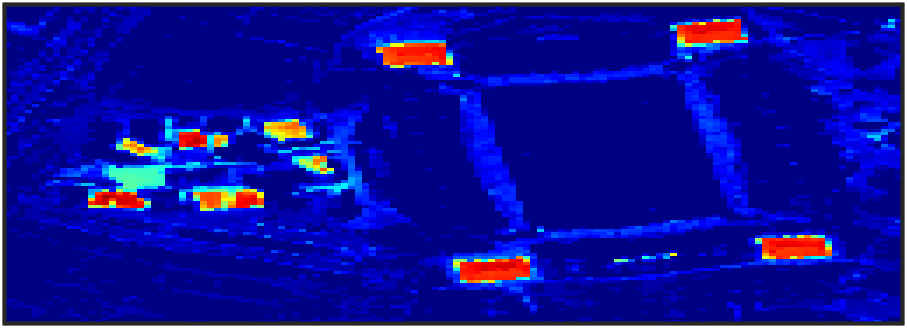}&
        \includegraphics[width=\picw\linewidth, height=0.065\textheight]{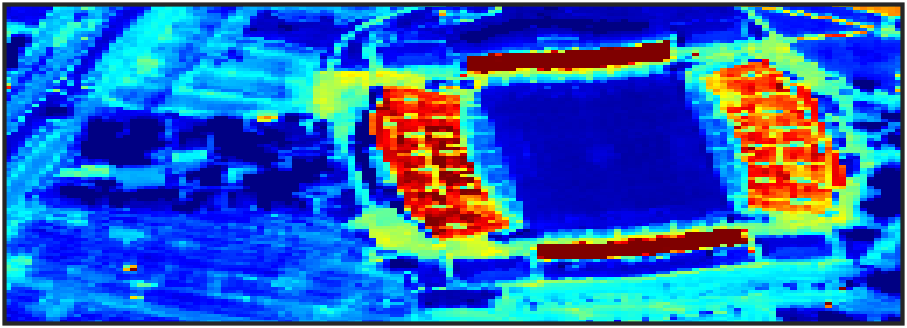}\\

        \raisebox{1.0\normalbaselineskip}[0pt][0pt]{\rotatebox{90}{\footnotesize \shortstack{Intra - \\ L1}}}&
        \includegraphics[width=\picw\linewidth,height=0.065\textheight]{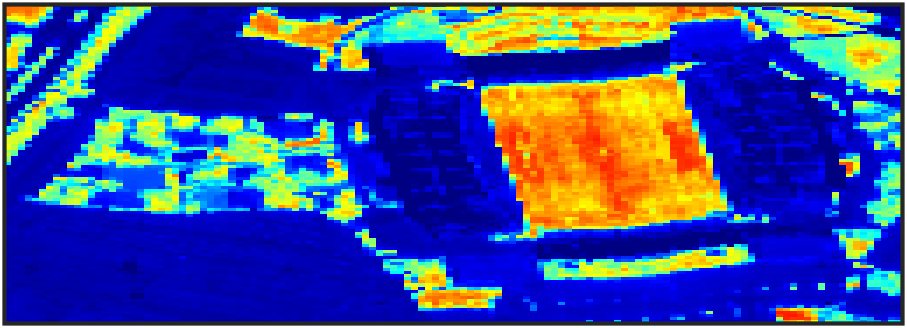}&
		\includegraphics[width=\picw\linewidth,height=0.065\textheight]{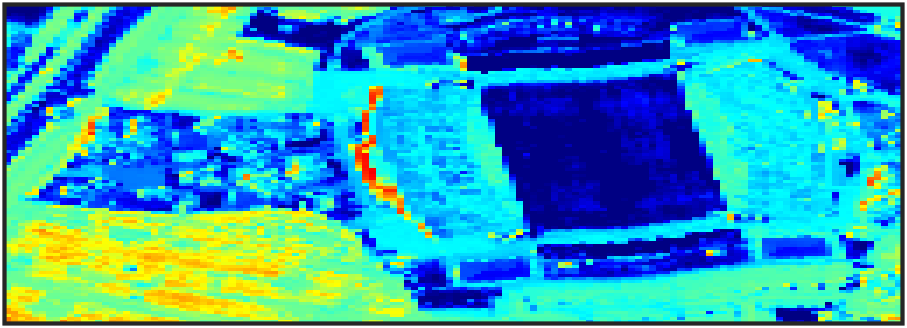}&
        \includegraphics[width=\picw\linewidth, height=0.065\textheight]{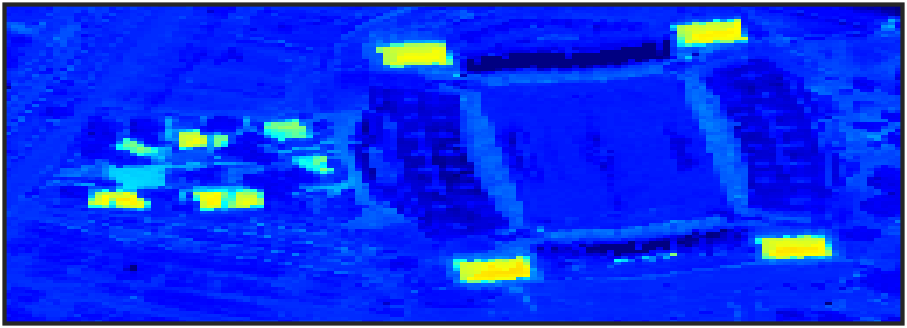}&
		\includegraphics[width=\picw\linewidth,height=0.065\textheight]{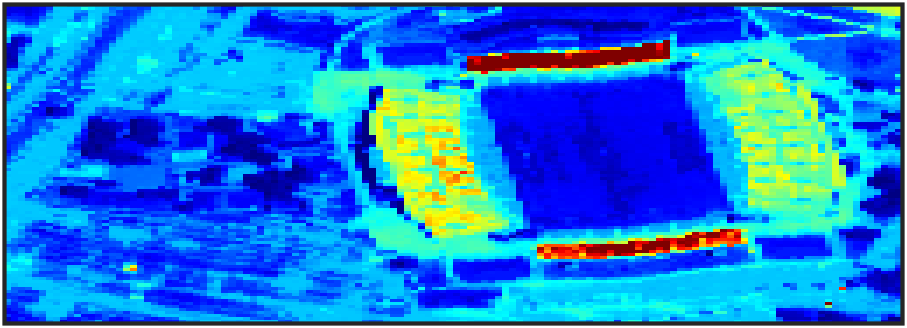}\\
	        
		\raisebox{0.5\normalbaselineskip}[0pt][0pt]{\rotatebox{90}{\footnotesize \shortstack{SWAG - \\Lq}}}&
        \includegraphics[width=\picw\linewidth,,height=0.065\textheight]{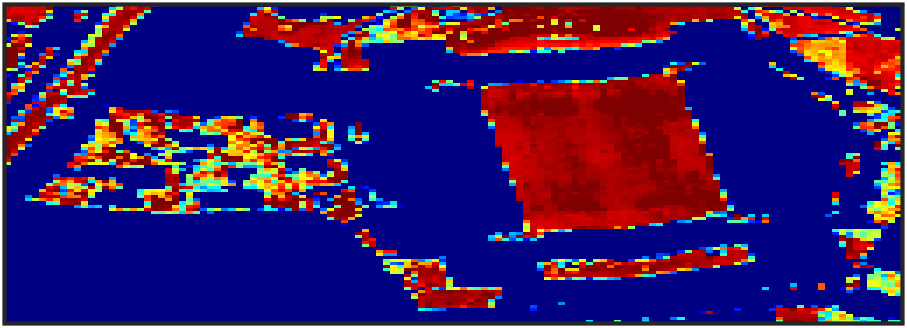}&
		\includegraphics[width=\picw\linewidth,height=0.065\textheight]{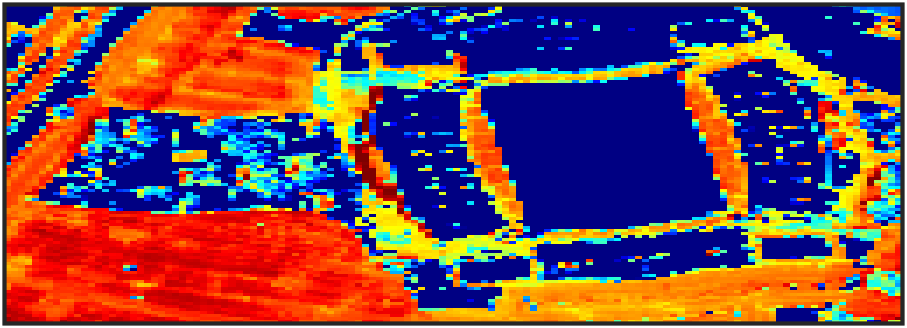}&
		\includegraphics[width=\picw\linewidth,height=0.065\textheight]{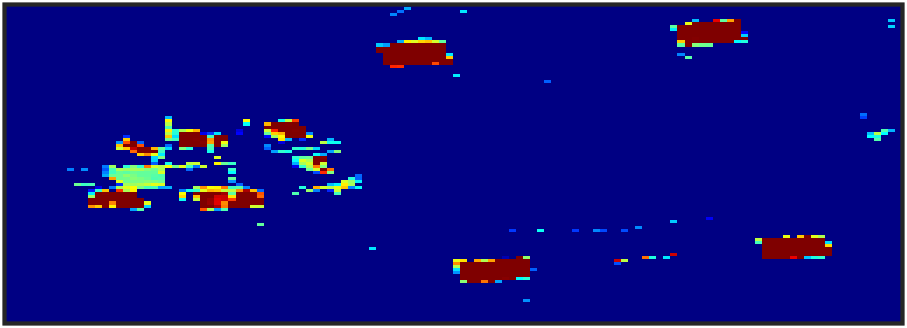}&
        \includegraphics[width=\picw\linewidth, height=0.065\textheight]{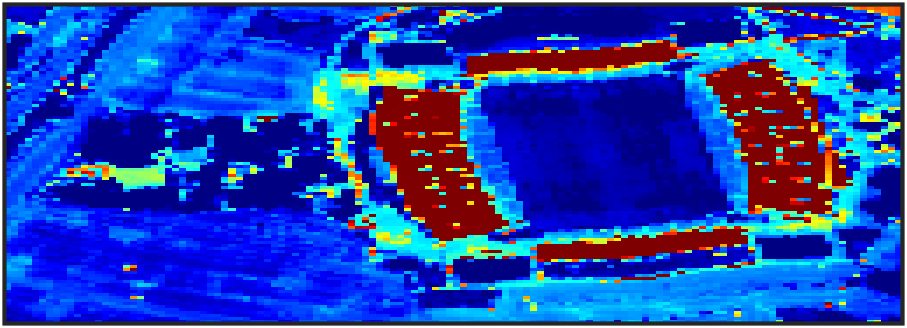}\\

        		\raisebox{1.0\normalbaselineskip}[0pt][0pt]{\rotatebox{90}{\footnotesize \shortstack{Inter -\\ TL1}}}&
        \includegraphics[width=\picw\linewidth,height=0.065\textheight]{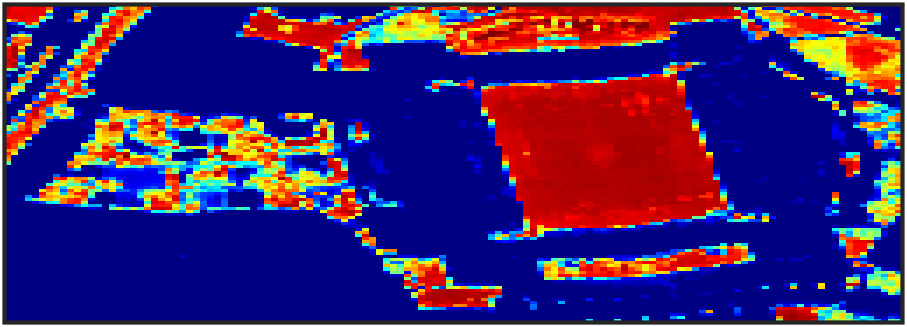}&
		\includegraphics[width=\picw\linewidth,height=0.065\textheight]{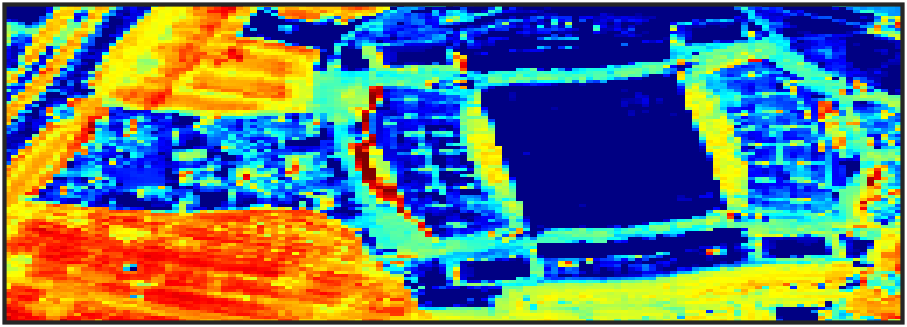}&
		\includegraphics[width=\picw\linewidth,height=0.065\textheight]{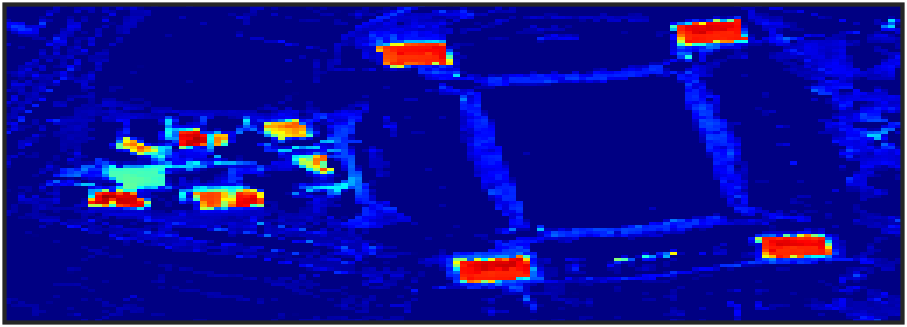}&
        \includegraphics[width=\picw\linewidth, height=0.065\textheight]{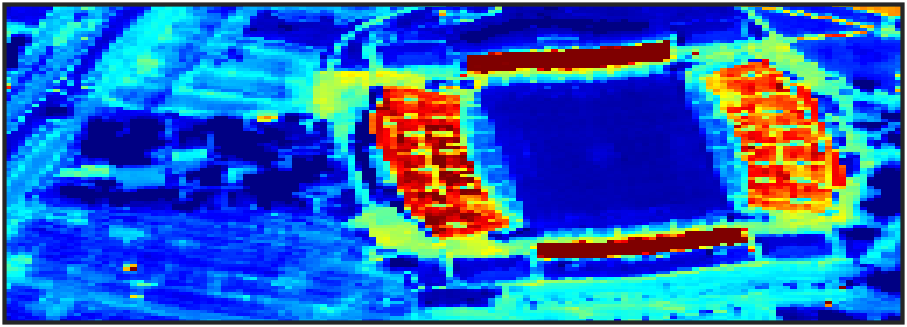}\\
		
		    \raisebox{1.0\normalbaselineskip}[0pt][0pt]{\rotatebox{90}{\footnotesize \shortstack{SWAG - \\TL1}}}&
        \includegraphics[width=\picw\linewidth,height=0.065\textheight]{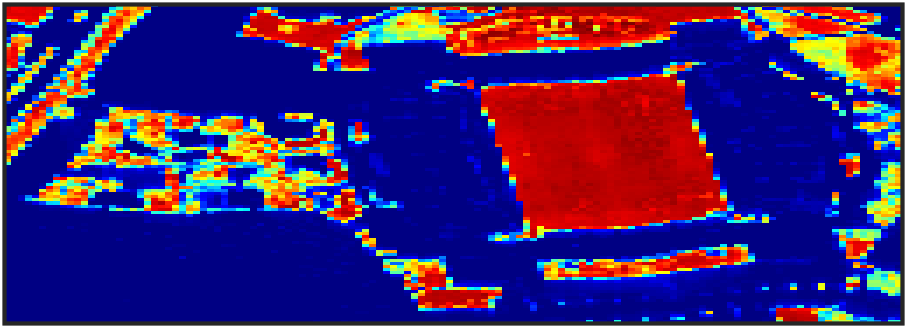}&
		\includegraphics[width=\picw\linewidth,height=0.065\textheight]{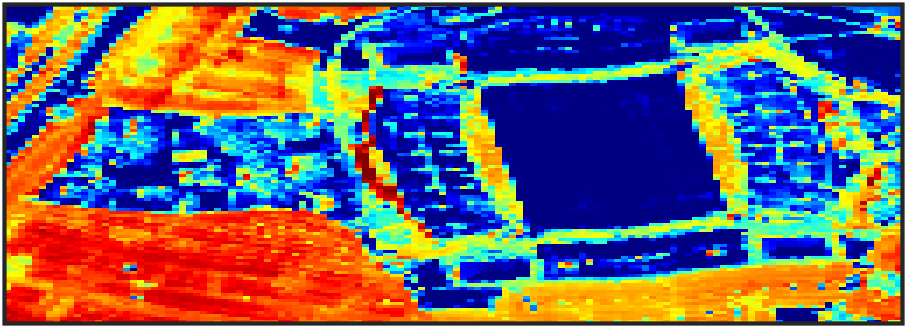}&
		\includegraphics[width=\picw\linewidth,height=0.065\textheight]{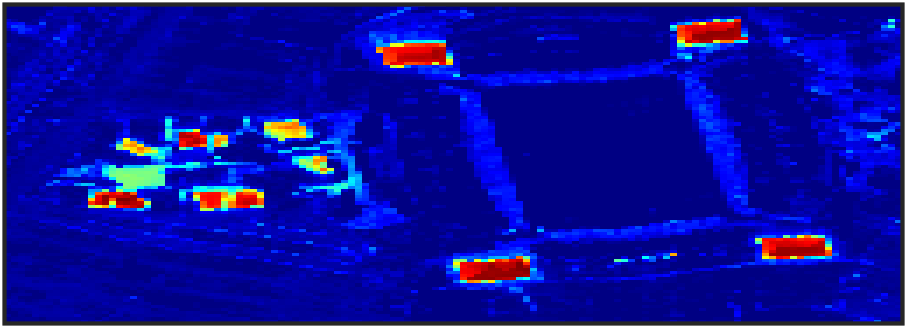}&
        \includegraphics[width=\picw\linewidth, height=0.065\textheight]{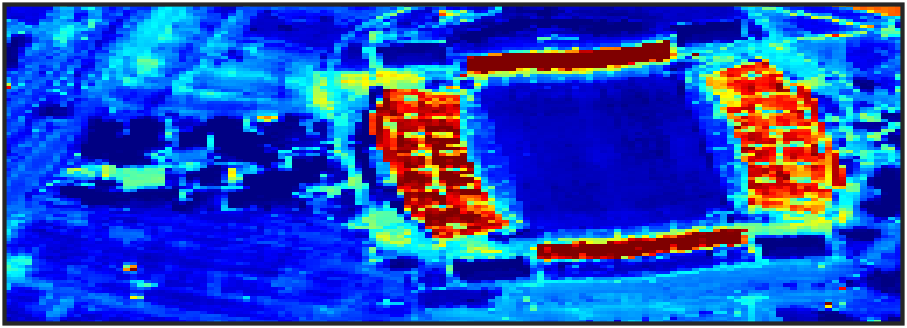}\\

	\begin{comment}
		\raisebox{1.0\normalbaselineskip}[0pt][0pt]{\rotatebox{90}{\footnotesize \shortstack{SWAG -\\ L12}}}&
        \includegraphics[width=\picw\linewidth,height=0.065\textheight]{houston/A_L12_im_Material1.png}&
		\includegraphics[width=\picw\linewidth,height=0.065\textheight]{houston/A_L12_im_Material2.png}&
		\includegraphics[width=\picw\linewidth,height=0.065\textheight]{houston/A_L12_im_Material3.png}&
        \includegraphics[width=\picw\linewidth, height=0.065\textheight]{houston/A_L12_im_Material4.png}\\

		\raisebox{1.0\normalbaselineskip}[0pt][0pt]{\rotatebox{90}{\footnotesize \shortstack{SWAG - \\Lq}}}&
        \includegraphics[width=\picw\linewidth,height=0.065\textheight]{houston/A_LP_im_Material1.png}&
		\includegraphics[width=\picw\linewidth,height=0.065\textheight]{houston/A_LP_im_Material2.png}&
		\includegraphics[width=\picw\linewidth,height=0.065\textheight]{houston/A_LP_im_Material3.png}&
        \includegraphics[width=\picw\linewidth, height=0.065\textheight]{houston/A_LP_im_Material4.png}\\
\end{comment}

	\end{tabular}
	 \caption{Abundance maps estimated by the competing methods on the Houston data.}
  \label{fig:houston_abundance}
\end{figure}

\section{Conclusion}
\label{conclusion}
In this article, we propose a general framework for incorporating group sparsity into the hyperspectral unmixing problem with endmember bundles, specifically designed to address endmember variability. In particular, the framework accommodates a variety of sparsity-promoting penalties to enforce either inter-group sparsity or SWAG. Moreover, we advocate the use of the transformed $\ell_1$ regularization, which performs particularly well for the bundle-based hyperspectral unmixing. We conduct extensive experiments on one synthetic dataset and three standard real-world datasets, and the results consistently demonstrate that SWAG is the most effective approach for addressing bundle-based spectral variability.
For future work, several promising directions can be explored and integrated into the proposed framework. First, in order to exploit the similarity of spectral information across different pixels, one can incorporate various graph-based regularization techniques \citep{qin2020blind} when solving \eqref{eq:general-bundle}. Second, one can develop a graph-based active learning framework \citep{chen2023graph, bhusal2024malady} to enhance performance even further while requiring minimal supervision. 

\appendix
\subsection{Proximal operators}\label{sect:prox-operators}
Given a univariate function $f:\mathbb R\to \mathbb R,$ the proximal mapping of $f$ with a positive parameter $\lambda$ is defined by 
%Recall that a proximal operator of a function $g(\cdot)$  is defined by
\[\prox_f(a;\lambda) \in \argmin_{v\in\mathbb R}\bigg(\lambda f(v) + \frac{1}{2}(v-a)^2\bigg),\]
for $a\in\mathbb R$. In the following, we review the proximal operators, corresponding to three sparsity-promoting regularizations, i.e., $\ell_1$, $\ell_{1/2},$ and TL1, used in the experiments. Since these three regularizations are separable---that is, they can be written as the sum of functions applied to each component---the proximal operator can be computed componentwise. With an abuse of notation, we denote the proximal operator applied to a vector $\bm a$ by $\prox_f(\bm a;\lambda)$.

\begin{itemize}
    \item The $\ell_1$ norm of a vector $\bm{a} $ is $\|\bm{a}\|_1= \sum_{i} |a_i|$, with its proximal operator is given by 
    \[\prox_{\ell_1}(\bm{a}, \lambda) = \text{sign}(\bm{a}) \circ \max(|\bm{a}| - \lambda , 0),\]
    where $\circ$ denotes the Hadamard operator for element-wise multiplication. 
    \item For a vector $\bm{a}$, the $\ell_{1/2}$ penalty is defined as $\|\bm{a}\|_{1/2} = \big(\sum_j \sqrt{|a_j|}\big)^2$. The proximal operator for $\ell_{1/2}^{1/2}$ has a closed-form solution \citep{xu2012l_half},
    \begin{equation}
    \label{prox_1/2}
    \begin{split}
    \prox_{\ell_{1/2}^{1/2}}(\bm{a};\lambda ) =&\frac{3\bm{a}}{4}\circ \bigg[\cos\bigg(\frac{\pi}{3}- \frac{\phi(\bm{a})}{3}\bigg)\bigg]^2 \\
    &\circ  \max\bigg(\bm{a}-\frac{3}{4} \lambda^{2/3},0 \bigg), 
    \end{split}
 \end{equation}
 where $\phi(\bm{a})= \arccos(\frac{\lambda}{8}(\frac{3}{|\bm{a}|})^{3/2}) $ with element-wise division. 
 %One can refer to \citep{chartrand2008iteratively, } for exhaustive references for $\ell_{\frac{1}{2}}$ regularization. 

% \item For a vector $\bm{a}$, the $\ell_1-\ell_2$ regularization is defined by:
%     \begin{equation}
%         \label{prox_L12}
%      r(\bm{a}) = \|\bm{a}\|_1 - \|\bm{a}\|_2.
%     \end{equation}

%   The proximal operator is given by:
%   \begin{itemize}
%       \item If $\|\bm{a}\|_{\infty} > \lambda$, we have $\bm{prox}_{\ell_1 - \ell_2}(\bm{a;\lambda}) = \frac{\bm{z(\|\bm{z}\|_2 + \lambda)}}{\|\bm{z}\|_2}$, where $\bm{z} = \bm{prox}_{\ell_1(\bm{a};\lambda)}$.
%       \item If \(\|\bm{a}\|_{\infty} \leq \lambda\), then \(\bm{x}^* := \bm{prox}_{\ell_1 - \ell_2}(\bm{a}, \lambda)\) is an optimal solution if and only if the following conditions hold:
% \begin{itemize}
 
%     \item \(x_i^* = 0\), \(|a_i| < \|\bm{a}\|_{\infty}\) and \(\|\bm{x}^*\|_2 = \|\bm{a}\|_{\infty}\) for all \(i\). 

% \end{itemize}

% The optimality condition implies that there are infinitely many solutions for \(\bm{x}^*\). Among these, we select the solution where:
% \begin{itemize}
%     \item \(x_i^* = \text{sign}(a_i) \|\bm{a}\|_{\infty}\) for the smallest index \(i\) satisfying \(|a_i| = \|\bm{a}\|_{\infty}\), and
%     \item the remaining coefficients are set to zero.
% \end{itemize}
%   \end{itemize}
%    Substantial work has been done to derive analytical solutions for the proximal operator of the L12 metric \citep{esser2013method, yin2014ratio, lou2015computing, lou2018fast}. 
%    In addition, various numerical algorithms have been proposed to compute these solutions.

\item The transformed $ \ell_1$ (TL1) \citep{zhang2018minimization,nikolova2000local,zhang2014minimization} for a vector $\bm{a} \in \mathbb{R}^r$ is defined as follows:
\begin{equation}
    \label{TL1}
    \text{TL1}_b(\bm{a}) =  \sum_{i=1}^n t_b(a_i),
\end{equation}
where the function $t_b(a_i) = \frac{(b+1)|a_i|}{b+ |a_i|}  $ with a nonnegative parameter  $b \in (0, +\infty)$ has the following asymptotic behaviors,
\begin{equation}
\label{TL1limit}
 \lim_{b\to 0^+} t_b(a_i) = I_{\{a_i \neq 0\}} , \hspace{0.5cm} \lim_{b\to + \infty} t_b(a_i) = |a_i|, 
\end{equation}
for the indicator function $ I_{\{a_i \neq 0\}}$. When acting on a vector $\bm{a} \in \mathbb{R}^r$, the TL1 function interpolates between the $\ell_0$ and $\ell_1$ norms, i.e.,
\begin{equation}
    \begin{split}
       & \lim_{b\to0^+}\text{TL1}_b(\bm{a}) = \sum_{i=1}^r  I_{\{a_i \neq 0\}}= ||\bm{a}||_0, \hspace{1.2cm}\\
    &\lim_{b\to+\infty }\text{TL1}_b(\bm{a}) = \sum_{i=1}^r |a_i|= ||\bm{a}||_1.
     \end{split}
\end{equation}
%One notable property of the TL1 penalty is that its proximal operator has a closed-form analytical solution for all values of $b$. 
The proximal operator for the TL1 functional  \citep{zhang2014minimization} is given by
\begin{equation}
\label{prox_tl1}
  \begin{split}
      &\prox_{\text{TL1}}(\bm{a}; \lambda)=\\
      &\begin{cases}
    \bigg[\frac{2}{3}(b+ |\bm{a}|)\text{ cos }\frac{\phi(\bm{a})}{3} -\frac{2}{3}b + \frac{|\bm{a}|}{3}\bigg] \quad &\text{if}\quad |\bm{a}| > \theta, \\
    0 &\text{if}\quad |\bm{a}|\leq \theta,
\end{cases} 
 \end{split}
\end{equation}
with
\begin{equation}
  \begin{split}
      &\phi(\bm{a}) = \text{arccos}\bigg(1-\frac{27\lambda b(b+1)}{2(b+|\bm{a}|)^3}\bigg) \quad \text{and}\quad \\
&\theta=\begin{cases}
    \lambda \frac{b+1}{b} \quad &\text{if}\quad \lambda \leq \frac{b^2}{2(b+1)},\\
    \sqrt{2\lambda(b+1)}-\frac{b}{2} \quad  &\text{if}\quad \lambda >\frac{b^2}{2(b+1)}.
    \end{cases}
     \end{split}  
\end{equation}
\end{itemize}

\subsection{Proximal operators for inter-group sparsity regularization}\label{sect:appendB}
For inter-group sparsity, the regularization $f(\cdot)$ is applied on a non-separable norm. Recall we have from \eqref{eq:A-update2} that
\begin{equation}\label{eq:A-update3}
\argmin_{\bm a}\sum_{l=1}^k \left(\lambda f(\|\bm{a}^{\mathcal{G}_l}\|_2) + \frac{\rho}{2}\|\bm{u}_i^{\mathcal{G}_l}-\bm{a}^{\mathcal{G}_l}-\bm{c}_i^{\mathcal{G}_l}\|_2^2\right),
\end{equation}
with $p=2$. This expression implies that for each group $l,$ it amounts to solving
\begin{equation}\label{eq:prox4norm}
   \argmin_{\bm a}\lambda f (\|\bm a\|_2) + \frac {\rho} 2 \|\bm a - \bm v\|_2^2,
\end{equation}
where $\bm v=\bm{u}_i-\bm c_i$ is given as input when minimizing over $\bm a.$ Following the derivations in \cite{ke2024generalized}, the optimal solution $\bm a$ is along the same direction as $\bm v$ with its magnitude being a proximal operator of $\|\bm v\|_2.$ In other words, the closed-form solution of \eqref{eq:prox4norm} is given by 
\[
\frac{\h v}{\|\h v\|_2} \prox_f\bigg(\|\h v\|_2; \frac {\lambda}{\rho}\bigg).
\]

% Assume that the components of a vector $\bm{a}\in \mathbb{R}^r$ are from $k$ groups denoted by $\mathcal{G}_l$ for $l \in \{1,\cdots, k\}$. The minimization problem such as (\ref{eq:A-update}) can be decomposed into groups such that for each $l \in \{1,\cdots, k\}$,
% \begin{equation}
%     \label{eqn:a_g-update}
% \bm{a}_{\mathcal{G}_l}  \in \arg\min_{\bm{a} \in \mathbb{R}^{|\mathcal{G}_l|}}\lambda f(\|\bm{a}_{\mathcal{G}_l}\|_2) + \frac{\rho}{2} \|\bm{a}_{\mathcal{G}_l}-\bm{u}_{\mathcal{G}_l}+\bm{c}_{\mathcal{G}_l}\|_2^2
% \end{equation}
% The $\bm{a}$-update for this problem is given by \citep{ke2024generalized},
% \[\bm{a}{\mathcal{G}_l}^{\tau+1} = \frac{\bm{u}_{\mathcal{G}_l}^{\tau} -\bm{c}_{\mathcal{G}_l}^{\tau}}{\|\bm{u}_{\mathcal{G}_l}^{\tau} -\bm{c}_{\mathcal{G}_l}^{\tau}\|}_2 \text{\prox}_f \bigg(\|\bm{u}_{\mathcal{G}_l}^{\tau} -\bm{c}_{\mathcal{G}_l}^{\tau}\|_2; \frac{\lambda}{\rho}\bigg)\]
% where $\prox$ denotes the proximal operator of the function $f$. For each column vector $\bm{a}$, we update the components of each group sequentially. As shown in \cite{ke2024generalized}, the solution $\bm{a}_{\mathcal{G}_l}^{\tau+1}$ is a stationary point of the $\bm{a}$-subproblem given in (\ref{eqn:a_g-update}).

% \ekaterina{Make sure the first letters of all words of the journals in the references are capitalized.} \gokul{done}

% \ekaterina{Make sure all preprints in the references are still preprints and not published somewhere.} \gokul{done}

%\bibliography{ref}
%\bibliographystyle{plain}
\bibliographystyle{IEEEtran}
\bibliography{IEEEabrv,ref}

\section{Biography}

\begin{IEEEbiography}
[{\includegraphics[width=1.5in,height=1.25in,clip,keepaspectratio]{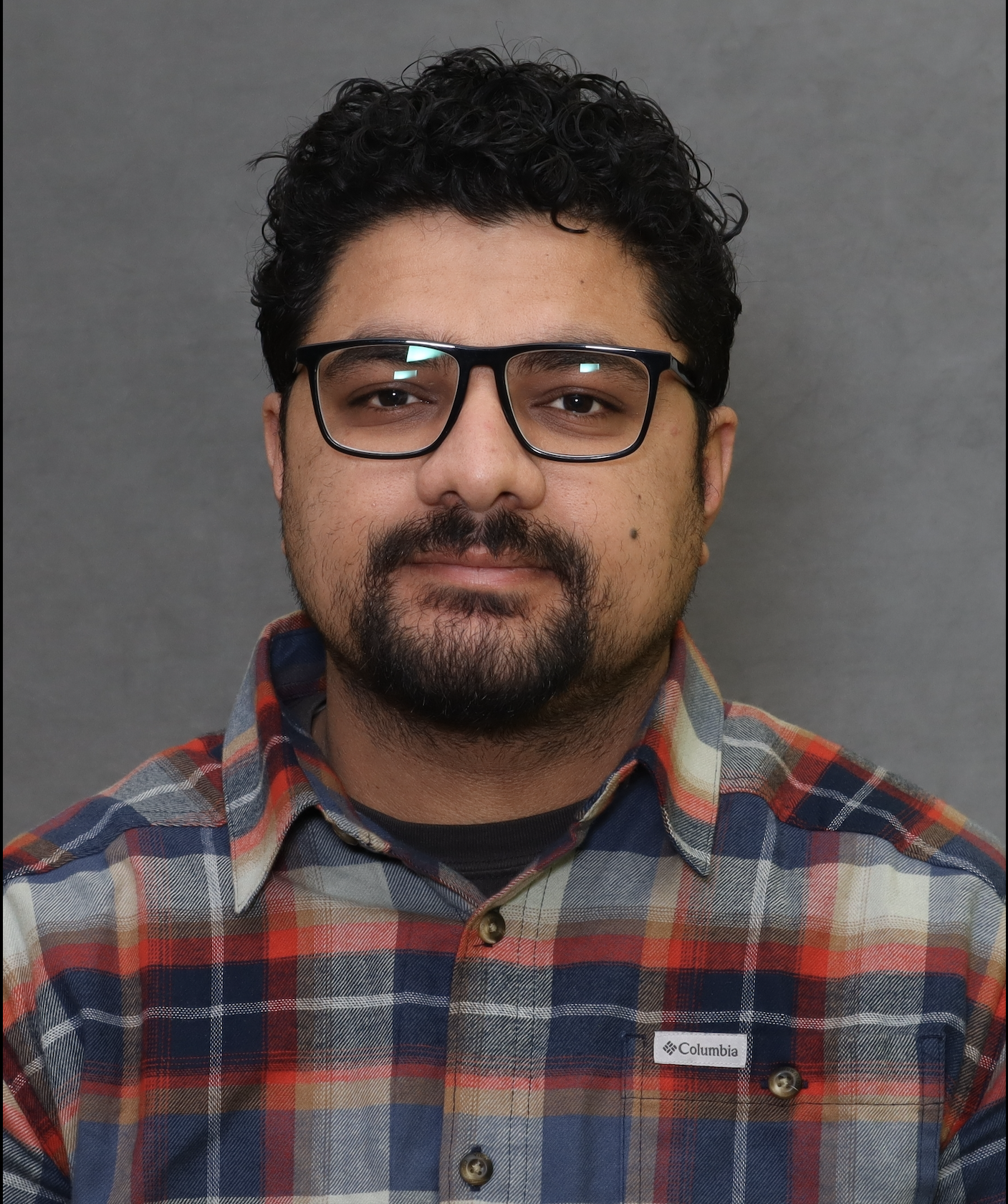}}]
{Gokul Bhusal} received  the B.S. degree in mathematics from the University of Southern Mississippi, Hattiesburg, MS, USA, in 2020. He is currently pursuing the Ph.D. degree in applied mathematics at Michigan State University, East Lansing, MI, USA. His research interests include graph-based methods, active learning, and Image processing. 
\end{IEEEbiography}
\begin{IEEEbiography}
[{\includegraphics[width=1in,height=1.25in,clip,keepaspectratio]{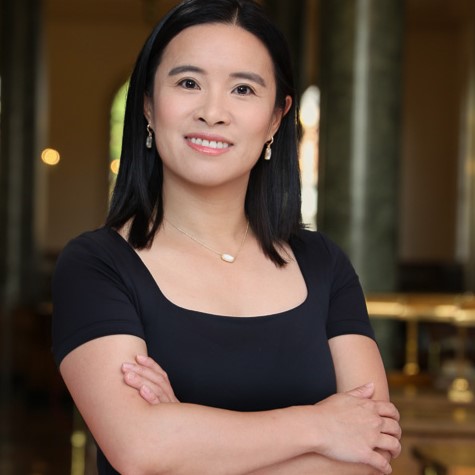}}]{Yifei Lou} holds a joint position in the Department of Mathematics and the School of Data Science and Society (SDSS) at the University of North Carolina at Chapel Hill. She was on the faculty in the Mathematical Sciences Department at  the University of Texas Dallas from  2014 to 2023, first as an Assistant Professor and then as an Associate Professor. She
received her Ph.D. in Applied Math from the University of California Los Angeles (UCLA) in 2010. 
After graduation, she was  a postdoctoral fellow at the School of Electrical and Computer Engineering Georgia Institute of Technology, followed by another postdoc training  at the Department of Mathematics, University of California Irvine from 2012-2014.  Dr.~Lou received the National Science Foundation CAREER Award in 2019.
Her research interests include compressive sensing and its applications,
image analysis (medical, hyperspectral, and seismic imaging),
 and (nonconvex) optimization algorithms. 
\end{IEEEbiography}

\begin{IEEEbiography}
[{\includegraphics[width=1in,height=1.25in,clip,keepaspectratio]{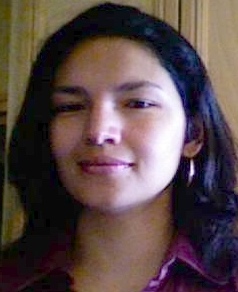}}]
{Cristina Garcia-Cardona} received the B.Sc. degree in electrical engineering from Universidad de Los Andes, Colombia, the M.Sc. degree in emergent computer sciences from Universidad Central de Venezuela, and the Ph.D. degree in computational science from Claremont Graduate University and San Diego State University Joint Program, CA, USA. She is currently a Staff Scientist with the Computer, Computational and Statistical Sciences Division, Los Alamos National Laboratory, Los Alamos, NM, USA. Her research interests include inverse problems, sparse representations, graph algorithms, and machine learning applications.
\end{IEEEbiography}

\begin{IEEEbiography}[{\includegraphics[width=1in,height=1in,clip,keepaspectratio]{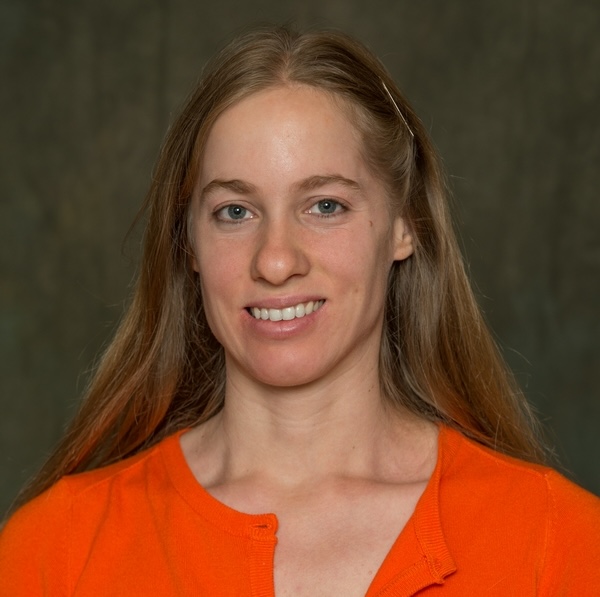}}]{Ekaterina Merkurjev} holds a joint position in the Department of Mathematics and the Department of Computational Mathematics, Science and Engineering (CMSE) at Michigan State University as an Assistant Professor. Prior to Michigan State University, she was a University of California President's Postdoctoral Fellow at the University of California, San Diego. She received her Ph.D. in Applied Mathematics from the University of California, Los Angeles, in 2015 under the guidance of Dr. Andrea Bertozzi. Her research interests include graph-based methods, semi-supervised learning, image processing, and designing algorithms for data with limited labeled samples.
 \end{IEEEbiography}
 
\end{document}